\documentclass[11pt,a4paper]{article} %fontsize option: 10pt,11pt,12pt
\usepackage{authblk}%mix affiliations
\usepackage{amsmath,amsthm}
\usepackage{amssymb}
\usepackage{color}

\usepackage{algorithm}
\usepackage{algorithmic}
\usepackage{enumerate}
\usepackage{graphicx}
\usepackage{subfig}
\usepackage{float}
\usepackage{booktabs}
\usepackage{multirow}
\usepackage{fancyhdr}
\usepackage{enumitem}

\usepackage[hidelinks]{hyperref}
\hypersetup{
  colorlinks   = true, 
  urlcolor     = red, 
  linkcolor    = red, 
  citecolor   = red 
}
\makeatletter
\renewcommand{\theequation}{\arabic{section}.\arabic{equation}}
\@addtoreset{equation}{section} \makeatother

\newcommand{\ud}{\mathrm{d}}

\newcommand{\dx}{\,\mathrm{d}x}
\renewcommand{\vec}[1]{\mbox{\boldmath$#1$}}

\DeclareMathAlphabet{\mathsfsl}{OT1}{cmss}{m}{sl}

\DeclareMathOperator{\Tr}{Tr}
\usepackage{enumitem}
\DeclareMathOperator*{\argmin}{arg\,min}
\setlist{nosep}

\makeatletter

\newcommand{\supp}{\operatorname{supp}}
\newcommand{\R}{\mathbb{R}}
\newcommand{\Rmnum}[1]{\expandafter\@slowromancap\romannumeral #1@}
\makeatother

\newcommand{\E}{\mathbb{E}}
\theoremstyle{plain}
\newtheorem{theorem}{Theorem}[section]
\newtheorem{proposition}[theorem]{Proposition}
\newtheorem{lemma}[theorem]{Lemma}

\theoremstyle{definition}
\newtheorem{definition}[theorem]{Definition}
\newtheorem{assumption}[theorem]{Assumption}
\theoremstyle{remark}
\newtheorem{remark}[theorem]{Remark}
\newtheorem*{proposition*}{Proposition}
\makeatletter
\renewenvironment{proof}[1][\proofname]{\par
  \pushQED{\qed}%
  \normalfont
  \topsep0pt\partopsep0pt % no space before
  \trivlist
  \item[\hskip\labelsep
        \itshape
    #1\@addpunct{.}]\ignorespaces
}{%
  \popQED\endtrivlist\@endpefalse
  \addvspace{6pt plus 6pt} % some space after
}
\makeatother

\makeatletter
\def\thanks#1{\protected@xdef\@thanks{\@thanks
        \protect\footnotetext{#1}}}
\makeatother



\begin{document}
%\abovedisplayskip=2.1pt plus 2.1pt
%\abovedisplayshortskip=0pt plus 2.1pt
%\belowdisplayskip=2.1pt plus 2.1pt
%\belowdisplayshortskip=2.1pt plus 2.1pt

\date{}

\title{DPOT: A DeepParticle method for Computation of Optimal Transport with convergence guarantee\thanks{YY's research is funded by China Scholarship Council NO. 202306280108. ZW's research is funded partially by NTU SUG-023162-00001, MOE AcRF Tier 1 Grant RG17/24.}}

\author[1,2]{ Yingyuan Li}
\author[2]{Aokun Wang}
\author[2]{Zhongjian Wang$^*$\thanks{\small Corresponding E-mail\,: zhongjian.wang@ntu.edu.sg (Z.Wang).}}
\affil[1]{\small\it School of Mathematics and Statistics, Xi'an Jiaotong University, Xi'an 710049, China}
\affil[2]{\small\it Division of Mathematical Sciences, School of Physical and Mathematical Sciences, Nanyang Technological University, 21 Nanyang Link, 637371, Singapore}

\maketitle

%\linenumbers
\noindent{\bf{Abstract.}} {In this work, we propose a novel machine learning approach to compute the optimal transport map between two continuous distributions from their unpaired samples, based on the DeepParticle methods. The proposed method leads to a min-min optimization during training and does not impose any restriction on the network structure. Theoretically we establish a weak convergence guarantee and a quantitative error bound between the learned map and the optimal transport map. Our numerical experiments validate the theoretical results and the effectiveness of the new approach, particularly on real-world tasks.}
\\


%---------------------------------------------------------------------
%            useful AMS classifications (2000)
%---------------------------------------------------------------------
%35Q30: Stokes and Navier--Stokes equations
%35R30: inverse problem for pde \\
%35K55 Nonlinear PDE of parabolic type
%49Q10: optimization of shape other than minimal surface\\
%49Q12: sensitivity analysis\\
%49J35:Minimax problem (existence)\\
%49K35:Minimax problem (necessary and sufficient condition for optim.)\\
%49K40:sensitivity,stability,well-posedness \\
%35B37: pde in connection with control problems ; \\
%65J15:equation with nonlinear operator;\\
%65J20 ill-posed problem;regularization\\
%46N10:applications in optimization ,convex analysis
%-----------------------------------------------------------------------
%\tableofcontents

\section{Introduction}
\label{intro}
Finding a continuous optimal transport (OT) map is challenging in machine learning and plays an essential role in many applications, such as image synthesis \cite{Fan2023neural, Shao2021SRWGANTV}, anomaly detection \cite{Huang2023IWGAN, haloui2018anomaly}, data augmentation \cite{Li2024RF, Bouallegue2020EEG, McHardy2023Augmentation}, domain adaptation \cite{Chai2023Global, Luo2018WGAN, Fan2023neural}. The continuous OT maps are pivotal not only in computer vision and data science but also in areas such as chemistry, physics and earth sciences. 
For instance, in modeling reaction-diffusion processes, flame dynamics, pollutant dispersion, turbulent transport phenomena, and subsurface flows.
However, it suffers from a severe computational burden for complex (e.g. non-logconcave) and continuous distributions.

The origin of OT traces back to the Monge problem.  
Let  $\mathcal{X}$ and $\mathcal{Y}$ denote two metric spaces and $\mu, \nu \in \mathcal{P}\left(\mathcal{X}\right), \mathcal{P}\left(\mathcal{Y}\right)$ denote two probability measures on $\mathcal{X}$ and  $\mathcal{Y}$ correspondingly.
Given a cost function $c: \mathcal{X} \times \mathcal{Y} \to \mathbb{R}$, which represents the cost of transporting one unit of mass from a location $x \in \mathcal{X}$ to a location $y \in \mathcal{Y}$, the classical Monge problem~\cite{monge1781memoire} seeks an OT map $T: \mathcal{X} \to \mathcal{Y}$ that minimizes the total transport cost 
\begin{align}\label{monge-T0}
I(T)=\int_{\mathcal{X}} c(x,T(x))d\mu.
\end{align}

In the literature, there are several major methodological directions in modeling continuous OT map \eqref{monge-T0}. %

\noindent\textsl{Adversarial and minimax OT formulations.} 
Adversarial approaches motivated by dual OT formulations (see Section \ref{sec 2}), most notably Wasserstein GANs (WGANs)~\cite{arjovsky2017wasserstein}, estimate transport maps via min–max optimization between a generator and a discriminator. 
These methods may lack theoretical guarantees for deterministic OT map and often encounter optimization challenges due to the adversarial structure.
While other minimax formulations including~\cite{fan2021scalable, rout2021generative} distinct from adversarial training, these approaches introduce auxiliary variables or require specific architectural constraints on the network structure.

\noindent\textsl{Entropic regularization.} 
To enhance stability and efficiency, entropy-regularized OT (EOT) adds a KL-divergence term to the transport plan $\gamma$, enabling efficient optimization via Sinkhorn iterations~\cite{cuturi2013sinkhorn}. 
This has led to non-minimax extensions including $\gamma$ learning via stochastic gradient ascent with barycentric projection~\cite{seguy2017large}, score-based Langevin sampling from $\gamma$~\cite{daniels2021score} and dual potential learning with energy-based models (EBMs)~\cite{mokrov2023energy}. Despite benefiting from smooth optimization, these methods introduce smoothing bias due to entropic regularization~\cite{xie2023randomized}. Alternatively, EOT can be framed as dynamic Schr{\"o}dinger bridge (SB) problem \cite{de2021diffusion, gushchin2024adversarial}. 
However, these are primarily designed for sampling rather than map recovery, and often involve costly iterative procedures. 

\noindent\textsl{Error estimates for neural OT maps.} 
While most methods focus on designing scalable and stable OT solvers, only a few works provide rigorous, quantitative error bounds for neural network learned OT maps. Classical theoretical results such as Corollary 5.23 in Villani~\cite{villani2009optimal} only establish weak convergence and offer no control over the approximation error.
Fan et al.~\cite{Fan2023neural} provide an error bound via duality gaps.
However, it relies on dual potentials. Korotin et al.~\cite{korotin2019wasserstein} focus on approximating dual potentials instead of transport map itself.
\medskip

\textbf{In this paper,}
we develop a novel method, named DPOT method, for learning Monge maps between two continuous distributions from their discrete unpaired samples and establish its convergence guarantee. 
In addition, we demonstrate the effectiveness of our method through numerical experiments for both synthetic and real-world models. Compared to previous research, our method offers several key advantages in both formulation and performance.

First, the training process under our loss function admits a stable min-min optimization. 
This circumvents oscillatory training process due to the adversarial formations and leads to a stable and interpretable optimization process with convergence guarantees.

Second, our method learns non-entropic optimal transport maps. 
In contrast to Schr{\"o}dinger bridge approaches which capture intermediate stochastic processes, our approach trains a parametric map $T_\theta$ end-to-end. 
This allows direct transport through a single forward pass, avoiding iterative sampling in diffusion-based. 
As a result, our method simplifies the generative process and improves computational efficiency. 

Third, our approach is insensitive with choice of neural network models. 
Unlike \cite{arjovsky2017wasserstein, korotin2019wasserstein} that require Lipschitz constraints or input-convex neural networks (ICNNs), it imposes no architectural or gradient requirements on neural networks during training and is compatible with standard architectures such as MLPs and ResNets.
This structural and computational flexibility makes our approach broadly applicable across diverse learning scenarios. 
\medskip

The remainder of this paper is organized as follows. 
Section \ref{sec 2} presents preliminary results on OT theory.
Section \ref{Alg sec} introduces the main error bound associated with the proposed loss function and our algorithm. 
In Section \ref{sec 4}, we validate the theoretical analysis and demonstrate the effectiveness of our model through numerical experiments. 
Finally, Section \ref{sec 5} concludes the paper by summarizing the key findings and outlining directions for future research. Additional preliminary results on OT theory used in our proofs are provided in Appendix \ref{appendix:proof}, and details of the training procedure are presented in Appendix \ref{appendix:training detail}. 

\paragraph{General notations}
Throughout the paper, we consider metric spaces $\mathcal{X}=\mathcal{Y}=\mathbb{R}^n$. We then assume $\mu,\nu\in\mathcal{P}_{ac,2}(\mathbb{R}^n)$, where $\mathcal{P}_{ac,2}(\mathbb{R}^n)$ is the space of absolutely continuous probability measures on $\mathbb{R}^n$ with finite second order moments. 
If a distribution $\mu$ is absolutely continuous with respect to the Lebesgue measure, then there exists a density function $\rho(x)$ such that $\ud \mu(x)=\rho(x) \ud x$. 
Given a measurable map $T: \mathbb{R}^n \rightarrow \mathbb{R}^n$ and a distribution $\mu$ on $\mathbb{R}^n$, its $L^2$ norm is defined as 
\begin{align*}
\|T\|_{L^2(\mu)}:=\left(\int_{\mathbb{R}^n}\|T(x)\|^2 \ud \mu(x)\right)^{1/2}.
\end{align*}
We denote the identity map by $\mathrm{Id}$ and the support of a measure $\mu$ by $\mathrm{supp}(\mu)$. And for simplicity, we write $a \lesssim b$ if there exists a constant $C > 0$, independent of the main variables under consideration, such that $a \leq C b$.

\section{Preliminaries}\label{sec 2}
In 1942, Kantorovich (see the review in~\cite{kantorovich2006problem})  proposed a relaxed formulation of Monge problem, known as the Monge–Kantorovich problem. 
It takes the form of an infinite-dimensional linear programming:
\begin{align}\label{Kan}
\mathcal{L}_c(\mu,\nu)=\min_{\gamma\in \Gamma(\mu,\nu)} I(\gamma)= \min_{\gamma\in \Gamma(\mu,\nu)}\int_{\mathbb{R}^n  \times \mathbb{R}^n } c(x, y) \ud \gamma(x, y),
\end{align}
where $\Gamma(\mu,\nu)$ is the set of transport plans $\gamma$ such that
\begin{align*}
    \Gamma(\mu,\nu):=\left\{\gamma \in \Gamma(\mathbb{R}^n \times \mathbb{R}^n ):\left(P_1\right)_{\sharp} \gamma=\mu,\left(P_2\right)_{\sharp} \gamma=\nu\right\},
\end{align*}
with $P_i$ denoting the projection onto the $i$-th coordinate, $P_i\left(x_1, x_2\right)=x_i$ for $i=1,2$. 
The notation $(P_i)_\sharp \gamma$ stands for the push-forward of $\gamma$ under $P_i$. The objective in \eqref{Kan} is to minimize the expected transport cost over all joint probability measures on $\mathbb{R}^n \times \mathbb{R}^n$ with fixed marginals $\mu$ and $\nu$. 
Kantorovich also formulated an equivalent dual problem as follows,
\begin{equation}\label{Kandual}
\mathcal{L}_c(\mu, \nu)=\sup_{(\varphi, \phi) \in \mathcal{R}(c)} \int_{\mathbb{R}^n} \varphi(x) \ud \mu(x)+\int_{\mathbb{R}^n} \phi(y) \ud \nu(y),
\end{equation}
where the set of admissible dual potentials is
\begin{align*}
\mathcal{R}(c) \stackrel{\text { def. }}{=}\{(\varphi, \phi) \in L^1(\mu) \times L^1(\nu): \forall(x, y), \varphi(x)+\phi(y) \leq c(x, y)\}.
\end{align*}
Here, $(\varphi, \phi)$ is a pair of continuous functions named Kantorovich potentials. 
The primal-dual optimality conditions allow us to track the support of the optimal transport plan, which is 
\begin{align}\label{plan-kan}
\operatorname{Supp}(\gamma) \subset\partial\mathcal{R}(c)=\{(x, y) \in \mathbb{R}^n \times \mathbb{R}^n: \varphi(x)+\phi(y)=c(x, y)\}.
\end{align}

In the default setting where the cost function is  $c(x, y) = \frac{1}{2}\|x - y\|^2$, the Monge problem can be written
\begin{align}
    \min_{T_\sharp\mu=\nu}&I(T), \text{ where,}\\
    I(T):&=\frac{1}{2}\int_{ \mathbb{R}^n} \|x-T(x)\|^2\ud\mu(x).\label{monge-T}
\end{align} 
We then define the Wasserstein-2 distance between $\mu$ and $\nu$ by \eqref{Kan} with the specific cost mentioned as 
\begin{align*}
W_2(\mu,\nu):=(\mathcal{L}_c(\mu,\nu))^{\frac{1}{2}}.
\end{align*}
Thanks to Theorem 2.9 in \cite{villani2009optimal} and related work in \cite{makkuva2020optimal}, we can express the Wasserstein-2 distance in a way similar to the duality form \eqref{Kandual},
\begin{equation}
\label{Kandual2}
W^2_2(\mu,\nu) = C_{\mu,\nu} - \inf_{\psi \in{\textrm{CVX}}(\mu)} \{ \E_\mu[\psi(X)] + \E_\nu[\psi^*(Y)] \},
\end{equation}
where $\textrm{CVX}(\mu)$ denotes the set of all convex functions in $L^1(\mu)$ and $C_{\mu,\nu}=\frac{1}{2}(\E_\mu[\|X\|^2]+\E_\nu[\|Y\|^2])$ and $\psi^*(y)=\sup_x\langle x, y\rangle-\psi(x)$ refers to the conjugate of convex function $\psi$. When the solution $\psi$ of \eqref{Kandual2} is differentiable, due to \eqref{plan-kan}, the solution of Monge problem \eqref{monge-T0} is given by 
\begin{align}\label{eq_grad_cvx}
T=\nabla\psi.
\end{align}
This approach naturally inspires using gradient of input-convex neural networks to approximate OT maps \cite{rout2021generative}. Finally, we list  the following proposition which ensures the existence and uniqueness of OT map under the quadratic cost $c(x,y)=\frac{\|x-y\|^2}{2}$, and show the \eqref{eq_grad_cvx}.
\begin{proposition}[Theorem 10.28 in \cite{villani2009optimal}] 
  Let $\mu \in \mathcal{P}_{ac,2}(\mathbb{R}^{n}), \nu \in \mathcal{P}_{ac,2}(\mathbb{R}^{n})$  be two continuous distributions where $\mu$ does not charge sets of dimension $n-1$. The density functions, $f$ satisfies $\mu=f\dx$ and $g$ satisfies $\nu = g\mathrm{d}y$. Then there exists a unique transport map $T$ solving \eqref{monge-T0}, such that $T=\nabla\psi$ $\mu-a.e.$ and,
  \begin{align*}
      \mathrm{det}(D^2\psi) = \frac{f}{g\circ\nabla\psi},\quad \mu-a.e.
  \end{align*}
  where $\psi$ is a lower semicontinuous convex function on $\supp(\mu)$. 
\label{prop1} 
\end{proposition} 
\noindent 
We also list some perturbative results of Proposition \ref{prop1} in Appendix \ref{sec:perturb} that we will use in the following up analysis.

\section{DPOT algorithm}\label{Alg sec}
 In Section \ref{sec2sub1}, we introduce our new learning objective and show its consistency.
 Subsequently, Section \ref{sec2sub2} offers both weak and strong convergence estimations for our approximated transport solutions.
 Section \ref{sec 3}
 introduces the discretization form of our learning objective and extends it to conditional settings.
\subsection{New learning objective and its consistency}\label{sec2sub1}
We consider the following functional with a constant $0<\lambda<1$ as the learning objective,
\begin{equation}
 P(T):=\lambda (I_\mu(T))^{\frac{1}{2}
}+W_2(T_{\sharp} \mu, \nu),
\label{DPOT-loss}
\end{equation}
where $I_{\mu}(T)$ is the transport cost of $T$ on $\mu$ defined in \eqref{monge-T}.

For the task of OT modeling, we will apply a neural network to model $T$ and minimize $P$ in \eqref{DPOT-loss}.  When $\lambda=0$, the learning process coincides with the DeepParticle methods \cite{wang2022deepparticle,wang2024deepparticle}. 
While in the proposed DPOT approach, we involve the addition term scaled with $\lambda>0$ to guarantee the optimality of the mapping by the following consistency theorem.

\begin{theorem}\label{thm1}
Under the assumptions of Proposition \ref{prop1}, problem \eqref{DPOT-loss} admits a unique solution $\bar{T}$, which is also the optimal solution of the Monge problem \eqref{monge-T0}, namely, $\forall\,\lambda\in(0,1)$,
\begin{equation*}
    \argmin_T P(T) = \argmin_{T:T_\sharp\mu=\nu} I_\mu(T).
\end{equation*}
\label{thm:minmin}
\end{theorem}
\begin{proof}
By proposition \ref{prop1}, there is a unique OT map $\bar{T}$, solving the Monge problem \eqref{monge-T0} and in this case $P(\bar T) = \lambda W_2(\mu,\nu)$.

For any push-forward map $T:\mathbb{R}^n\to\mathbb{R}^n$, we have 
\begin{align}
P(T) & =\notag\lambda  \left(\int_{\mathbb{R}^n}\frac{\|x-T(x)\|^2}{2} \ud\mu\right)^{\frac{1}{2}}+W_2(T_{\sharp} \mu, \nu)  \notag \\  
& \geq\lambda W_2(T_{\sharp} \mu, \mu)+W_2(T_{\sharp} \mu, \nu) \label{p_ineq1} \\
& \geq\lambda W_2(\mu, \nu)+(1-\lambda) W_2\left(T_{\sharp} \mu, \nu\right) \label{p_ineq2} \\ 
& \geq\lambda W_2(\mu, \nu),\label{p_ineq3}
\end{align}
where we use the definition of $W_2$ distance in \eqref{p_ineq1} and triangle inequality in \eqref{p_ineq2}, and hence \eqref{p_ineq3} implies 
\begin{align*} 
P(T)\geq P(\bar T), \quad\forall \, T.
\end{align*}
We then proceed to examine the necessary condition for $P(T) = P(\bar{T})$ to hold in \eqref{p_ineq3}. \eqref{p_ineq1} is equality if and only if 
\begin{align*}
\lambda\left(\int_{\mathbb{R}^n}\frac{\|x-T(x)\|^2}{2} \ud\mu\right)^{\frac{1}{2}}=\lambda W_2(T_{\sharp}\mu,\mu).
\end{align*}
Then due to the uniqueness, $T$ is the Monge map between $T_\sharp \mu$ and $\mu$. Furthermore, the equality in \eqref{p_ineq3} holds when $W_2(T_{\sharp} \mu, \nu)=0$, which suggests $T_{\sharp} \mu= \nu$. Combining these observations, we conclude that the solution $T$ of problem \eqref{DPOT-loss} is the Monge map between $\mu$ and $\nu$.
\end{proof}

\noindent Compared to traditional adversarial minimax OT methods relying on dual potentials (e.g., WGANs), \eqref{DPOT-loss} yields a stable min-min optimization problem, where minimizations are over $T$ and transport plan $\gamma$ within the Wasserstein-$2$ term. This avoids saddle-point instability and enables the use of unconstrained network architectures other than ICNNs. 

\subsection{Convergence bound}\label{sec2sub2}
In this subsection, we turn to the analyses of the robustness of Theorem \ref{thm1}. First we denote a series of approximated transport map by $T_\epsilon$, which admits the following gap to optimality,
\begin{equation}\label{total_epsilon}
P(T_\epsilon)- \lambda W_2(\mu,\nu)=\epsilon.
\end{equation}
In light of proof of Theorem \ref{thm1}, we decompose $\epsilon$ into three non-negative terms, namely, $\epsilon =\epsilon_1 + \epsilon_2 +\epsilon_3$, where
\begin{align}\label{opt-gap}
&\left\{
\begin{array}{l}
\epsilon_1=(1-\lambda) W_2({T_\epsilon}_\sharp \mu, \nu), \\
\epsilon_2=\lambda\left(\left( \int_{\mathbb{R}^n} \frac{\|x-T_\epsilon(x)\|^2}{2} \ud \mu\right)^{\frac{1}{2}}-W_2({T_\epsilon}_\sharp \mu, \mu)\right), \\
\epsilon_3=\lambda\left(W_2({T_\epsilon}_\sharp \mu, \mu)+W_2({T_\epsilon}_\sharp \mu, \nu)-W_2(\mu, \nu)\right).
\end{array}\right.
\end{align}
Now we denote the OT map from $\mu$ to $\nu_\epsilon = {T_\epsilon}_\sharp\mu$ by $\bar{T}_{\epsilon}$, whose existence follows directly from the Proposition \ref{prop1}.
The convergence bound for $T_\epsilon$ boils down to,
\begin{align}
\label{err_split}
    \|T_{\epsilon}-\bar{T}\|_{L^2(\mu)}\leq\|T_{\epsilon}-\bar{T}_{\epsilon}\|_{L^2(\mu)}+\|\bar{T}_{\epsilon}-\bar{T}\|_{L^2(\mu)}.
\end{align}  

As the first step, we provide an estimate $\|T_\epsilon-\bar{T}_{\epsilon}\|_{L^2(\mu)}^2$ in \eqref{err_split} by $\epsilon_2$, which is a direct consequence of Proposition \ref{prop3.3} and the definition of $\epsilon_2$ as \eqref{opt-gap}. To this end, we make the following assumption to ensure the uniform regularity of optimal transport map $\mu\to\nu_\epsilon$ with respect to $\epsilon$.
\begin{assumption}\label{regularity assumption nu epsilon}
    $\mathrm{supp}(\mu)$, $\{\mathrm{supp}(\nu_\epsilon) \}_\epsilon$ are both $C^2$ and uniformly convex.
On the support we assume that $\mu$ and $\{\nu_\epsilon \}_\epsilon$ have $C^{0,\alpha}$ densities, for some $\alpha\in(0,1)$, satisfying
\begin{align*}
0<c\leq\|\frac{\mathrm{d}\nu_\epsilon}{\mathrm{d}\mathcal{L}^n}\|_\infty\leq C,\quad
0<\bar{c}\leq\|\frac{\mathrm{d}\mu}{\mathrm{d}\mathcal{L}^n}\|_\infty\leq \bar{C},
\end{align*}
where the constants $c, C, \bar{c}, \bar{C}$ are independent of $\epsilon$ and $\mathcal{L}^n$ denotes the Lebesgue measure on $\mathbb{R}^n$.
\end{assumption}
Note that except the uniform convexity of the $\supp(\nu_\epsilon)$, they can be achieved by the regularity of $T_\epsilon$ represented by neural network. The consistency result in Theorem~\ref{thm1} does not require these assumptions.

\begin{lemma}\label{lem:quant bound 1}
Under the assumption of Proposition~\ref{prop1}, and Assumption~\ref{regularity assumption nu epsilon} we have,
\begin{equation}\label{tbd-0}
    \|T_\epsilon-\bar{T}_{\epsilon}\|_{L^2(\mu)}^2\ \lesssim \epsilon_2.
\end{equation}
\end{lemma}

What worth mentioning for the proof of Lemma \ref{lem:quant bound 1} is that the uniformity of the supremum convex modulus of $\psi_\epsilon$ related to $\epsilon$ is derived from the upper bound of $|\mathrm{det}(D^2\psi_\epsilon)|$ as the same argument in \cite{gigli2011holder}.

\medskip
Then we derive bounds for $\|\bar{T}_{\epsilon}-\bar{T}\|_{L^2(\mu)}$ in \eqref{err_split}. Depending on the assumptions, we have two types of approximations.

\subsubsection{Weak convergence}\label{weak bound}
We start with the following proposition from \cite{villani2021topics}, using the cost function specified as $c(x, y) = \frac{1}{2} \|x - y\|^2$. The proof is supplemented in Appendix \ref{proof to quant bound 2} for completeness.
\begin{proposition}[Exercise 2.17 of \cite{villani2021topics}]
 \label{quant bound 2}
Under the assumptions of Proposition~\ref{prop1} and $\nu_{\epsilon}$ converging weakly to some $\nu \in \mathcal{P}_{ac,2}(\mathbb{R}^n)$, $\bar{T}_\epsilon$ converges to $\bar{T}$ in measure with respect to $\mu$, that is,
\begin{align*}
\forall \,\delta>0,\quad \mu[|\bar{T}_{\epsilon}-\bar{T}|\geq \delta]\stackrel{\epsilon\rightarrow 0}{\longrightarrow} 0.
\end{align*}
\end{proposition}

As a direct consequence of Proposition \ref{quant bound 2} and Lemma \ref{lem:quant bound 1}, we have the following theorem.
\begin{theorem}\label{thm: weakly converge}
Let $T_\epsilon$ be a sequence of measurable maps as defined in \eqref{total_epsilon}. Assume that the second order moments of $\nu_\epsilon$ are uniformly bounded with respect to $\epsilon$. Under the assumption of Proposition~\ref{prop1} and Assumption~\ref{regularity assumption nu epsilon}, $T_\epsilon$ converges weakly to $\bar{T}$.
For instance,
\begin{align*}
    \forall \,\delta>0,\quad \mu[|{T}_{\epsilon}-\bar{T}|\geq \delta]\stackrel{\epsilon\rightarrow 0}{\longrightarrow} 0.
\end{align*}
\end{theorem}
\begin{proof}
    $\epsilon_1\to 0$ implies that $W_2(T_{\epsilon\sharp}\mu,\nu)\to0$,
    from which we can derive that $\nu_\epsilon$ converge weakly to $\nu$.
    Then we can use Proposition~\ref{quant bound 2} to show weak convergence from $\bar{T_\epsilon}$ to $\bar{T}$. Then together with Lemma~\ref{lem:quant bound 1}, we have the weak convergence of $T_\epsilon$ to $\bar T$.
\end{proof}

\subsubsection{Quantitative bound}\label{strong bound}
We also provide a quantitative bound given stronger but more technical assumptions.
\begin{theorem}\label{the:quant_thm} 
Under the assumption of Theorem~\ref{thm: weakly converge}, we further assume that $\supp(\nu)\subseteq \supp(\nu_\epsilon)$. Then the following estimate holds:
\begin{equation}\label{L2_eps}
\|T_{\epsilon}-\bar{T}\|_{L^2(\mu)}\lesssim\sqrt{\epsilon_1}+\sqrt{\epsilon_2},
\end{equation}
where $\bar{T}$ is the OT map from $\mu$ to $\nu$.
\end{theorem}

\begin{proof}
Let $(\bar{\psi},\bar{\psi}^*)$ achieves the minimum in the dual form \eqref{Kandual2} between $\mu$ and $\nu$, and $(\bar{\psi}_{\epsilon},\bar{\psi}_{\epsilon}^*)$ solves \eqref{Kandual2} between $\mu$ and $\nu_\epsilon$.
We then define the following expressions,
\begin{align}
S_{\mu,\nu_{\epsilon}}(\bar{\psi}_{\epsilon})&=\int_{\mathbb{R}^n} \bar{\psi}_{\epsilon}(x)\ud\mu(x)+\int_{\mathbb{R}^n} \bar{\psi}_{\epsilon}^{*}(y)\ud\nu_{\epsilon}(y) 
={M(\mu)+M(\nu_\epsilon)}-W_2^2(\mu,\nu_{\epsilon}), \label{eq:1}\\
S_{\mu,\nu}(\bar{\psi})&=\int_{\mathbb{R}^n} \bar{\psi}(x)\ud\mu(x)+\int_{\mathbb{R}^n} \bar{\psi}^{*}(y)\ud\nu(y) ={M(\mu)+M(\nu)}-W_2^2(\mu,\nu), \label{eq:2}\\
S_{\mu,\nu}(\bar{\psi}_{\epsilon})&=\int_{\mathbb{R}^n} \bar{\psi}_{\epsilon}(x)\ud\mu(x)+\int_{\mathbb{R}^n} \bar{\psi}_{\epsilon}^{*}(y)\ud\nu(y) \geq S_{\mu,\nu}(\bar{\psi}),\label{eq:3}
\end{align}
 where $M(\mu'):=\E_{X\sim\mu'}\frac{|X|^2}{2}$ and consider the following decomposition,
\begin{equation*}
\begin{aligned}
S_{\mu,\nu}(\bar{\psi}_{\epsilon})-S_{\mu,\nu}(\bar{\psi})
=\underbrace{S_{\mu,\nu}(\bar{\psi}_{\epsilon})-S_{\mu,\nu_{\epsilon}}(\bar{\psi}_{\epsilon})}_{\Rmnum{1}}
+\underbrace{S_{\mu,\nu_{\epsilon}}(\bar{\psi}_{\epsilon})-S_{\mu,\nu}(\bar{\psi})}_{\Rmnum{2}}.
\end{aligned}
\end{equation*}
Direct computations show, 
\begin{align}
\Rmnum{1}&=\int_{\mathbb{R}^n} \bar{\psi}_{\epsilon}^{*}(y)\ud\nu(y)-\int_{\mathbb{R}^n} \bar{\psi}_{\epsilon}^{*}(y)\ud\nu_\epsilon(y),\label{Rmnum1}\\
\Rmnum{2}& =M(\nu_\epsilon)-M(\nu)+W_2^2(\mu,\nu)
-W_2^2(\mu,\nu_\epsilon)\label{Rmnum2}.
\end{align}
To estimate the first term $\Rmnum{1}$, let $H_{\epsilon}$ be the OT map from $\nu$ to $\nu_\epsilon$, then by the change of variables, we have
\begin{align*}
\int_{\mathbb{R}^n} \bar{\psi}_{\epsilon}^{*}\left(y\right)\ud\nu_{\epsilon}\left(y\right)=\int_{\mathbb{R}^n} \bar{\psi}_{\epsilon}^{*}\left(H_{\epsilon}(y)\right)\ud\nu\left(y\right).
\end{align*} 
Substituting this into equation \eqref{Rmnum1}, we obtain
\begin{align*}
\Rmnum{1}& = \int_{\mathbb{R}^n} (\bar{\psi}_{\epsilon}^{*}(y)-\bar{\psi}_{\epsilon}^{*}(H_{\epsilon}(y)))\ud\nu(y).
\end{align*}
Since $\bar{\psi}^*_{\epsilon}$ is convex and $\supp(\nu)\subseteq\supp(\nu_\epsilon)$, we apply the first-order condition of convexity:
\begin{align*}
 \bar{\psi}_{\epsilon}^{*}(H_{\epsilon}(y))\geq \bar{\psi}_{\epsilon}^{*}(y) + \nabla \bar{\psi}_{\epsilon}^{*}(y) \cdot (H_{\epsilon}(y)-y).   
\end{align*}
Therefore, 
\begin{align*}
I \leq \int_{\mathbb{R}^n} \nabla \bar{\psi}_{\epsilon}^*(y) \cdot\left(y-H_{\epsilon}(y)\right) \ud \nu(y) .
\end{align*}
Using Cauchy–Schwarz inequality leads to
\begin{align}
\Rmnum{1} &\leq  \left(\int_{\mathbb{R}^n} \|\nabla\bar{\psi}_{\epsilon}^{*}(y)\|^2 \ud\nu(y)\right)^{\frac{1}{2}}\left(\int_{\mathbb{R}^n}\|y-H_{\epsilon}(y)\|^{2}\ud\nu(y)\right)^{\frac{1}{2}}\nonumber\\
&\leq R(\mu)\left(\int_{\mathbb{R}^n}\|y-H_{\epsilon}(y)\|^{2}\ud\nu(y)\right)^{\frac{1}{2}}\leq \frac{ R(\mu)}{1-\lambda}\epsilon_1,
\label{tbd-1}
\end{align} 
where $R(\mu)$ stands for a bounded constant with respect to $\mu$ and we employ the following bound in the second inequality
\begin{align*}
\sup_{y\in\supp(\nu)}\|\nabla\bar{\psi}_{\epsilon}^{*}(y)\|\leq R(\mu),
\end{align*}
noting that $\nabla \bar{\psi}_{\epsilon}^{*}$ pushes $\nu_\epsilon$ forward to $\mu$.

Next, for the second term $\Rmnum{2}$, we first approximate:
\begin{align}
M(\nu)-M(\nu_\epsilon)& =\int_{\mathbb{R}^n}\left(\frac{\| y\|^2}{2}-\frac{\| H_{\epsilon}(y) \|^2}{2}\right)\ud\nu(y)\nonumber\\
&\leq\int_{\mathbb{R}^n}\frac{\| y \|+\| H_{\epsilon}(y) \|}{2}(\|y-H_{\epsilon}(y)\|) \ud\nu(y)\nonumber\\
&\leq \max(R(\nu),R(\nu_\epsilon)) W_1(\nu, \nu_\epsilon)\nonumber\\
&\leq \frac{\max(R(\nu),R(\nu_\epsilon)) }{1-\lambda}\epsilon_1.\label{tbd-2}
\end{align} 
Here, $W_1$ denotes the Wasserstein-$1$ distance and we use $W_1(\nu, \nu_\epsilon) \leq W_2(\nu, \nu_\epsilon)$ in the last inequality. Then with the triangle inequality $W_2(\mu,\nu)-W_2(\mu,\nu_\epsilon)\leq W_2(\nu,\nu_\epsilon) $, we estimate the difference of the squared Wasserstein-2 distances as
\begin{equation*}
\begin{aligned}
W_2^2(\mu,\nu)-W_2^2(\mu,\nu_\epsilon)
\leq&(W_2(\mu,\nu)+W_2(\mu,\nu_\epsilon)) W_2(\nu, \nu_\epsilon)\\
\leq & (2 W_2(\mu,\nu)+W_2(\nu, \nu_\epsilon)) W_2(\nu, \nu_\epsilon).
\end{aligned}
\end{equation*}
Because $W_2(\mu,\nu)$ is bounded and $\epsilon_1=(1-\lambda)W_2(\nu_\epsilon,\nu)$, it follows that
\begin{align}\label{tbd-4}
W_2^2(\mu,\nu)-W_2^2(\mu,\nu_\epsilon) \lesssim \frac{\epsilon_1}{1-\lambda}.
\end{align}
Combining the bounds from \eqref{tbd-1}, \eqref{tbd-2} and \eqref{tbd-4}, 
we have
\begin{equation}
 S_{\mu,\nu}(\bar{\psi}_{\epsilon})-S_{\mu,\nu}(\bar{\psi})
\lesssim \epsilon_1.
\end{equation}
By Proposition \ref{prop 10} (under the assumption that $\psi_\epsilon$ is strongly convex as required by the proposition),
\begin{equation}\label{tbd-6}
\left\|\bar{T}_\epsilon-\bar{T}\right\|_{L^2(\mu)}^2 =\|\nabla \bar{\psi}_{\epsilon}-\nabla \bar{\psi}\|_{L^2(\mu)}^2
\lesssim  S_{\mu,\nu}(\bar{\psi}_{\epsilon})-S_{\mu,\nu}(\bar{\psi})
\lesssim \epsilon_1.
\end{equation}
Finally, by adding up \eqref{tbd-0} and \eqref{tbd-6}, we arrive at \eqref{L2_eps}. 
\end{proof}

\begin{remark}
\label{convex-extension}
In theorem \ref{the:quant_thm}, we assume $\supp(\nu)\subset\supp(\nu_\epsilon)$ in order to derive the regularity of $\psi_\epsilon^*$ on the support of $\nu$. The assumption is purely technical, and can be relaxed by the following argument. By Theorem 12.50 (iii) in \cite{villani2009optimal}, assuming that both $\supp(\mu)$ and $\supp(\nu_\epsilon)$ are of class $C^{k+2}$ and uniformly convex, and that density functions of $\mu$ and $\nu_\epsilon$ belong to $C^{k,\alpha}$ and are bounded above and below (assured by smoothness of learning map $T_\epsilon$), one obtains $\psi_\epsilon\in C^{k+2,\alpha}$. Combining this conclusion with Proposition~\ref{prop1}, we can get the strong convexity of $\psi_\epsilon$ without assumption. 
Furthermore, the Monge-Amp\`{e}re equation implies the uniform convexity of $\psi_\epsilon^*$ on $\supp (\nu_\epsilon)$. By applying standard convex extension of $\psi_\epsilon^*$~\cite{azagra2017extension}, it follows that Lipschitz constant of $\psi_\epsilon^*$ on support of $\nu$ is also bounded.
\end{remark}

\subsection{Discrete Algorithm}\label{sec 3}
In this section, we provide a discretization of loss \eqref{DPOT-loss} in practical training, while the details about training strategy is described in Section \ref{al strategy}.
%\subsection{Training objective}\label{al objective}
We also extend the original loss function \eqref{DPOT-loss} to a conditioned setting, where the goal is to learn an OT map conditioned on some physical parameter $\kappa$ (problem dependent). Specifically, given the pairs of distribution $(\mu_\kappa,\nu_\kappa)$ parameterized by $\kappa$ in an admissible set $\mathcal{O}$, we are seeking the optimal transport map parametrized by $\kappa$,
\begin{equation*}
    T(\cdot|\kappa)_\sharp \mu_\kappa = \nu_\kappa, \quad \forall \,\kappa \in \mathcal{O}.
\end{equation*}
Accordingly, the continuous conditional loss function is defined as
\begin{equation}
 \min_T P(T_\theta(\cdot| \kappa))=\mathbb{E}_\kappa\big[\lambda\left(I\left(T_\theta(\cdot|\kappa)\right)\right)^{\frac12}+W_2\left(T_\theta(\cdot| \kappa)_{\sharp} \mu_\kappa, \nu_\kappa\right)\big],
\label{condition-DPOT-loss}
\end{equation}
with
\begin{equation*}
\begin{aligned}
&I\left(T_\theta(\cdot| \kappa)\right):=\frac{1}{2} \int_{\mathbb{R}^n}\left\|x-T_\theta(x|\kappa)\right\|^2 \ud \mu_\kappa(x), \\
&W_2\left(T_\theta(\cdot| \kappa)_{\sharp} \mu_\kappa, \nu_\kappa\right)=\min_{\gamma\in \Gamma(\mu_\kappa,\nu_\kappa)}  \int_{\mathbb{R}^n \times \mathbb{R}^n} c(x, y) \ud \gamma(x, y).
\end{aligned}
\end{equation*}

Note that our method aims at learning OT maps between continuous distributions $\mu_\kappa$ and $\nu_\kappa$. In practice, however, we only have access to finite samples from the conditioned distributions. To bridge this gap, we approximate the continuous loss function by empirical measures \cite{fournier2013rate} and optimize it via random batches. Specifically, we first draw $\{\kappa_r\}_{r=1}^{n_\kappa}$ samples from some distribution of admissible set $\mathcal{O}$, then for each $\kappa$, we draw i.i.d. samples $\{x_{i,r}\}_{i=1}^{N_0} \sim \mu_{\kappa_r}$ and $\{y_{j,r}\}_{j=1}^{N_0}\sim\nu_{\kappa_r}$. Although our computation relies on discrete samples, the transport map is parametrized by a neural network that defines a continuous mapping over the entire space. This approach is often referred to as a continuous OT solver \cite{mokrov2023energy}, as it generalizes beyond the support of the discrete samples. 

Then a discretization of loss \eqref{condition-DPOT-loss} inspired by DeepParticle \cite{wang2022deepparticle} reads,
\begin{equation}\label{trainingloss}
\begin{aligned}
\hat P(T_\theta(\cdot|\kappa_r)) = &\frac{\lambda}{n_\kappa} \sum_{r=1}^{n_\kappa} \left( \frac{1}{2N} \sum_{i,j=1}^N \|T_\theta(x_{i,r}|\kappa_r) - x_{j,r}\|^2 \right)^{1/2} \\
&+ \frac{1}{n_\kappa} \sum_{r=1}^{n_\kappa} \left( \frac{1}{2N} \min_{\gamma_r \in \Gamma} \sum_{i,j=1}^N \|T_\theta(x_{i,r}|\kappa_r) - y_{j,r}\|^2 \gamma_{ij,r} \right)^{1/2},
\end{aligned}
\end{equation}
where $\Gamma$ here is the set of $N \times N$ doubly stochastic matrices, representing the discrete approximation of the continuous coupling set $\Gamma(\mu_{\kappa_r},\nu_{\kappa_r})$. Elements in $\Gamma$, $\gamma_r$ with entries $\gamma_{ij,r}$ admits the following constraints,
\begin{equation*}
\begin{cases}
\gamma_{i j,r} \geq 0 \quad \forall i, j=1,\dots,N ,\\
\sum_{i=1}^N \gamma_{i j,r}=1  \quad \forall j=1,\dots,N, \\
\sum_{j=1}^N \gamma_{i j,r}=1\quad \forall i=1,\dots,N.\\
\end{cases}
\end{equation*} 

Since our loss function involves solving a $N\times N$ discrete OT plan $(\gamma_{ij,r})$, which may be computational costly, we can reduce training cost by updating $\gamma_r$ every $n_\gamma$ iterations utilizing gradient descent of $\theta$ in each iteration. 
This optimization scheme maintains convergence while significantly lowering computational overhead. The OT plan is implemented via the POT library~\cite{flamary2021pot}. The detailed training techniques are provided in Appendix \ref{al strategy}. Alternative scalable solvers, such as Randomized Block Coordinate Descent (ARBCD) \cite{xie2023randomized}, improved Alternating Minimization (AM) \cite{guminov2021combination}, and Accelerated Primal-Dual Alternating Minimization Descent (APDAMD) \cite{lin2022efficiency}, can also be incorporated to further accelerate training in parallel, ensuring efficient.

\section{Numerical experiments}\label{sec 4}
We next validate our theories in Section~\ref{Alg sec} and illustrate the performance of the discrete algorithm through several experiments presented. 
In Section \ref{sec: synthetic examples}, we present some synthetic experiments which are inspired by PDE-based solvers for Monge–Amp{\`e}re equations from \cite{benamou2014numerical}.
Section \ref{sec:inv} extends DPOT loss \eqref{DPOT-loss} by a cycle-consistency term enables the modeling of inverse OT mapping. 
Finally, we turn to some real-world problems in Section~\ref{sec:practical}.
\subsection{Synthetic Examples}\label{sec: synthetic examples}
\subsubsection{Mapping non-uniform to uniform on a square}\label{sec:square}
We begin by validating our method on an analytically solvable problem on $\Omega = (-0.25,0.25)^2$. The target distribution $\nu$ is uniform on $\Omega$, while the source distribution $\mu$ follows $\ud\mu=\rho_X \ud x$ where
\begin{align}
&\rho_X\left(x_1, x_2\right)=1+4\left(q^{\prime \prime}\left(x_1\right) q\left(x_2\right)+q\left(x_1\right) q^{\prime \prime}\left(x_2\right)\right)\nonumber \\
&+16(q\left(x_1\right) q\left(x_2\right) q^{\prime \prime}\left(x_1\right) q^{\prime \prime}\left(x_2\right)-q^{\prime}\left(x_1\right)^2 q^{\prime}\left(x_2\right)^2)
\label{square_density}
\end{align}
with 
\begin{equation*}
q(z)=\left(-\frac{1}{8 \pi} z^2+\frac{1}{256 \pi^3}+\frac{1}{32 \pi}\right) \cos (8 \pi z)+\frac{1}{32 \pi^2} z \sin (8 \pi z).
\end{equation*}
And the ground truth OT map $\bar{T}$ from $\mu$ to $\nu$ reads as
\begin{equation}\label{square_exact}
\bar{T}(x_1, x_2) =
\begin{bmatrix}
x_1 + 4 q^{\prime}(x_1) q(x_2) \\
x_2 + 4 q(x_1) q^{\prime}(x_2)
\end{bmatrix},
\end{equation}
which helps with generation of the discrete samples in the training data. 

We apply a unconditioned ResNet network \cite{He2016CVPR} with parametric ReLU (PReLU) activation function to learn this map through \eqref{trainingloss}. Details on the network architecture and training hyper-parameters are provided in Appendices \ref{appendix for NN} and \ref{appendix for square}.

\begin{figure}[htbp]
\centering
\includegraphics[width=0.8\columnwidth]{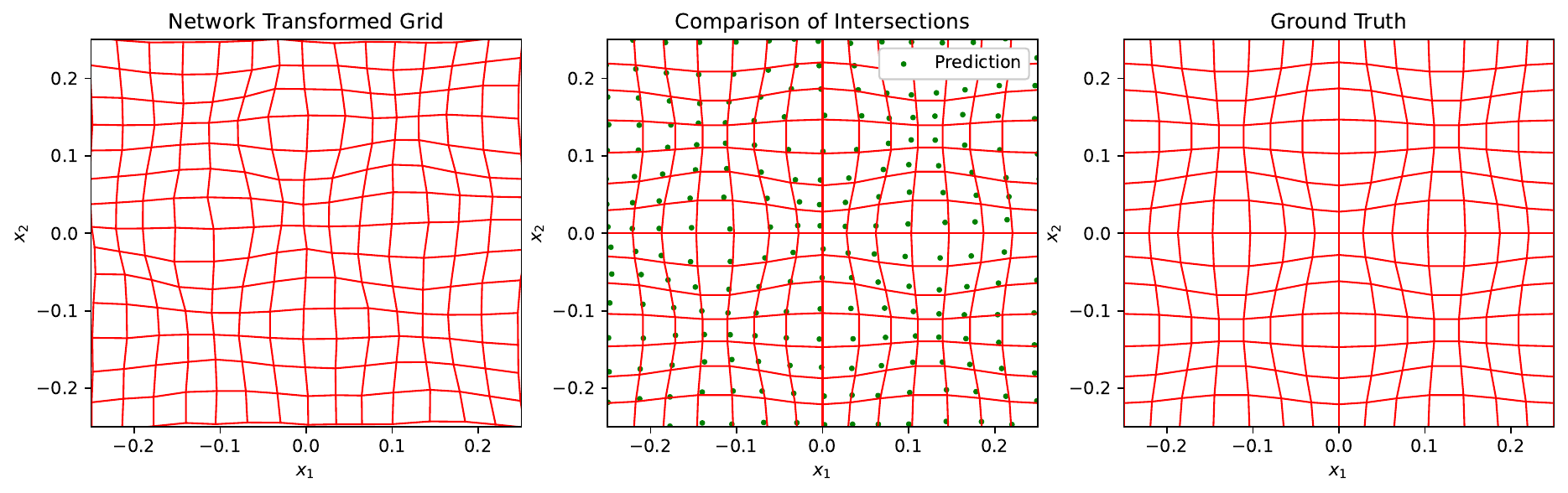}
\caption{Comparison of the transformed Cartesian mesh under $T_\theta(x)$ with $\lambda=0.3$ and $\bar{T}(x)$ (Left: network transformed. Middle: green points are the network predicted of grid intersections. Right: ground truth.) }
\label{fig:square_mesh_lam03}
\end{figure}

Figure~\ref{fig:square_mesh_lam03} illustrates the predicted transformation of a $15 \times 15$ Cartesian mesh using $\lambda = 0.3$, and the relative error of $T_\theta$ under $L^2(\mu)$ metric is $3.5 \times 10^{-2}$. 
Furthermore, in Figure~\ref{square_eps_epoch}, we present that the $L^2$ relative error between $T_\theta$ and $\bar{T}$ correlates well with the optimality gap $\epsilon$, supporting the theoretical result \eqref{L2_eps}.

In Figure~\ref{squ_mesh_com}, we also compare the performance of DPOT with various $\lambda$.
 As $\lambda\rightarrow 1$, the learned map increasingly resembles the identity map $\mathrm{Id}$, which is consistent with the error decomposition \eqref{opt-gap}: when $\lambda=1$, there is no guarantee of $W_2(T_\sharp \mu,\nu)\to0$. Conversely, when $\lambda=0$ (DPOT turns to the conventional DeepParticle \cite{wang2022deepparticle} methods), the absence of regularization leads to failure to finding the OT map.  

\subsubsection{Mapping from one ellipse to another ellipse}\label{sec:ep}
This experiment evaluates the proposed method's ability to learn both unconditioned and conditioned OT maps between two elliptical distributions in $\mathbb{R}^2$. Each ellipse is obtained by applying a transformation to the unit ball. In both settings, the source distribution $\mu$ is fixed and transformed by a matrix $M_x$, while the target distribution $\nu$ is transformed by $M_y(0.2)$ for the unconditioned case and $\nu_\kappa$ belongs to a parameterized family defined by elliptical transformations $M_y(\kappa)$ under the conditioned setting:
\begin{align*}
M_x = \begin{bmatrix} 0.8 & 0\\
0 & 0.4
\end{bmatrix},\quad
 M_y(\kappa) = \begin{bmatrix}0.6 & \kappa\\
\kappa & 0.8
\end{bmatrix}.
\end{align*}
The physical parameter $\kappa\in \mathcal{O}=[-0.5,0.5]$ controls the off-diagonal entries of the target covariance matrix, where varying $\kappa$ induces different shear and rotation effects on the target ellipse.

The closed form of this OT problem can be derived explicitly from
\begin{equation}\label{opt_benamou}
     \bar{T}(x) = M_yR_a M_x^{-1}x,
\end{equation}
where $R_a$ is a rotation matrix given by
\begin{align*}
R_a 
  &= \begin{pmatrix}
      \cos(a) & -\sin(a)\\
      \sin(a) & \cos(a)
     \end{pmatrix}, 
\end{align*}
and the angle $a$ is defined as
\begin{align*}
 \tan(a) = \Tr(M_x^{-1}M_y^{-1}J)/\Tr(M_x^{-1}M_y^{-1})\quad \text{where}\quad J = R_{\pi/2} = 
\begin{pmatrix}0&-1\\1&0\end{pmatrix}. 
\end{align*}

During the training, the hyper-parameter $\lambda$ is set to $0.3$. We exploit two ResNet networks with PRelu activation function and use parameters detailed in Appendix \ref{appendix for ep}.

We first consider unconditioned $T_\theta$ and study the influence of the update frequency $n_\gamma$ and number of batches $n_\kappa$. We find that increasing the number of batches from $n_\kappa = 1$ to $n_\kappa = 10$ reduces the $L^2$ relative error from $6.5 \times 10^{-2}$ to $4.3 \times 10^{-2}$. Among the tested update frequencies $n_\gamma \in \{5, 10, 50, 100\}$, we conclude that $n_\gamma = 10$ yields the best trade-off from Figure~\ref{ngamma_convergence}. It achieves a lower relative error than $n_\gamma = 50$ and $n_\gamma =100$ ($3.4\times 10^{-1}$ and $3.6\times 10^{-1}$), while exhibiting more stable convergence behavior than $n_\gamma = 5$. 

\begin{figure}[htbp]
\centering
\subfloat[$\kappa=-0.15$\label{ellispe_kap1}]{
\includegraphics[height=2.5cm,width=2.5cm]{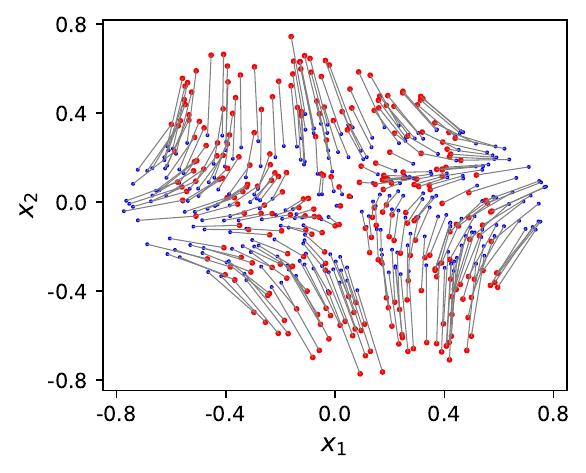}}
\subfloat[$\kappa=-0.47$\label{ellispe_kap2}]{
\includegraphics[height=2.5cm,width=2.5cm]{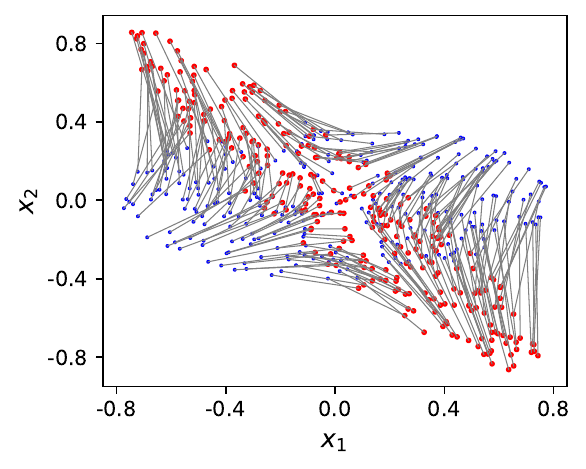}}
\subfloat[$\kappa=0.11$\label{ellispe_kap3}]{
\includegraphics[height=2.5cm,width=2.5cm]{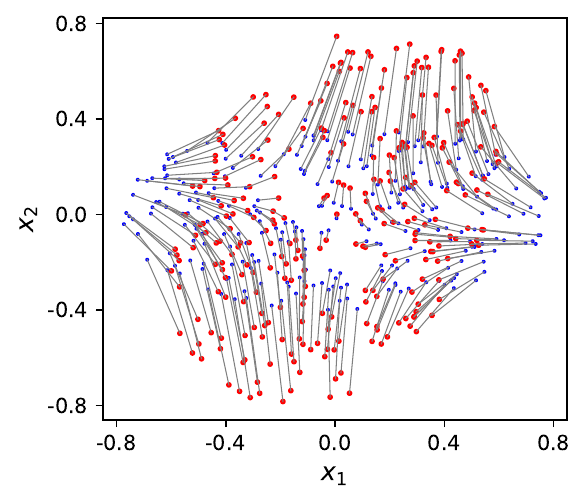}}
\subfloat[$\kappa=0.32$\label{ellispe_kap4}]{
\includegraphics[height=2.5cm,width=2.5cm]{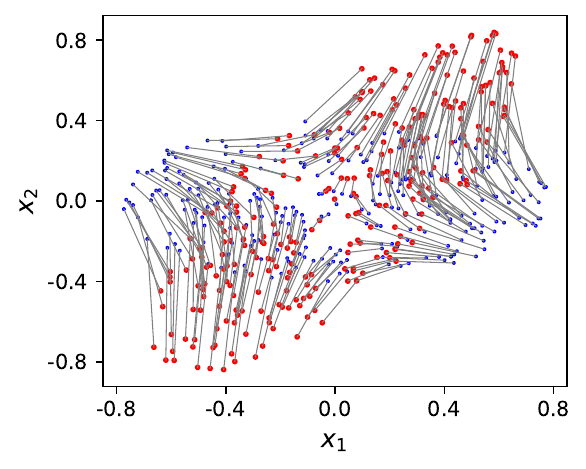}}
\caption{Transport map inferred by conditioned network $T_\theta(\cdot|\kappa)$ (Here, blue and red points represent the source distribution and neural network predictions, respectively.)}
\label{ellispe_mulicon_map_com}
\end{figure}

\begin{figure}[htbp]
\centering
\subfloat[Square transport\label{square_eps_epoch}]{
\includegraphics[height=4cm,width=6cm]{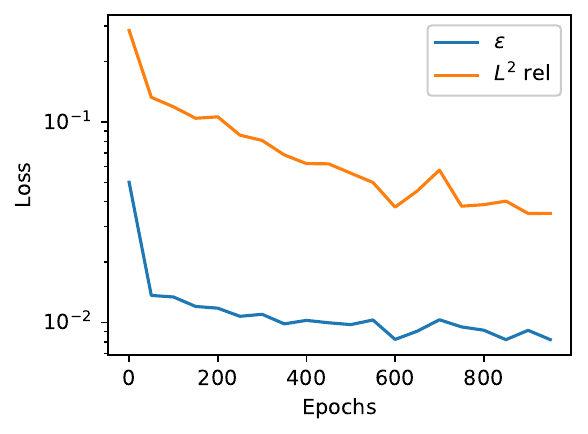}}
\subfloat[Ellipse transport\label{ellipse_cond_loss}]{   \includegraphics[height=4cm,width=6cm]{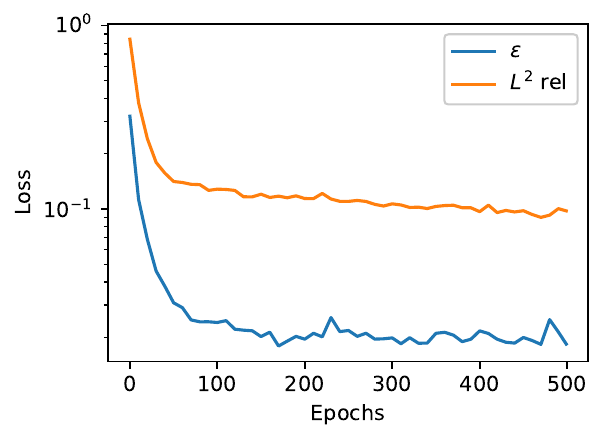}  } 
\caption{Total optimality gap $\epsilon$ and $L^2$ relative error vs. epochs}
\end{figure}

We next consider conditioning on five random values sampled from the admissible set $\mathcal{O} = [-0.5, 0.5]$. Based on findings from the unconditioned case, we use $n_\kappa = 10, n_\gamma=10$ here. After training, we randomly sample new values of $\kappa$ and generate $1\times 10^6$ data as input of the network. As displayed in Figure~\ref{ellispe_mulicon_map_com}, the network recovers conditioned transport maps  within the admissible range. Figure~\ref{ellipse_cond_loss} presents the convergence behavior under this conditioned setting.

\subsubsection{Mapping disjoint two half circles onto the circle } \label{sec:disjoint}
To show the ability of our method modelling discontinuous transport maps, we then consider recovering a mapping from two disjoint half circles to a circle, which is a famous example of singular OT by Luis A. Caffarelli~\cite{villani2009optimal}. 

The experiment includes both an unconditioned scenario, where we minimize \eqref{DPOT-loss}, and a conditioned one, where \eqref{condition-DPOT-loss} is minimized with a geometric parameter $\kappa$ that encodes the distance between the source half-circles.
In both cases, the target distribution $\nu$ is taken as the uniform distribution on a disk of radius $0.85$ centered at the origin. For the unconditioned case, the source distribution $\mu$ is constructed by shifting the left (resp. right) half-disk leftward (resp. rightward) by $0.25$. In the conditioned case, the source distribution $\mu_\kappa$ is generated in the same way, but with each half shifted by $\kappa/2$ in opposite directions.

We set $\lambda=0.3, n_\kappa=30$ and employ two standard MLP architectures for this experiment. A full description of the network configuration and training setup is available in Appendix~\ref{appendix for dis} for reproducibility, as well as the additional experiment results. 

\begin{figure}[bpt]
\centering
\subfloat[$\kappa=0.2$]{ 
\includegraphics[width=0.25\columnwidth]{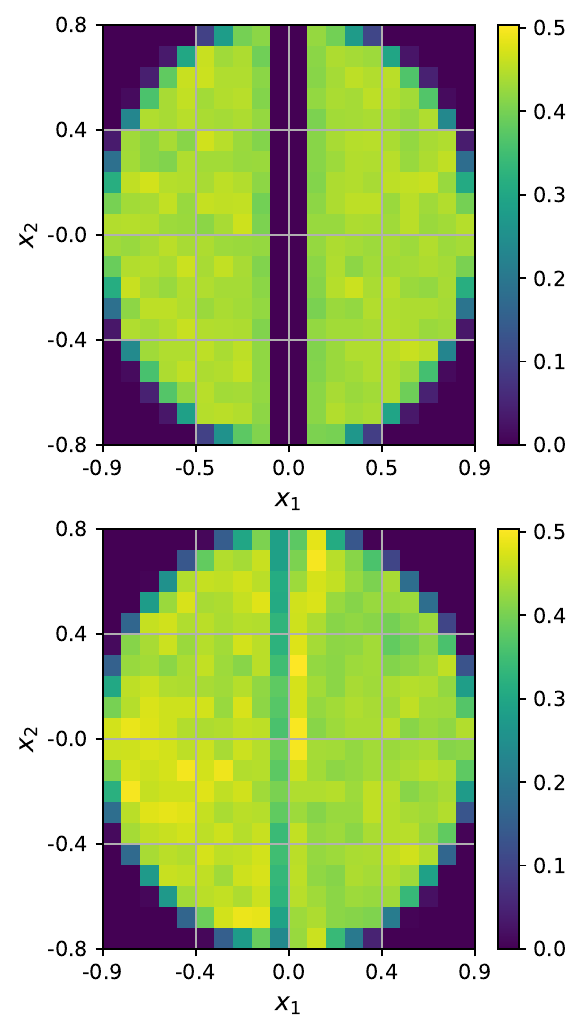}}
\subfloat[ $\kappa=0.4$]{ 
\includegraphics[width=0.25\columnwidth]{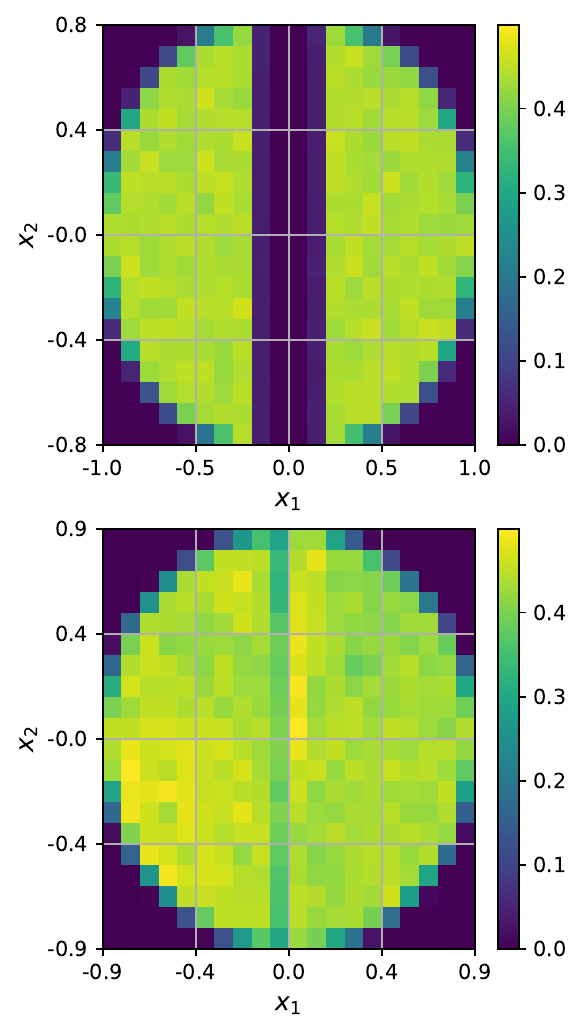}}
\subfloat[$\kappa=0.6$]{ 
\includegraphics[width=0.25\columnwidth]{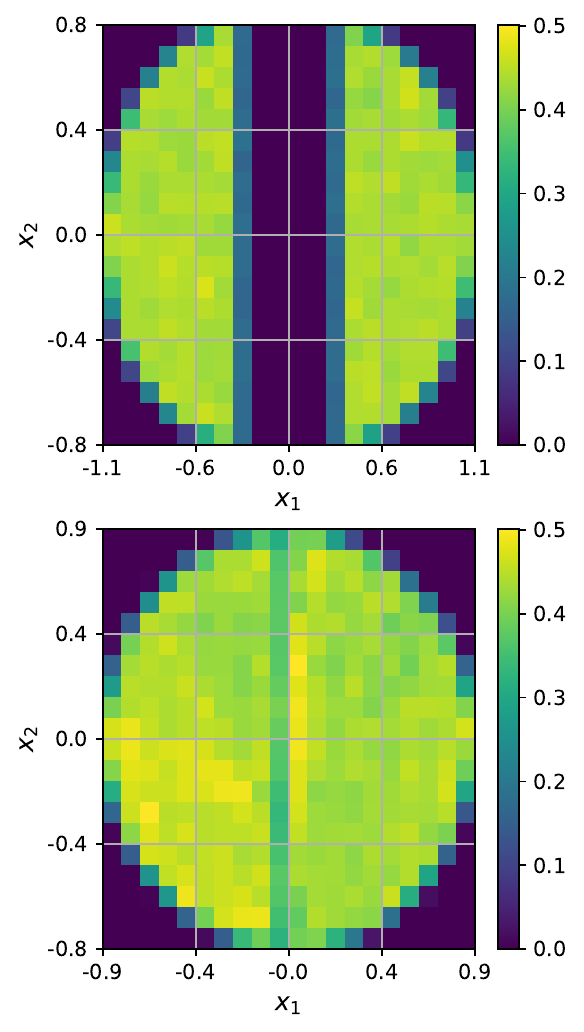}}
\subfloat[$\kappa=0.8$]{ 
\includegraphics[width=0.25\columnwidth]{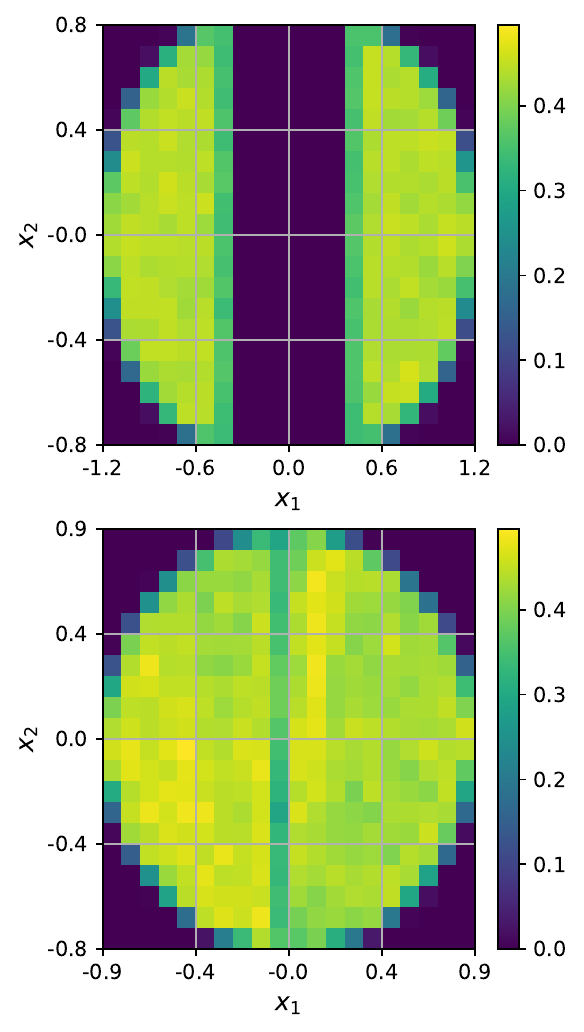}}
\caption{Comparison of source and prediction inferred by conditioned network $T_\theta(\cdot|\kappa)$ with specific $\kappa$ (The first row: source histograms. The second row: neural network predicted histograms.) } 
\label{dis_multcond_hist}
\end{figure}

Under the conditioned case, $n_\kappa =30$ values of $\kappa$ are randomly sampled from the admissible set $\mathcal{O} = [0.0, 1.0]$. Once trained, we evaluate the generalization of the conditioned map $T_\theta(\cdot|\kappa)$ on four unseen values, $\kappa = 0.2,\ 0.4,\ 0.6,\ 0.8$, to assess performance beyond the training samples. As displayed in Figure \ref{dis_multcond_hist}, the predicted distributions closely approximate the target uniform distribution on a disk for all tested $\kappa$ values, illustrating strong generalization of the model to unseen conditioning inputs. 

\subsection{Inverse mapping}\label{sec:inv}
The DPOT method is also capable of recovering inverse mappings $T^{-1}$ by slight modification of learning objective as \cite{korotin2019wasserstein}. More precisely, we consider the training objective $\mathcal{L}=\mathcal{L}_{\mathrm {fwd}}+\mathcal{L}_{\mathrm {inv}}$, where
\begin{equation}\label{res1}
\mathcal{L}_{\mathrm{fwd}}=P(T_\theta)+\|T_\theta^{-1}\circ T_\theta-\mathrm{Id}\|^2_{L^2(\mu)},
\end{equation}
and 
\begin{equation}\label{res2}
\mathcal{L}_{\mathrm{inv}}=P(T_\theta^{-1})+\|T_\theta\circ T_\theta^{-1}-\mathrm{Id}\|^2_{L^2(\nu)}.
\end{equation}
To be noted in \eqref{res2}, $P(T_\theta^{-1})$ is the DPOT objective \eqref{DPOT-loss} defined from $\nu$ to $\mu$.

Analogous to cycle-consistency regularization proposed in \cite{korotin2019wasserstein}, the residue terms added to $P$ in \eqref{res1} and \eqref{res2} are designed to ensure invertibility, namely,
\begin{align*}
    T_\theta^{-1} \circ T_\theta(x) \approx x , \quad T_\theta \circ T_\theta^{-1}(y) \approx y.
\end{align*}

We test learning objective $\mathcal{L}$ in the example mapping a Gaussian mixture at four corners to a single Gaussian centered at origin in \cite{benamou2014numerical}, detailed in Appendix~\ref{Appdendix for inv}. 
During training, we use $\lambda=0.3$ in \eqref{condition-DPOT-loss} and a modified MLP \cite{e2017deepritzmethoddeep} introduced in Appendix \ref{appendix for NN}. Figure \ref{inv_map} displays the  predicted forward and inverse mappings, suggesting the reversibility of our method. The residual errors defined in \eqref{res1} and \eqref{res2} converge to $6.6 \times 10^{-3}$ and $9.0 \times 10^{-3}$, respectively.

\begin{figure}[htbp]
\centering
\includegraphics[width=0.7\columnwidth]{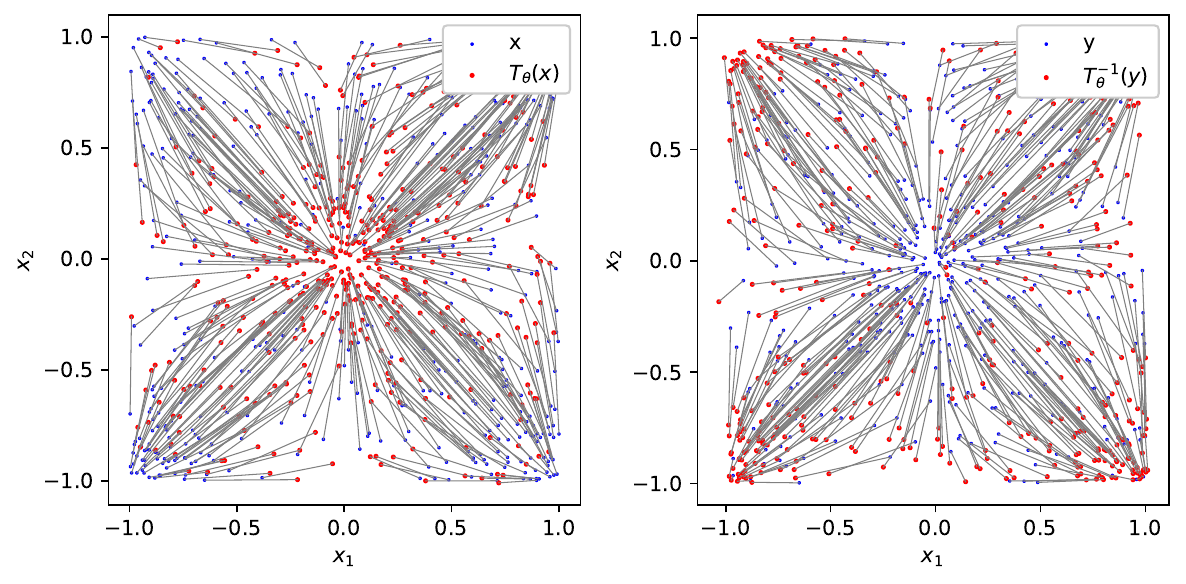}
\caption{Network predicted map for inverse transport}
\label{inv_map}
\end{figure}
\subsection{Practical examples}\label{sec:practical}
\subsubsection{Compartmental Susceptible-Infected-Removed model} \label{sec:csir}
Though DPOT method aims to solve the OT problems, it can also be applied as a continuous sampler trained from discrete data. In this experiment, we employ the DPOT algorithm to approximate the posterior distribution of the parameters in a compartmental Susceptible-Infectious-Recovered model (CSIR)~\cite{cui2023self}, with $d\in\mathbb{N}$ denoting the number of compartments.
The interaction among the individuals is governed by the following ordinary differential equations (ODEs):
\begin{equation}
\left\{\begin{array}{l}
\frac{\ud S_i(t)}{\ud t}=-\beta_i S_i I_i+\frac{1}{2} \sum_{j \in \mathcal{J}_i}\left(S_j-S_i\right), \\
\frac{\ud I_i(t)}{\ud t}=\beta_i S_i I_i-\zeta_i I_i+\frac{1}{2} \sum_{j \in \mathcal{J}_i}\left(I_j-I_i\right), \\
\frac{\ud R_i(t)}{\ud t}=\zeta_i I_i+\frac{1}{2} \sum_{j \in \mathcal{J}_i}\left(R_j-R_i\right),
\end{array}\right.
\label{CSIR_model}
\end{equation}
where $S_i(t), I_i(t), R_i(t)$ represent the susceptible, infected, and recovered individuals in the $i$-th $(i=1, \ldots,d)$ compartment at a given time $t$. Parameters $\beta_i$ and $\zeta_i$ denote the $i$-th infection and recovery rates, respectively. The summation terms in \eqref{CSIR_model} account for the diffusive interaction between the $i$-th compartment and its neighboring compartments set $\mathcal{J}_i=\{i-1,i+1\}$ with $Z_{d+1}=Z_1$ for $Z\in\{S,\ I,\ R\}$. The initial condition of \eqref{CSIR_model} is fixed as,\begin{equation*}
S_i(0)=99-d+i, \quad I_i(0)=d+1-i, \quad R_i(0)=0, \quad i=1, \ldots, d. 
\end{equation*}

One  important application of the CSIR model is to infer the unknown parameters $y=(\beta_1, \zeta_1, \ldots, \beta_d, \zeta_d) \in \mathbb{R}^{2 d}$ from noisy observations of $I_i(t), i=1, \ldots, d$, at 6 equidistant time points on $[0,5]$. 
To this end, we impose a uniform prior $\rho_X$ 
\begin{align*}
\rho_X({x})=\prod_{i=1}^{2d} 1_{[0,2]}\left(x_i\right),
\end{align*}
and simulate \eqref{CSIR_model} with the fourth-order Runge-Kutta method.
Posterior samples are then generated via accept–reject technique using the likelihood in \eqref{CSIR likehood}.

\begin{figure}[htbp]
\centering
\subfloat[$d=1$ \label{fig:csir_marg_1}]{ 
\includegraphics[width=0.3\columnwidth]{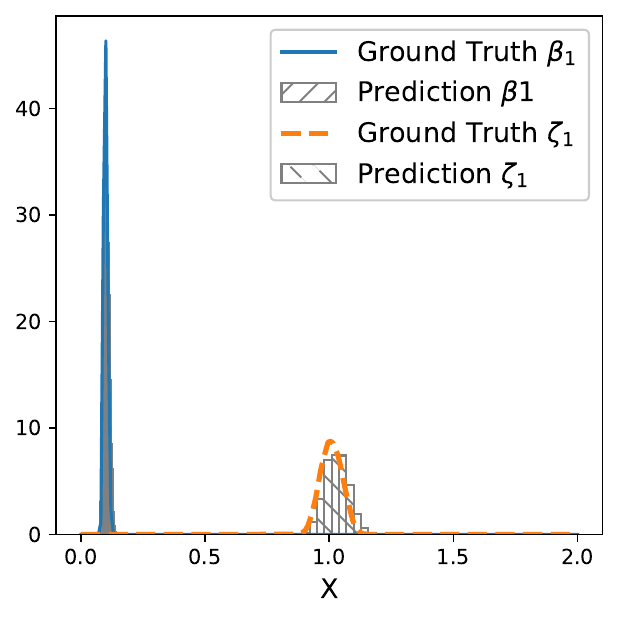}
}
\subfloat[$d=2$ \label{fig:csir_marg_2}]{ 
\includegraphics[width=0.59\columnwidth]{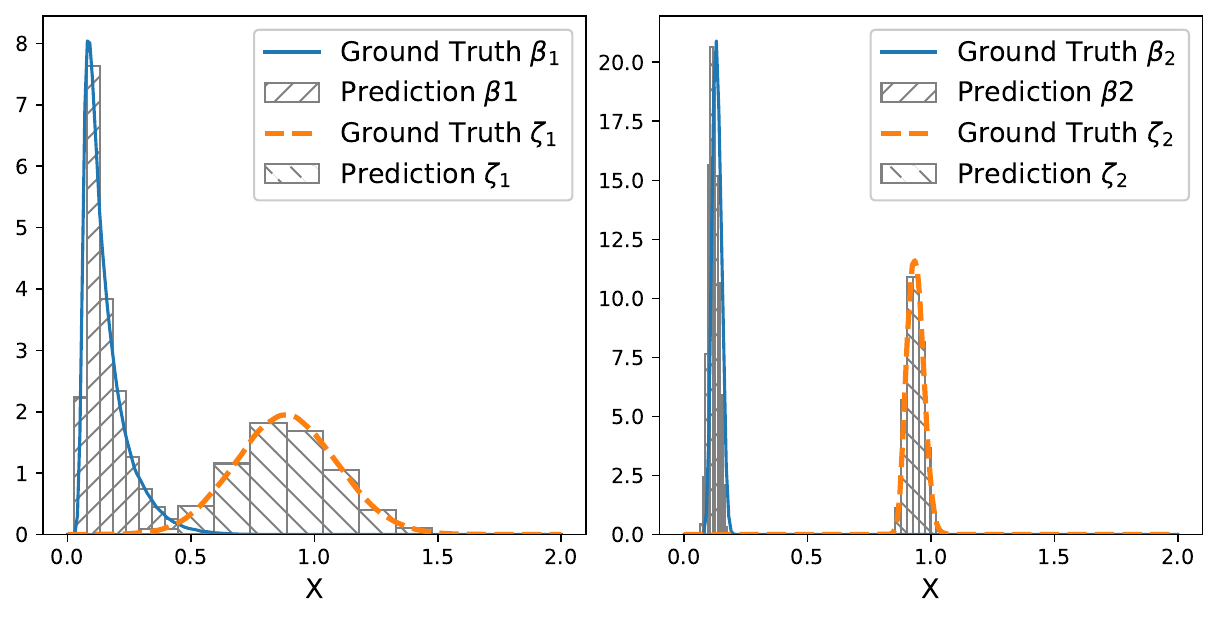}
}

\subfloat[$d=3$ \label{fig:csir_marg_3}]{ 
\includegraphics[width=0.9\columnwidth]{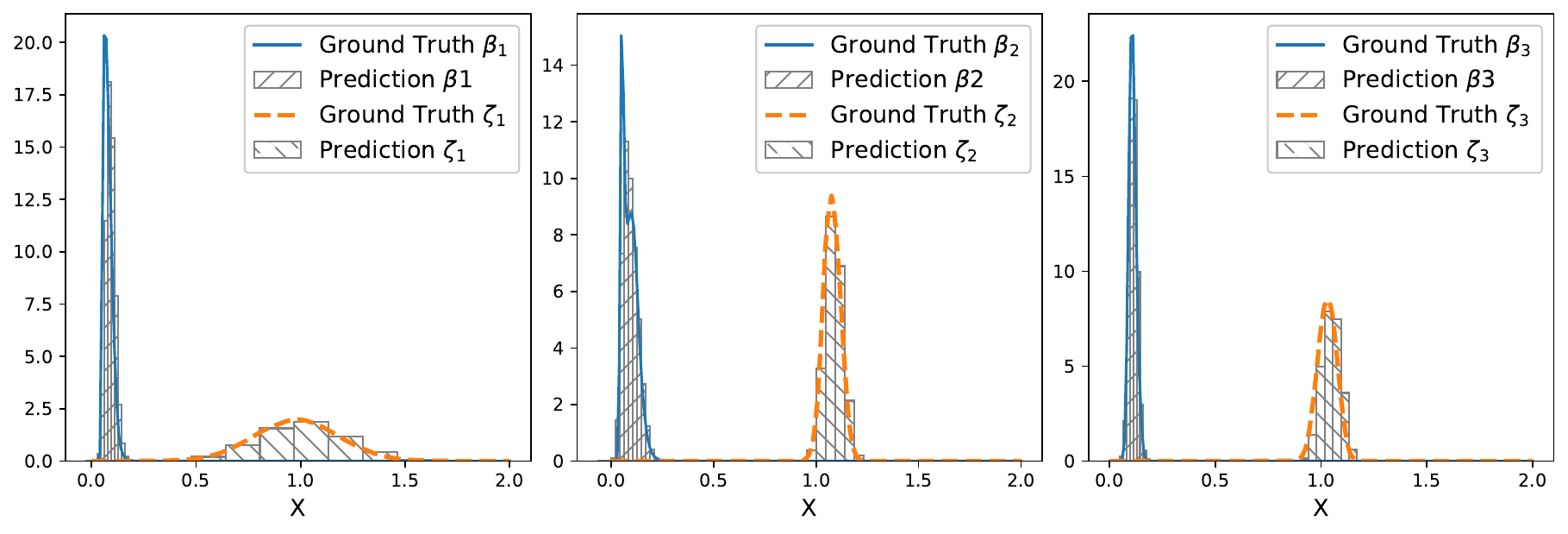}
}

\subfloat[$d=4$ \label{fig:csir_marg_4}]{ 
\includegraphics[width=0.9\columnwidth]{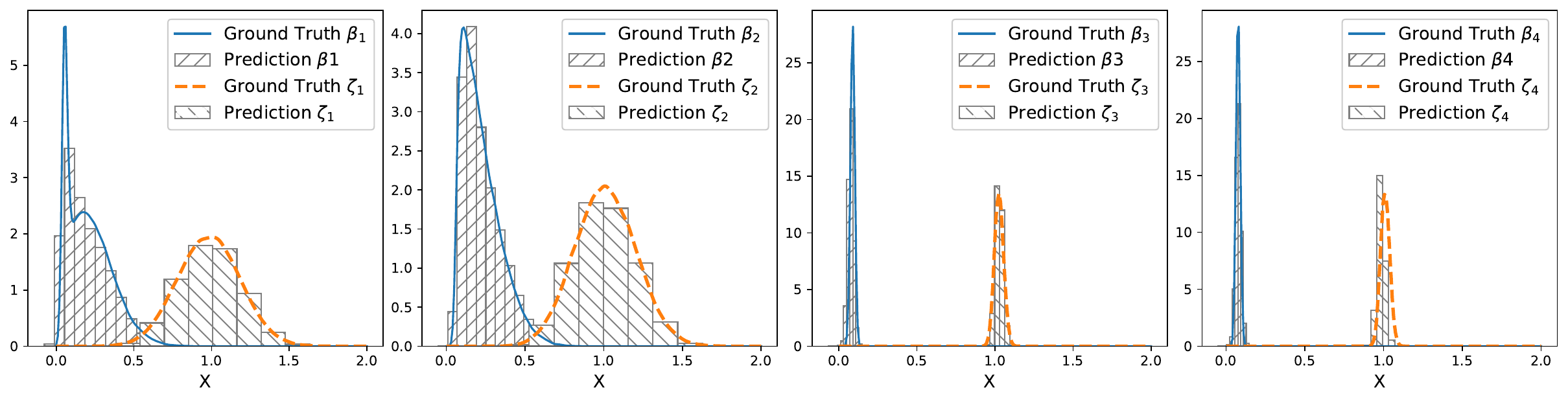}
}
\caption{Marginal density comparison of predicted posterior and ground truth (In each experiment, the total dimension of posterior is $2d$.)}
\label{fig:csir_marg_com}
\end{figure}

We report the performance of DPOT method modeling the posterior distribution of CSIR model in Table \ref{tab:posterior_gen_time}. For each $d = 1, 2, 3, 4$, we compare the traditional accept-reject (AR) sampling method with our learned transport map $T_\theta$, trained in advance.  Remarkably, even when generating a substantially larger number of samples ($1 \times 10^6$), our approach achieves orders-of-magnitude speedups over AR sampling. To further evaluate the quality of the learned transport map, Figure \ref{fig:csir_marg_com} compares the predicted and ground truth posterior marginal distributions for the $i$-th pair $(\beta_i,\zeta_i), i=1,\ldots, d$ with $d=1, 2, 3, 4$. Notably, the learned map $T_\theta$ accurately captures the bimodal characteristics of the posterior distributions across all cases, demonstrating the robustness and efficiency of DPOT comparing with the conventional AR when increasing dimension. 
\begin{table}[htbp]
\centering
\caption{Posterior generation time: accept-reject (AR) vs. neural network $T_\theta$ (NN $T_\theta$).}
\begin{tabular}{cccc}
\toprule
$d$ & Method & Number of samples & Total time (s) \\
\midrule
\multirow{2}{*}{1} & AR &$ 4\times 10^4$ & $5.22\times 10^3$ \\
                  & NN $T_\theta$ & $ 1\times 10^6$ & 0.11 \\
\midrule
\multirow{2}{*}{2} & AR & $ 4\times 10^4$ & $1.01\times 10^4$ \\
                  & NN $T_\theta$ & $ 1\times 10^6$ & 0.11 \\
\midrule
\multirow{2}{*}{3} & AR & $ 4\times 10^4$ & $1.08\times 10^4$ \\
                  & NN $T_\theta$ &$ 1\times 10^6$ & 0.12 \\
\midrule
\multirow{2}{*}{4} & AR & $ 4\times 10^4$ & $1.54\times 10^4$ \\
                  & NN $T_\theta$ & $ 1\times 10^6$ & 0.12 \\
\bottomrule
\end{tabular}
\label{tab:posterior_gen_time}
\end{table}

\subsubsection{Image-to-Image color transfer}\label{sec:image}
\begin{figure}[htbp]
\centering
\subfloat[Original images\label{ColorMLP0}]{ 
\includegraphics[width=0.45\columnwidth]{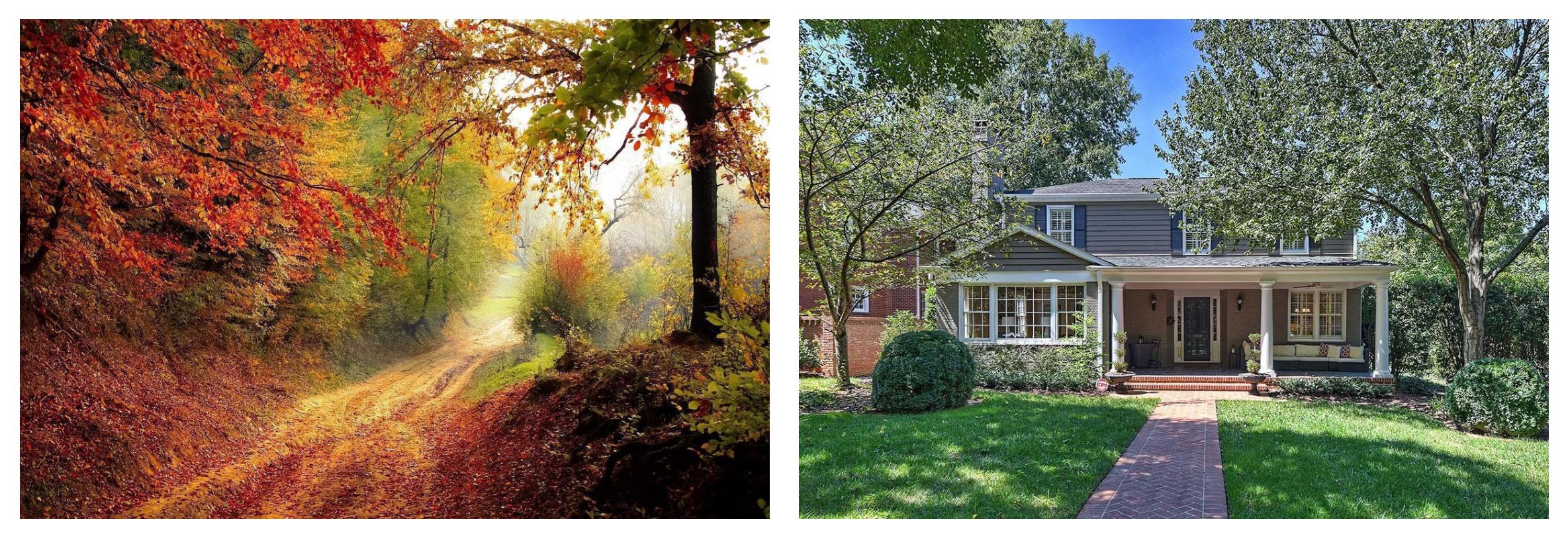}}
\subfloat[MLP+$P_{\lambda=0.3}(T_\theta)$\label{color_fost_MLP}]{ 
\includegraphics[width=0.45\columnwidth]{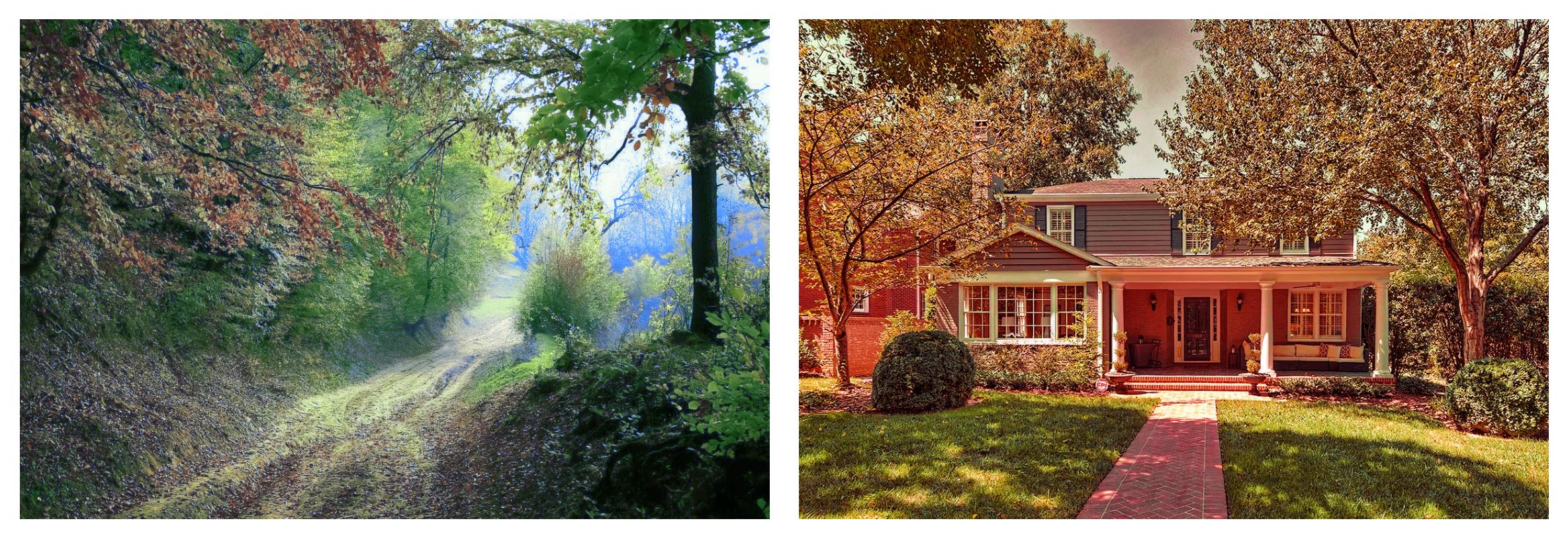}}

\subfloat[ICNN+$P_{\lambda=0.3}(T_\theta)$\label{color_icnnlambN0_fost}]{ 
\includegraphics[width=0.45\columnwidth]{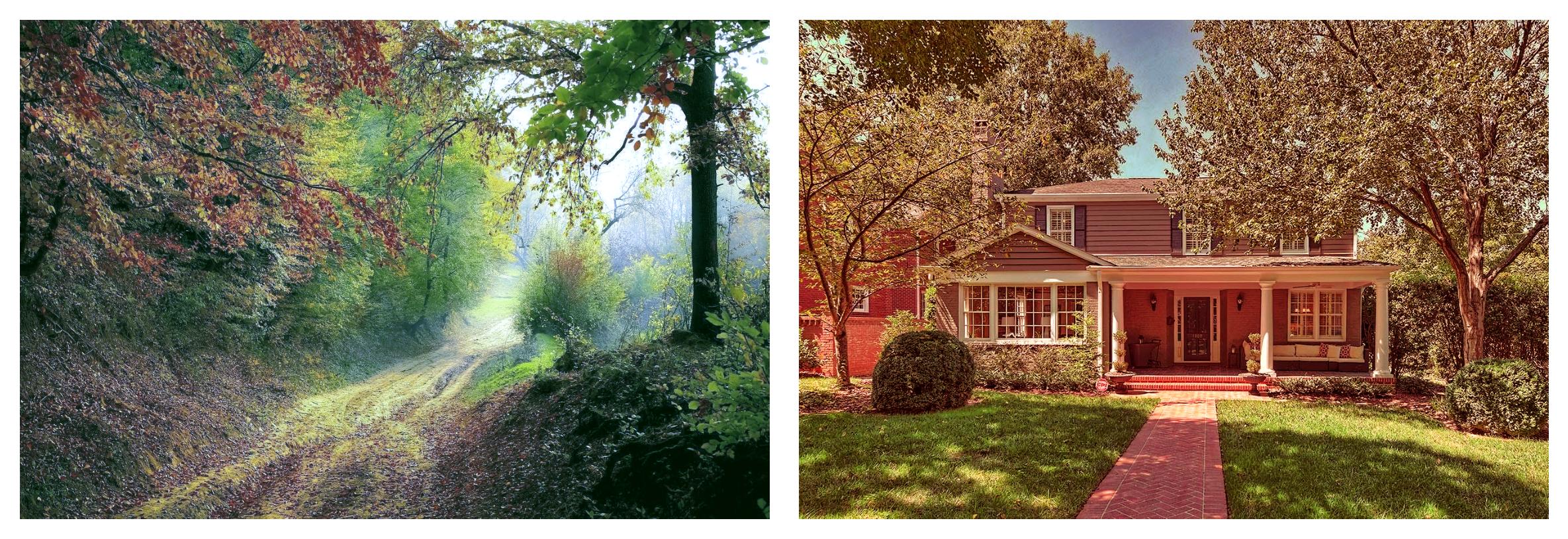}}
\subfloat[ICNN+W2L\eqref{W2GN-loss}\label{color_icnnw2l_fost}]{ 
\includegraphics[width=0.45\columnwidth]{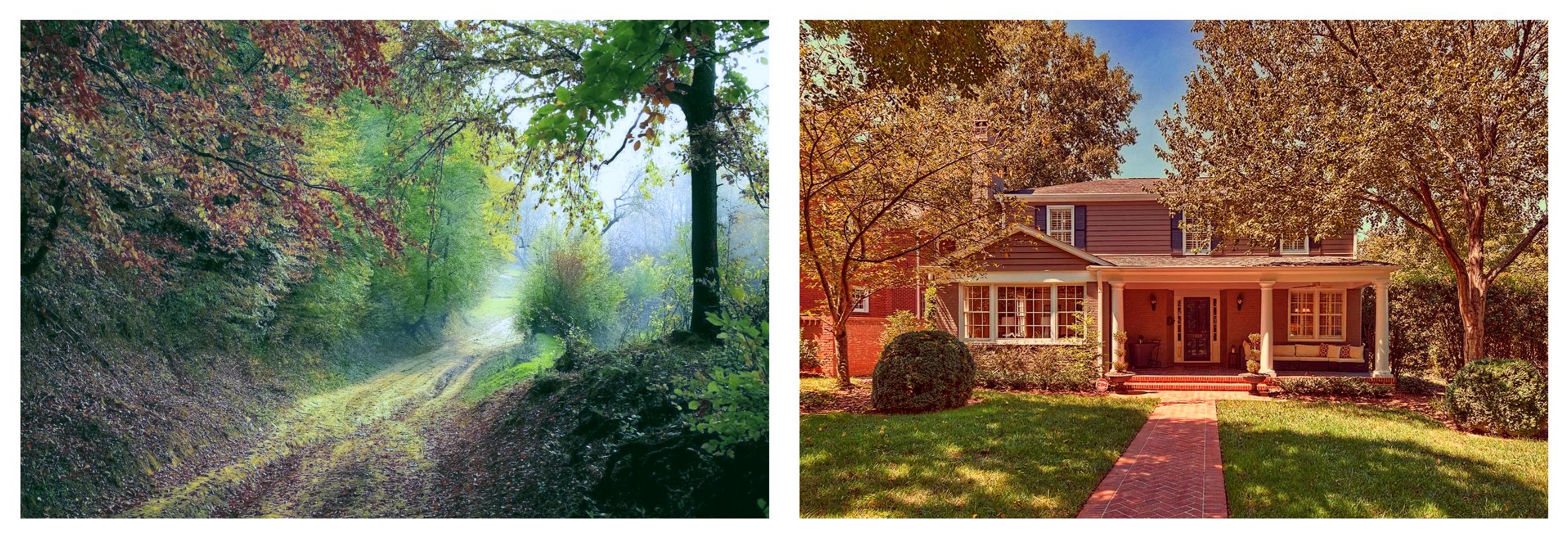}}
\caption{Comparison of color transferred results for the first task (The transferred sky color in Figure \ref{color_icnnw2l_fost} is incorrect.)}
\label{color_com1}
\end{figure}

An important application of OT in image processing is to find optimal color transfer plan between images. Here we illustrate how DPOT performs in the task.
The original images are presented in Figure \ref{ColorMLP0}.
The color on each pixels of the images can be viewed as discrete samples of the color space of the image. Then We can employ the loss functions \eqref{res1} and \eqref{res2} mentioned in Section \ref{sec:inv} to find the continuous OT map between color space of images, which provides the ``optimal" way to exchange the color. 
To illustrate the flexibility and superiority of our loss function, we conduct comparative experiments across different neural network architectures:  

We first apply a modified MLP trained with our loss $P(T_\theta)$ using regularization parameter $\lambda = 0.3$. 
As shown in Figure~\ref{color_fost_MLP}, the MLP successfully achieves accurate color transfer, demonstrating that our loss can guide learning without the need for restrictive architectural conditions.
\medskip

As a baseline, we also evaluate the ICNN architecture with the W2L loss~\cite{korotin2019wasserstein}, 
since it is analogous to our loss function:
\begin{align}
\textrm{W2L}=\min _{\theta,\omega}\bigg[\int_{\mathbb{R}^n} \psi_\theta(x) d \mu(x)+\frac{\eta}{2} R_{\mathcal{Y}}(\theta,\omega)
+\int_{\mathbb{R}^n}\left[\left\langle\nabla \psi_\omega^*(y), y\right\rangle-\psi_\theta\left(\nabla \psi_\omega^*(y)\right)\right] d \nu(y))\bigg],
\label{W2GN-loss}
\end{align}where
\begin{align}
\eta>0, \quad R_{\mathcal{Y}}=\int_{\mathbb{R}^n}\left\|\nabla \psi_\theta \circ \nabla \psi_\omega^*(y)-y\right\|^2 d \nu(y).  
\label{cycle-cons} 
\end{align} 
In \eqref{W2GN-loss} and \eqref{cycle-cons}, $\psi_\theta$ and $\psi_\omega^*$ are parametrized by two DenseICNN networks. Appendices \ref{appendix for NN} and \ref{appendix for ct} provide the descriptions of network structure and training details.

During the experiments, we found W2L takes least training time, as it does not require solving the discrete OT problems as required by DPOT. While the performance of W2L is highly task-dependent, due to the limitation of ICNN (required by W2L framework). In addition, we also test with the setting ICNN$+P_{\lambda=0.3}(T_\theta)$ for completeness as shown in Figure~\ref{color_icnnlambN0_fost}.

\begin{figure}[htbp]
\centering
\subfloat[Original images\label{palette_boat_src}]{ 
\includegraphics[width=0.45\columnwidth]{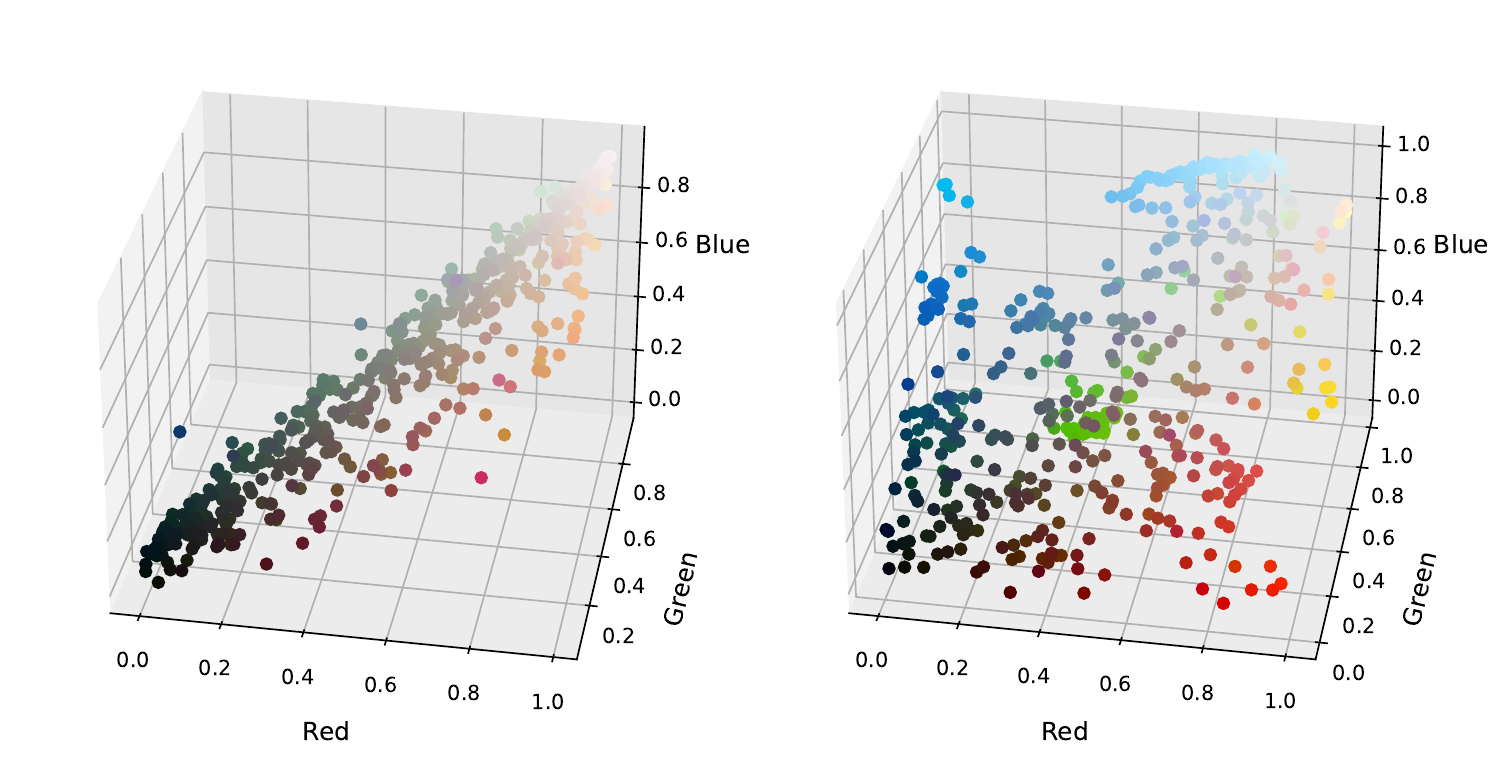}}
%\vspace{-1em}
\subfloat[MLP+$P_{\lambda=0.3}(T_\theta)$\label{palette_boat_MLP}]{ 
\includegraphics[width=0.45\columnwidth]{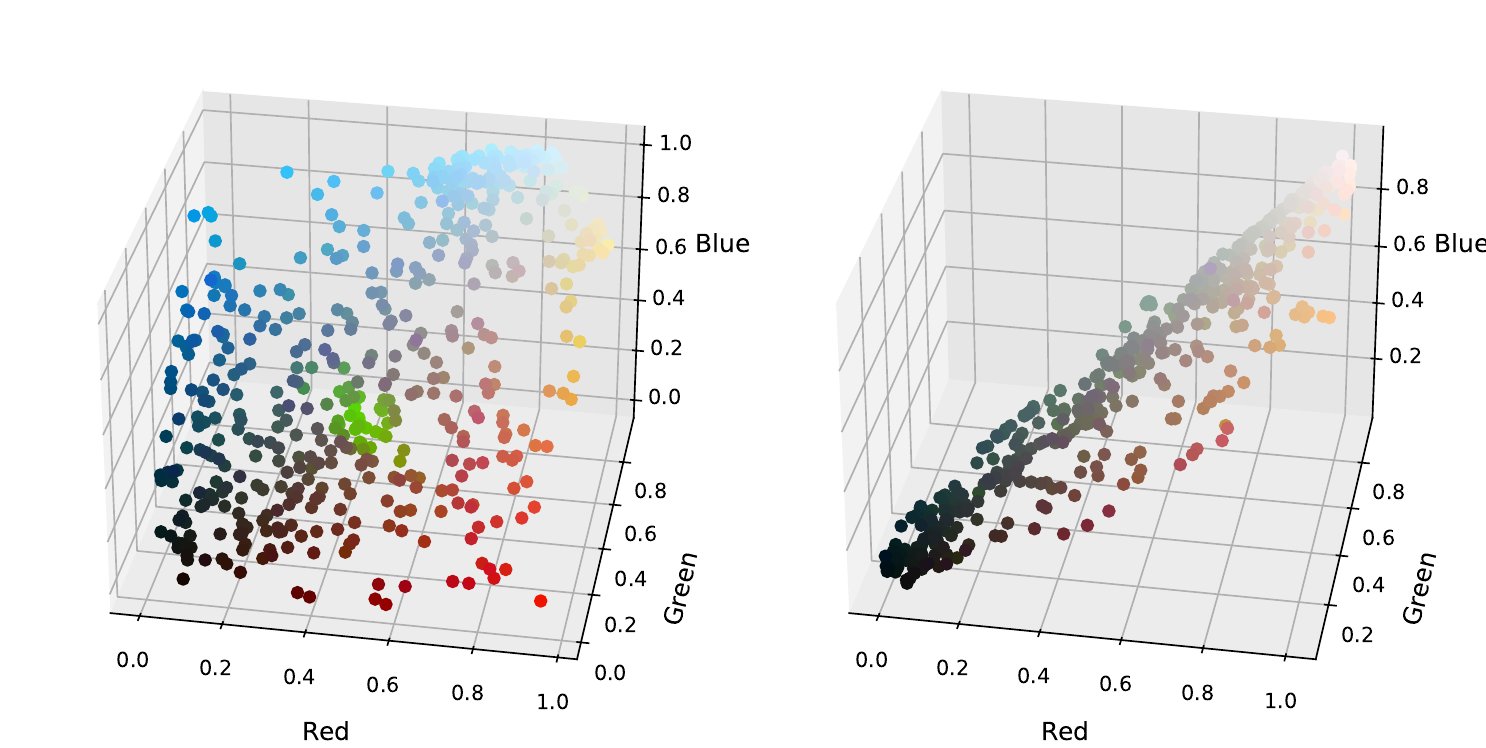}}

\subfloat[ICNN+$P_{\lambda=0.3}(T_\theta)$\label{palette_icnnlambN0_boat}]{ 
\includegraphics[width=0.45\columnwidth]{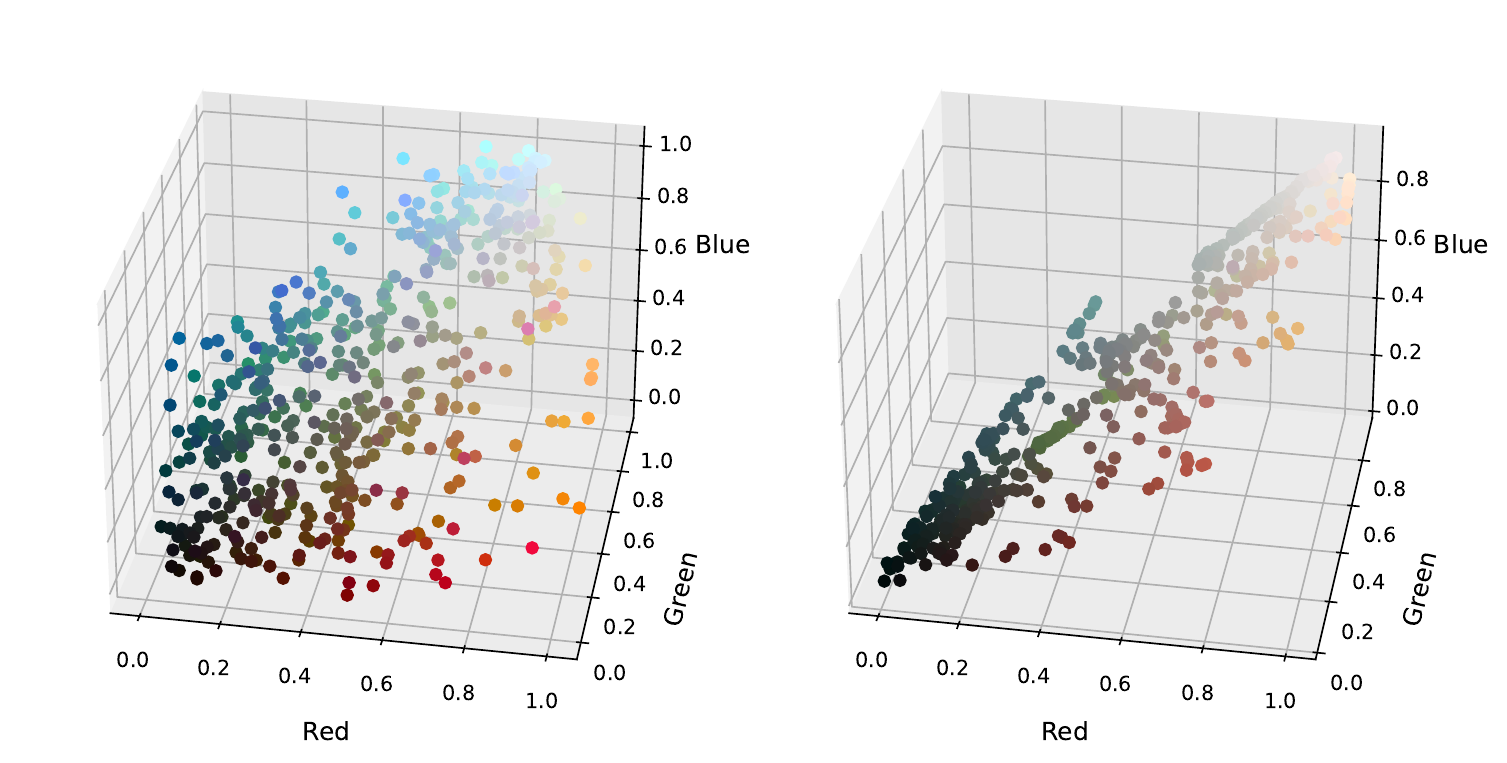}}
%\vspace{-1em}
\subfloat[ICNN+W2L\eqref{W2GN-loss}\label{palette_icnnw2l_boat}]{
\includegraphics[width=0.45\columnwidth]{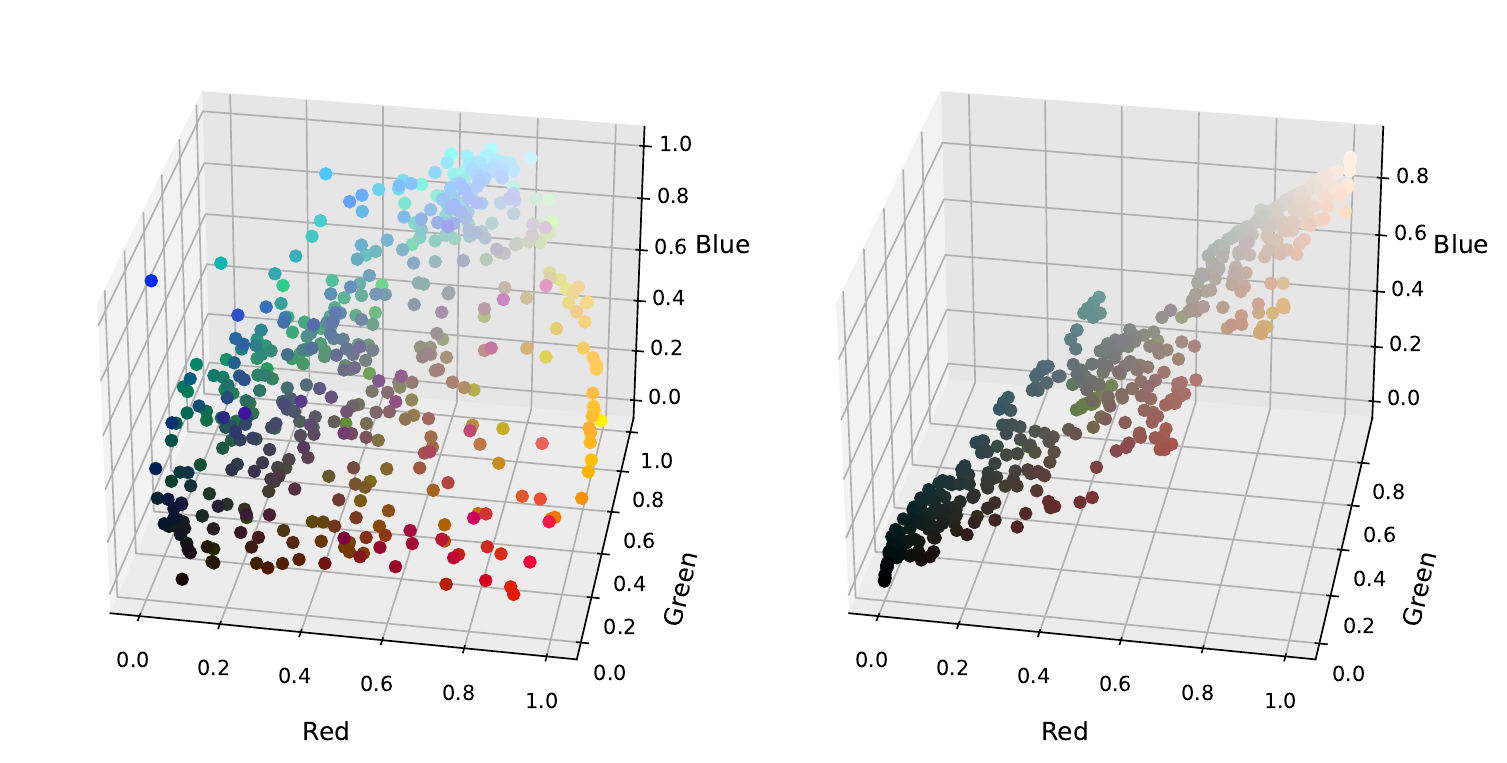}}
\caption{Comparison of transferred color palettes (500 random pixels) for the second task (Left: corresponds to the left column of Figure~\ref{color_com2}. Right: corresponds to the right column of Figure~\ref{color_com2}.)}
\label{colorpalettes_com}
\end{figure}

Moreover, in Figure~\ref{colorpalettes_com}, we compare the RGB scatter of the various settings for another set of experiment, the original images of which are provided in Figure \ref{ColorMLP0_boat}. The MLP$+P_{\lambda=0.3}(T_\theta)$ mapping produces clusters that are closer to the original images, see the green dots in the center of second picture of Figure~\ref{palette_boat_src} only recur in the center of first picture of Figure~\ref{palette_boat_MLP}.
Also only the second picture of Figure~\ref{palette_boat_MLP} has the similar structure of the first picture of Figure~\ref{palette_boat_src} (blue part), while those in Figure~\ref{palette_icnnlambN0_boat} and Figure~\ref{palette_icnnw2l_boat} are not in the structure like a straight line. 

Summing up, both W2L loss and our loss $P(\cdot)$ can fulfill the color transfer task roughly. While the ICNN, which is required by the W2L \cite{korotin2019wasserstein}  framework, introduce difficulties during training and
 MLP$+P_{\lambda=0.3}(T_\theta)$ yields the best accuracy compared to other situations.

Besides, for the purpose of illustrating the exchange process and potential application, we present the displacement interpolation \cite{mccann1997convexity}, $\tilde{T}_t(x)=(1-t)x+tT(x)$, in Figures \ref{color_process1} and \ref{color_process2}.

\section{Conclusion and future work}\label{sec 5}
This work proposes an end-to-end min-min framework to learn continuous OT maps by neural network. We introduce a loss function combining the Monge problem formulation with the Wasserstein-2 distance from $T_{\sharp}\mu$ to $\nu$. We proved that our learned map $T_\theta$ converges weakly to the OT map $\bar{T}$, and further established a quantitative error bound. Specifically, we showed that the $L^2$-distance between $T_\theta$ and $\bar{T}$ is controlled by the duality gap. Numerical experiments across multiple architectures, including MLP, ResNet, ICNN, demonstrate the effectiveness and flexibility of the proposed loss function, validating the theoretical analysis and confirming the high-dimension scalability. The generalization of convergence theorem to stochastic flow through Schr{\"o}dinger bridge may be a further research direction.

\section*{Statements \& Declarations}
\paragraph{Data Availability Statement} A representative example implementation, including code and data is available at \href{Github}{https://github.com/liyingyuan/DPOT.git}. The complete dataset and code for all experiments are available upon reasonable request.

\paragraph{Author Contributions} All authors contributed to the study conception and design. 

\paragraph{Conflict of Interest} The authors declare that they have no known competing financial interests or personal relationships that could have appeared to influence the work reported in this paper.
 
\bibliographystyle{spmpsci} %
\bibliography{bib}

\appendix
\renewcommand{\thesection}{\Alph{section}} %
\renewcommand{\theequation}{\Alph{section}.\arabic{equation}}
\renewcommand{\thetheorem}{\Alph{section}.\arabic{theorem}}
\renewcommand{\thefigure}{\Alph{section}.\arabic{figure}}
\renewcommand{\thetable}{\Alph{section}.\arabic{table}}

\counterwithin{figure}{section}

\section{Supplementary results and proofs} \label{appendix:proof}
We first recall some crucial definitions to characterize convexity and differentiability properties of probability measures from \cite{villani2009optimal}.
\begin{definition}
 Let $\mathcal{X}, \mathcal{Y}$ be two sets, and $c: \mathcal{X} \times \mathcal{Y} \rightarrow$ $(-\infty,+\infty]$. A function $\psi: \mathcal{X} \rightarrow \mathbb{R} \cup\{+\infty\}$ is said to be $c$-convex if it is not identically $+\infty$, and there exists $\xi: \mathcal{Y} \rightarrow \mathbb{R} \cup\{ \pm \infty\}$ such that
\begin{align}\label{c-convex}
\forall x \in \mathcal{X}, \quad \psi(x)=\sup _{y \in \mathcal{Y}}(\xi(y)-c(x, y)).
\end{align}
Then its c-transform is the function $\psi^c$ defined by
\begin{align}\label{c-transform}
\forall y \in \mathcal{Y}, \quad \psi^c(y)=\inf _{x \in \mathcal{X}}(\psi(x)+c(x, y)),
\end{align}
and its c-subdifferential is the c-cyclically monotone set defined by
\begin{align*}
\partial_c \psi:=\left\{(x, y) \in \mathcal{X} \times \mathcal{Y} ; \quad \psi^c(y)-\psi(x)=c(x, y)\right\}.
\end{align*}
The functions $\psi$ and $\psi^c$ are said to be c-conjugate.

Moreover, the c-subdifferential of $\psi$ at point $x$ is
\begin{align*}
\partial_c \psi(x)=\left\{y \in \mathcal{Y} ; \quad(x, y) \in \partial_c \psi\right\},
\end{align*}
or equivalently
\begin{align*}\label{c-subdifferential}
\forall z \in \mathcal{X}, \quad \psi(x)+c(x, y) \leq \psi(z)+c(z, y).
\end{align*}
\end{definition}

\begin{proposition}
\cite[Theorem 2.12]{villani2021topics}. Let $\mu, \nu$ be probability measures on $\mathbb{R}^n$, with finite second order moments, in the sense of \begin{equation*}
C_{\mu,\nu}:=\int_{\mathbb{R}^n} \frac{|x|^2}{2} d \mu(x)+\int_{\mathbb{R}^n} \frac{|y|^2}{2} d \nu(y)<+\infty .
\end{equation*} We consider the Monge-Kantorovich transportation problem associated with a quadratic cost function $c(x, y)=\frac{\|x-y\|^2}{2}$. Then,
\begin{itemize}
    \item [(i)] (Knott-Smith optimality criterion) $\gamma \in \Gamma(\mu, \nu)$ is optimal if and only if there exists a convex lower semi-continuous function $\psi$ such that 
    \begin{align*}
    \operatorname{Supp}(\gamma) \subset \operatorname{Graph}(\partial \psi),
    \end{align*}
    or equivalently: 
    \begin{align*}
    \text{for} \;\ud\gamma\text{-almost all} \; (x, y), \quad y \in \partial \psi(x).
       \end{align*}
      
Moreover, in that case, the pair $\left(\psi, \psi^*\right)$ has to be a minimizer in the problem \eqref{Kandual}.

\item[(ii)] (Brenier's theorem) If $\mu$ admits a density with respect to the Lebesgue measure, then there is a unique optimal $\gamma$, which is
\begin{equation*}
d \gamma(x, y)=d \mu(x) \delta[y=\nabla \psi(x)],
\end{equation*}
or equivalently,
\begin{equation*}
\gamma=(\mathrm{Id} \times \nabla \psi)_\sharp \mu,
\end{equation*}
where $\nabla \psi$ is the unique (i.e. uniquely determined $d \mu$-almost everywhere) gradient of a convex function which pushes $\mu$ forward to $\nu$ : $\nabla \psi_\sharp \mu=\nu$. Moreover,
\begin{equation*}
\operatorname{Supp}(\nu)=\overline{\nabla \psi(\operatorname{Supp}(\mu))}.
\end{equation*}
\item[(iii)] As a corollary, under the assumption of (ii), $\nabla \psi$ is the unique solution to the Monge problem \eqref{monge-T0}.

\item[(iv)] Finally, if $\nu$ admits a density with respect to the Lebesgue measure, then, for $\ud\mu$-almost all $x$ and $\ud \nu$-almost all $y$,
\begin{equation*}
\nabla \psi^* \circ \nabla \psi(x)=x, \quad \nabla \psi \circ \nabla \psi^*(y)=y,
\end{equation*}
and $\nabla \psi^*$ is the ($\ud \nu$-almost everywhere) unique gradient of a convex function which pushes $\nu$ forward to $\mu$, and also the solution of the Monge problem for transporting $\nu$ onto $\mu$ with a quadratic cost function.

\end{itemize}
\end{proposition}

\begin{proposition}
\cite[Theorem 7.8]{maggi2012sets} If $f: \mathbb{R}^n \rightarrow \mathbb{R}^m$ is a Lipschitz function and $x$ is a Lebesgue point of the weak gradient $\nabla f$, then $f$ is differentiable at $x$ (in particular, $f$ is differentiable a.e. on $\mathbb{R}^n$ ), with
\begin{align*}
\mathrm{d} \psi_x[\tau]=\nabla f(x)[\tau], \quad \forall \,\tau \in \mathbb{R}^n.
\end{align*}
\end{proposition}

\subsection{Perturbative results on OT maps}\label{sec:perturb}
Here we list some perturbative results on OT maps from which we develop our analysis in Section \ref{Alg sec}.

\begin{proposition}[Proposition 3.3 in \cite{gigli2011holder}]\label{prop3.3}
Let $\mu$ and $\nu_\epsilon$ be two distributions on $\mathbb{R}^n$ with finite transport cost and finite second order moments. Assume that $\supp(\mu)$ and $\supp(\nu_\epsilon)$ (i.e. the smallest closed sets on which $\mu$ and $\nu_\epsilon$ are concentrated) are both $C^2$ and uniformly convex.
Let $\varphi_\epsilon\in C^{2,\alpha}(\supp(\mu))$ be a smooth function whose gradient is the optimal transport map from $\nu_\epsilon$ to $\mu$, let $M > 0$ be the modulus of uniform convexity of $\varphi_\epsilon$ (i.e $M$ is the supremum of $M'$ such that $y\mapsto \varphi_\epsilon(y)-\frac{1}{2}M' |y|^2$ is convex on $\supp(\mu)$) and let $\bar{T}_{\epsilon}:=(\nabla\varphi_\epsilon)^{-1}$. Then for every transport map $T_\epsilon$ from $\mu$ to $\nu_\epsilon$ the following holds:
\begin{equation*}
    \| T_\epsilon-\bar{T}_{\epsilon}\|^2_{L^2(\mu)}\leq\frac{2}{M}(\| T_\epsilon-\mathrm{Id} \|^2_{L^2(\mu)}-\| \bar{T}_{\epsilon}-\mathrm{Id} \|^2_{L^2(\mu)}).
\end{equation*}
\end{proposition}

\begin{proposition}\cite[Proposition 10]{hutter2021minimax}\label{prop 10}
Fix two constants $K\geq2,\eta>1$.
Let $\mathcal{M}=\mathcal{M}(K)$ be the set of all probability measures $P$ whose support $\Omega_P\subseteq KB_1$ is a bounded and connected Lipschitz domain, and that admit a density $\rho_P$ with respect to the Lebesgue measure such that $K^{-1}\leq\rho_P(x)\leq K$ for almost all $x\in\Omega_P$. Assume that the measure $P\in\mathcal{M}$. 
For any $P\in\mathcal{M}$ with support $\Omega_P$, let $\tilde\Omega_P$ denote a convex set with Lipschitz boundary such that $\tilde\Omega_P\subseteq KB_1$, and $\Omega_P+K^{-1}B_1\subseteq\tilde\Omega_P$, where we denote by $B_1$ the unit-ball with respect to the Euclidean distance in $\mathbb{R}^n$. Let $\tilde{\mathcal{T}}=\tilde{\mathcal{T}}(K)$ be the set of all differentiable functions $T:\tilde\Omega_P\rightarrow \mathbb{R}^n$ such that $T=\nabla \psi$ for some differentiable convex function $\psi:\tilde\Omega_P\rightarrow\mathbb{R}^n$ and
\begin{itemize}
    \item[(i)] $|T(x)|\leq K$ for all $x\in\tilde\Omega_P$,

    \item[(ii)] $K^{-1}\preceq DT(x) \preceq K$ for all $x\in\tilde\Omega_P$.
\end{itemize}
Let $\mathcal{X}= \mathcal{X}(K)$ be the set of all twice continuously differentiable functions $\psi:\tilde{\Omega}_p\to\R$ such that
\begin{itemize}
    \item[(i)] $|\psi(x)|\leq 2K^2$ and $|\nabla \psi(x)|\leq K$ for all $x\in\tilde\Omega_P$,

    \item[(ii)] $K^{-1}\preceq D^2\psi(x) \preceq K$ for all $x\in\tilde\Omega_P$.
\end{itemize}
Then there exists a Kantorovich potential $\bar{\psi}\in\mathcal{X}(K)$ and a map $\bar{T}\in\tilde{\mathcal{T}}(K)$ such that $\bar{T}=\nabla \bar{\psi}$ is the exact optimal transport map from source $P$ to target $Q$.
And $\forall \psi\in\mathcal{X}(2K)$, we have 
\begin{equation}\label{result1}
    \frac{1}{8K}\|\nabla \psi(x)-\nabla \bar{\psi}(x)\|^2_{L^2(P)}\leq S(\psi)-S(\bar{\psi})\leq 2K\|\nabla \psi(x)-\nabla \bar{\psi}(x)\|^2_{L^2(P)}
\end{equation}
and
\begin{equation}\label{result2}
    \frac{1}{4K}\|\nabla \psi^*(y)-\nabla \bar{\psi}^*(y)\|^2_{L^2(Q)}\leq S(\psi)-S(\bar{\psi}).
\end{equation}
\end{proposition}

\subsection{Proof to Proposition \ref{quant bound 2}}\label{proof to quant bound 2}
Since the sequence ${\nu_\epsilon}$ is assumed to converge, Prohorov’s theorem \cite[Theorems 6.1, 6.2]{billingsley1999convergence} suggests that the family ${\nu_\epsilon}$ is tight. By \cite[Lemma 4.4]{villani2009optimal}, which states that the set of all transport plans between two tight families of probability measures is itself tight in the product space. Therefore, the corresponding OT plans ${\gamma_\epsilon}$, namely each coupling $\mu$ and $\nu_\epsilon$, form a tight family in $\mathbb{R}^n\times \mathbb{R}^n$.

By tightness and weak compactness, there exists a subsequence $\{\gamma_{\epsilon_n}\}$ that converges weakly to some measure $\gamma^{\prime} \in \Gamma(\mu, \nu)$. Because every $\gamma_{\epsilon_n}$ is an OT plan for the quadratic cost, it is concentrated on a $c$-cyclically monotone set~\cite{Ambrosio2003existence}. Consequently, for all integer $N \geq 1$, the product measure $\gamma_{\epsilon_n}^{\otimes N}$ is concentrated on a set $C(N) \subset (\mathbb{R}^n\times \mathbb{R}^n)^N$ defined by 
\begin{align*}
C(N)=\left\{((x_1, y_1), \ldots,(x_N, y_N)): \sum_{i=1}^N c\left(x_i, y_i\right) \leq \sum_{i=1}^N c(x_i, y_{i+1})\right\},
\end{align*}
where indices are cyclic, when $i=N$, $y_{N+1}= y_1$. The continuity of the transport cost $c$ implies that the function
\begin{align*}
F((x_1, y_1), \ldots,(x_N, y_N)):=\sum_{i=1}^N\left[c(x_i, y_i)-c(x_i, y_{i+1})\right]
\end{align*} is also continuous. Hence $C(N) = {F \leq 0}$ is a closed set. By weak convergence of  $\gamma_{\epsilon_n}^{\otimes N}$ to $\gamma'^{\otimes N}$ and concentration of $\gamma_{\epsilon_n}^{\otimes N}$ on $C(N)$, the Portmanteau theorem ensures $\gamma'^{\otimes N}$ is concentrated on $C(N)$. Let $\Gamma' = \supp(\gamma')$ denote the support of $\gamma'$. Then $\Gamma'^N = (\supp(\gamma'))^{\otimes N}=  \supp(\gamma'^{\otimes N}) \subset C(N)$, implying that $\Gamma'$ is $c$-cyclically monotone. 

By the Theorem 5.10 in \cite{villani2009optimal}, the support of an OT plan $\gamma$ is a $c$-cyclically monotone set, and this property also ensures the uniqueness of the OT plan. Thus, $\gamma' = \gamma$, where $\gamma$ is the unique OT plan between $\mu$ and $\nu$.

We now establish weak convergence of the maps $\bar{T}_\epsilon$ to $\bar{T}$. For any given $\xi>0$ and $\delta>0$, Lusin’s Theorem ensures the existence of a closed set $K_\xi \subset \mathbb{R}^n$ satisfying $\mu[K_\xi] > 1 - \xi$ and $\bar{T}$ is continuous when restricted to $K_\xi$. Define the set 
\begin{align*}
A_\delta=\{(x,y)\in K_\xi\times \mathbb{R}^n:|y-\bar{T}(x)|\geq \delta\},
\end{align*} 
which is closed in $\mathbb{R}^n \times \mathbb{R}^n$. Since the limit plan $\gamma$ is concentrated on the graph of $\bar{T}$, it holds that $\gamma[A_\delta] = 0$. Therefore, by lower semi-continuity of measures on closed sets, we obtain
\begin{align*}
0 = \gamma[A_\delta] &\geq \limsup_{\epsilon \to 0} \gamma_\epsilon[A_\delta].
\end{align*}
Noting that $\gamma_\epsilon = (\mathrm{Id}, \bar{T}_\epsilon)_\sharp \mu$, it follows that
\begin{align*}
\gamma_\epsilon[A_\delta]= \mu[\{x \in K_\xi \left| \,|\bar{T}_\epsilon(x) - \bar{T}(x)| \right. \geq \delta \}].
\end{align*}
Thus, 
\begin{align*}
\limsup_{\epsilon \rightarrow 0} \mu[\{x \in K_{\xi}\left| \,|\bar{T}_\epsilon(x)-\bar{T}(x)| \right. \geq \delta\}]=0.
\end{align*}
Now observe that
\begin{align*}
\mu[\{x \in \mathbb{R}^n\left|\, |\bar{T}_\epsilon(x)-\bar{T}(x)| \right.\geq \delta\}] \leq \mu[\{x \in K_{\xi}\left|\, |\bar{T}_\epsilon(x)-\bar{T}(x)| \right. \geq \delta\}]+\mu[\mathbb{R}^n \backslash K_{\xi}].
\end{align*}
By definition of $K_\xi$, we have $\mu[\mathbb{R}^n \setminus K_\xi] < \xi$, so the second term is less than $\xi$. Therefore,
\begin{align*}
\limsup _{\epsilon \rightarrow 0} \mu[\left\{x \in \mathbb{R}^n\left|\, |\bar{T}_\epsilon(x)-\bar{T}(x)\right| \geq \delta\right\}] \leq \xi.
\end{align*}
At last, due to the selection of $\xi > 0$ is arbitrary, letting $\xi \to 0$ yields
\begin{align*}
\mu[\left|\bar{T}_\epsilon-\bar{T}| \right.\geq \delta] \rightarrow 0,
\end{align*}
i.e., $\bar{T}_\epsilon \to \bar{T}$ in measure, as desired.

\section{Details of experiments} \label{appendix:training detail}
\subsection{Training strategy}\label{al strategy}
In order to train the parameterized transport map $T_\theta$ efficiently, we adopt a mini-batch strategy, where each training batch contains a relatively small number of samples. The reduced batch size permits the use of discrete OT solvers. This makes the computation tractable when solving the full-sample Monge OT problem would be impractical. 

In the unconditioned setting, the neural network is tasked with learning a single, fixed OT map between a given source distribution $\mu$ and target distribution $\nu$. 
As a result, training can be carried out using a standard single-batch formulation. 
As for conditioned setting, it involves learning a family of transport maps indexed by a conditioning variable $\kappa$. 
To handle this variability, we implement a multi-batch strategy. 
During training, batches are constructed from a collection of source–target pairs corresponding to different conditioning values $\kappa$. 
Each batch thus contains $N =3000$ samples from several pairs $(\mu_{\kappa_r}, \nu_{\kappa_r})$ for $r = 1, \dots, n_\kappa$, where each $\kappa_r \in \mathcal{O}$ represents a distinct physical or experimental configuration.

This multi-batch structure equips the model to learn $\kappa$-dependent transport maps $T_{\theta}(\cdot|\kappa)$ within a unified framework and generalize effectively across varying conditions. In addition, averaging the OT costs over $n_\kappa > 1$ batches helps reduce the variance in empirical OT approximations and mitigates the effect of batch-level stochastic noise in the second term of Equation~\eqref{trainingloss}. 

Crucially, our method decouples the computationally expensive training phase from the fast inference phase. 
Although training may involve higher computational cost due to repeated $\gamma_r$ updates, particularly in high-dimensional settings, this is a one-time trade-off cost, analogous to offline training \cite{luo2023optimal} in deep learning. 
Once trained, the model defines a continuous map $T_\theta(\cdot|\kappa)$ that can push forward new samples from $\mu_{\kappa}$ to $\nu_{\kappa}$ without solving OT again, even for previously unseen values of $\kappa$. 
Besides, the trained model can be applied to much larger inference datasets (up to 300 times larger than the training mini-batch) through a single-pass inference. This makes our method scalable to real-world or large-scale applications since new samples can be processed rapidly without re-running expensive OT solvers. 
Crucially, because $\kappa$ is treated as an explicit input to the neural network, the learned model can be applied not only to the $\kappa$-values sampled during training, but also any desired $\kappa$, thereby supporting generalization across a continuous family of transport problems. 

\subsection{Network architectures}\label{appendix for NN} 
Because our proposed loss functions \eqref{DPOT-loss} and \eqref{trainingloss} are insensitive to the choice of network architecture, we adapt the network design to suit each task, including fully connected multi-layer perceptron (MLP), modified MLP, ResNet and DenseICNN.
\paragraph{Modified MLP}
This modified MLP incorporates a unique blending mechanism between two independent parameter sets, resulting in a more flexible architecture. The weights $W$ for each layer are initialized using the Xavier (Glorot) initialization method:
\begin{align*}
W \sim \mathcal{N}\left(0, \frac{1}{\sqrt{\frac{n_{\textrm {in }}+n_{\textrm {out }}}{2}}}\right),
\end{align*}
where $n_{\textrm {in }}$ and $n_{\textrm{out }}$ are the input and output dimensions. 
And two additional independent sets of parameters $(U_1, b_1)$ and $(U_2, b_2)$ are initialized for the blending mechanism. At each hidden layer, the network computes two intermediate activations:
\begin{align*}
U=\sigma\left(X U_1+b_1\right), \quad V=\sigma\left(X U_2+b_2\right),
\end{align*}
where $\sigma$ denotes the activation function, chosen to be the ReLU in this setup. The final hidden layer activation combines these two intermediate activations using a dynamic blending scheme:
\begin{align*}
\text { Output }=Z \cdot U+(1-Z) \cdot V,
\end{align*}
where $Z=\sigma(W X+b)$ is the standard activation computed from the layer's weights $W$ and biases $b$.

\paragraph{ResNet}
We use a fully-connected ResNet architecture to model the transport map. Each residual block applies a skip connection over a sequence of fully connected layers with PReLU activations initialized with a $\mathrm{slope}=-0.25$ in the first layer.

\paragraph{DenseICNN}
Full architectural specifications for DenseICNN follow the descriptions in Appendix B of~\cite{korotin2019wasserstein}. All activation functions are set to CELU. 

\subsection{Mapping nonuniform to uniform on a square}\label{appendix for square}

The ResNet architecture employed in this example consists of four residual blocks, each containing five dense layers with $64$ neurons. 

We generate a training dataset using an accept-reject sampling procedure from the source density~\eqref{square_density}. Specifically, we draw $N_0=80000$ source samples, paired with $N_0= 80000$ uniformly distributed target samples. 
The model is trained for $\mathrm{steps}=1000$ iterations in total. 
And we renew both the training batch and the transport coupling matrix $\gamma$ every $n_\gamma=50$ iterations in order to reduce the computation cost. 
 
\begin{figure}[htbp]
\centering
\subfloat[$\lambda=0.0$]{\includegraphics[width=0.8\columnwidth]{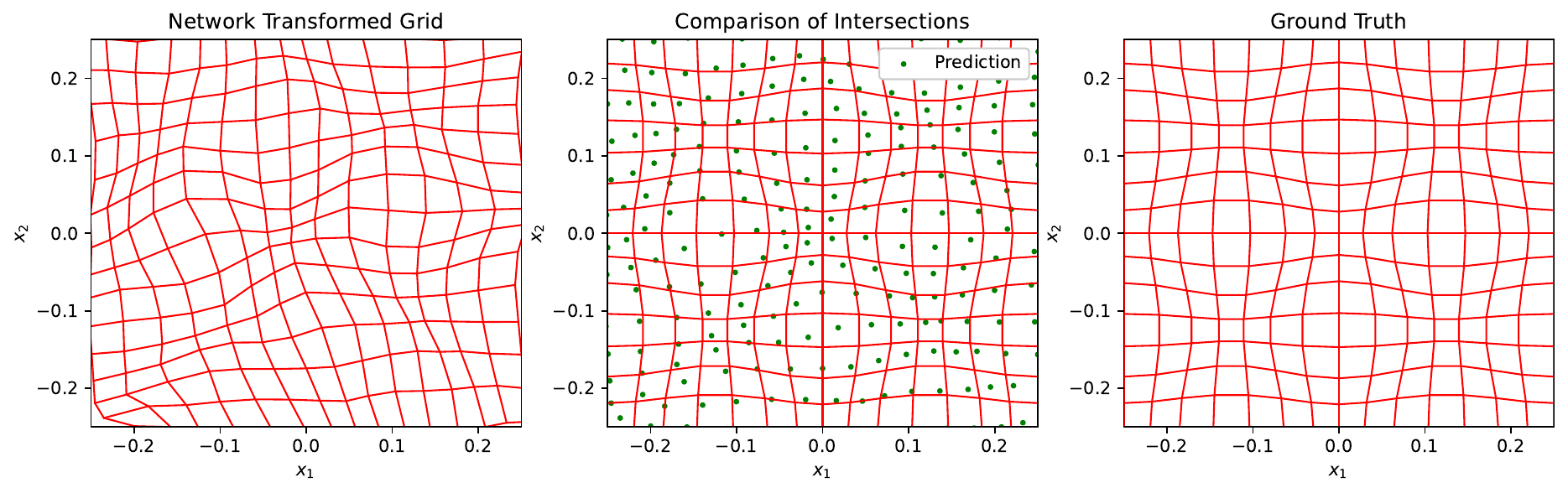}}

\subfloat[$\lambda=0.1$]{\includegraphics[width=0.8\columnwidth]{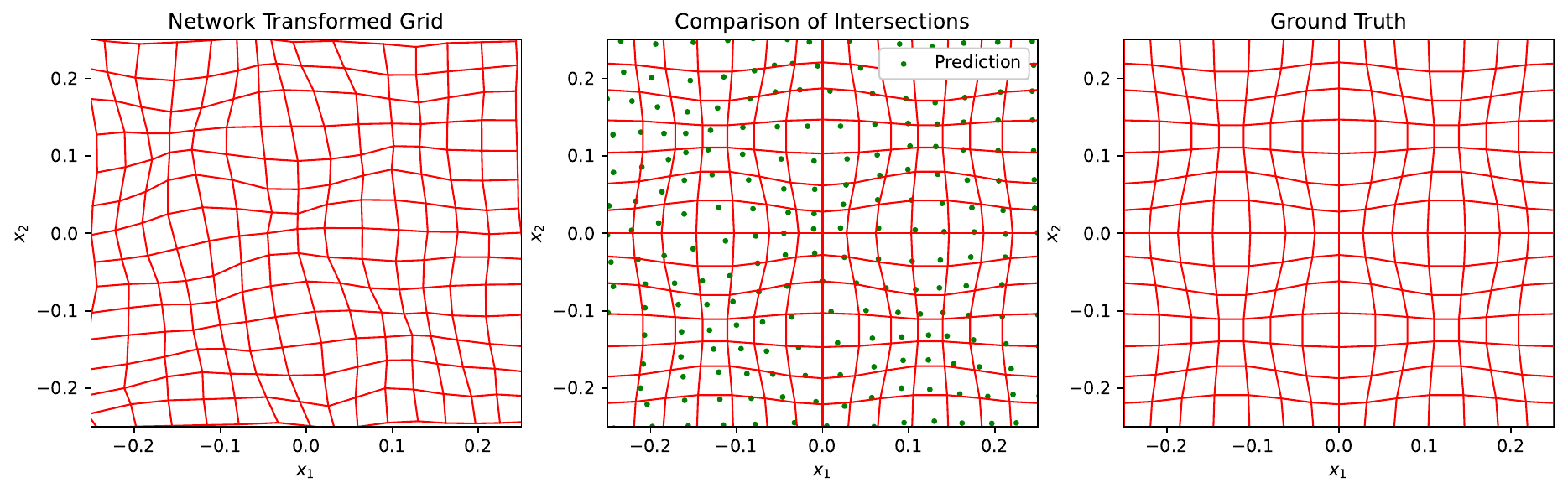}}

\subfloat[$\lambda=0.7$]{\includegraphics[width=0.8\columnwidth]{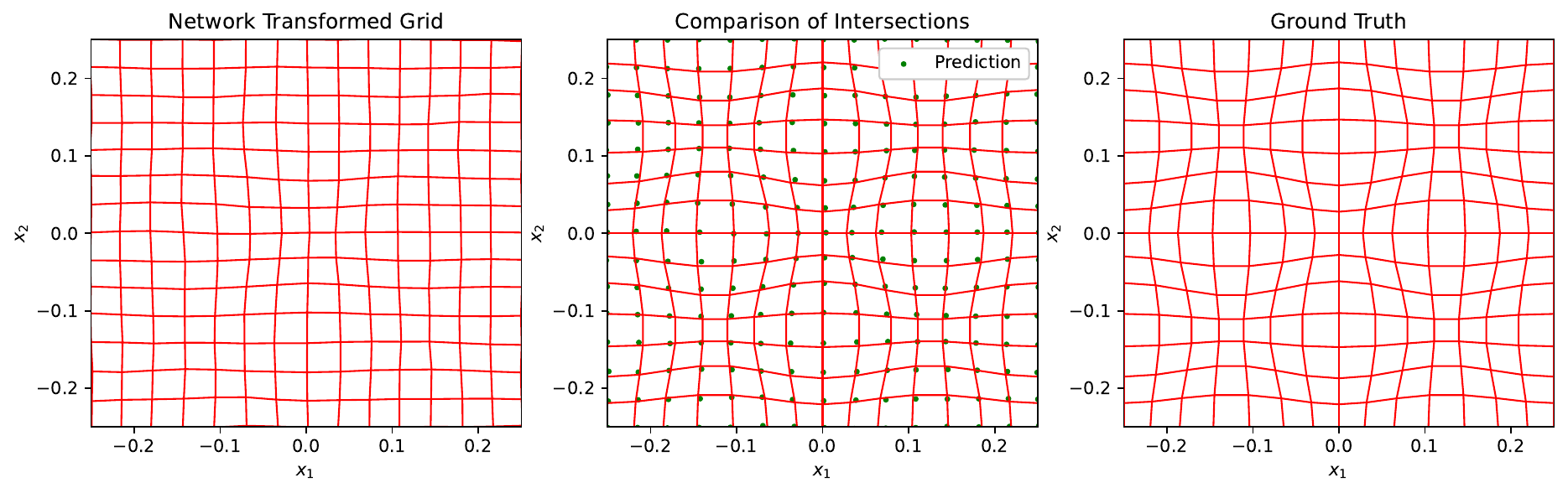}}

\subfloat[$\lambda=1.0$]{\includegraphics[width=0.8\columnwidth]{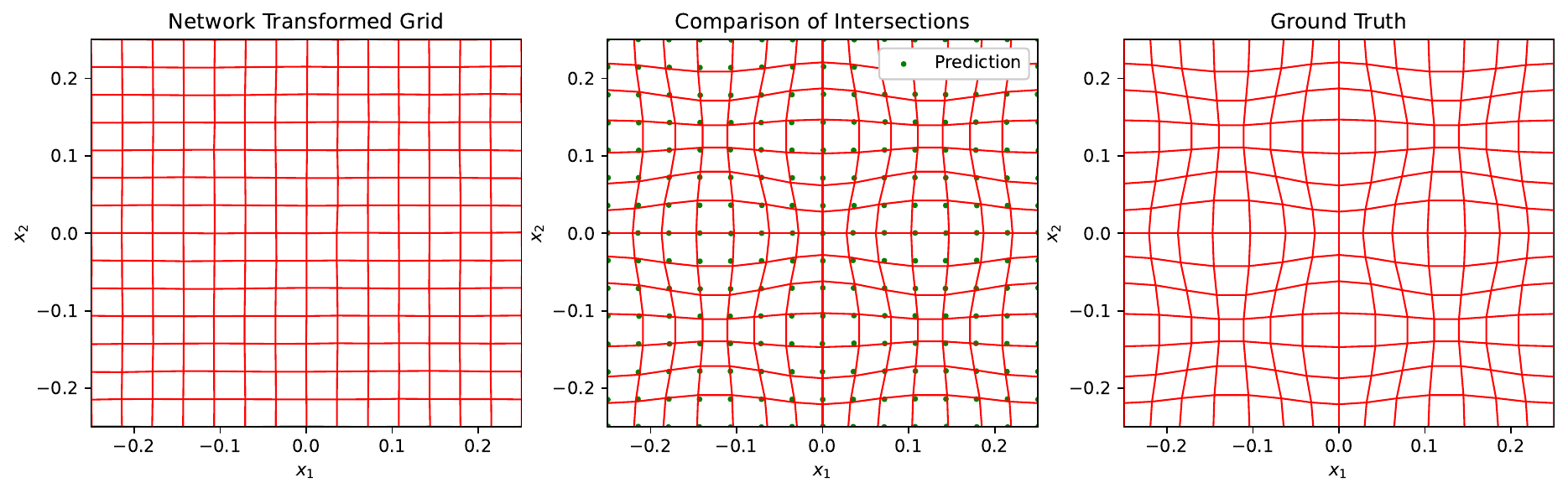}}
\caption{Comparison of the transformed Cartesian mesh under $T_\theta(x)$ and $\bar{T}(x)$ (Left: network transformed. Middle: green points are the network prediction of grid intersections. Right: ground truth.) }
\label{squ_mesh_com}
\end{figure}

\begin{figure}[htbp]
\centering
\subfloat[$\lambda=0.0$]{\includegraphics[width=0.4\columnwidth]{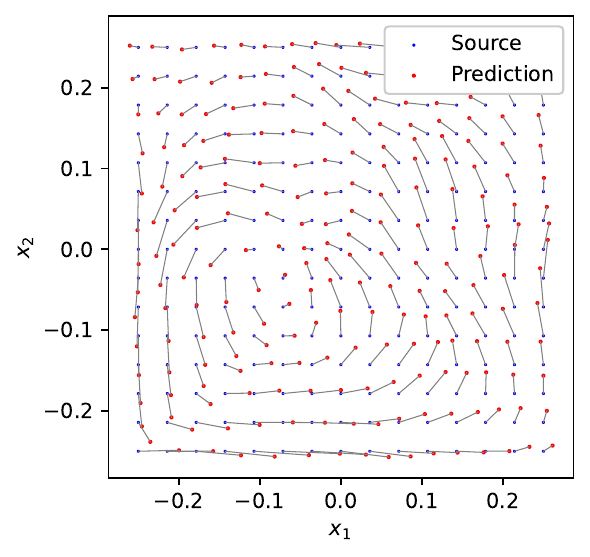}}
\subfloat[$\lambda=0.1$]{\includegraphics[width=0.4\columnwidth]{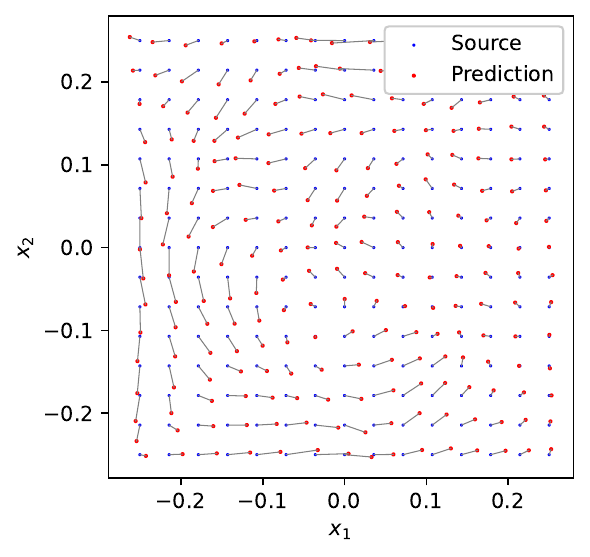}}

\subfloat[$\lambda=0.3$]
{\includegraphics[width=0.4\columnwidth]{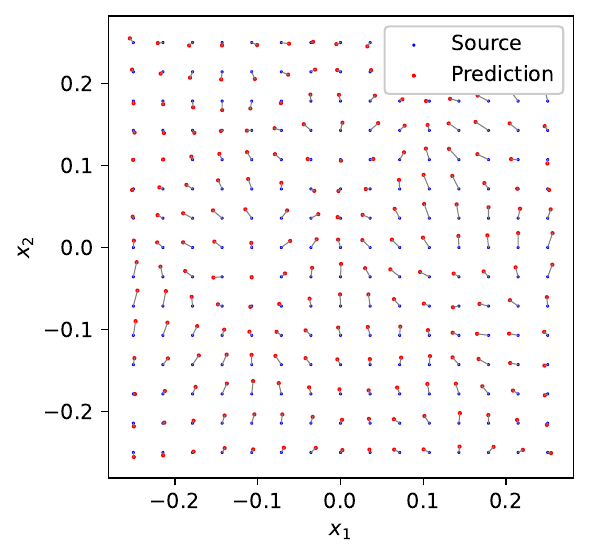}}
\subfloat[$\lambda=0.7$]{\includegraphics[width=0.4\columnwidth]{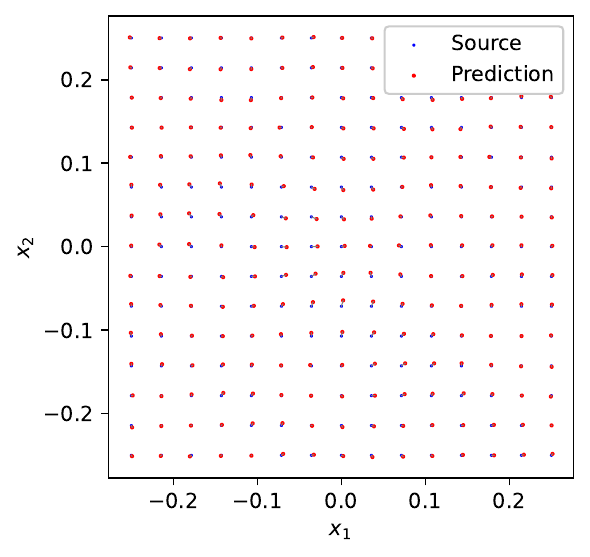}}

\subfloat[$\lambda=1.0$]{\includegraphics[width=0.4\columnwidth]{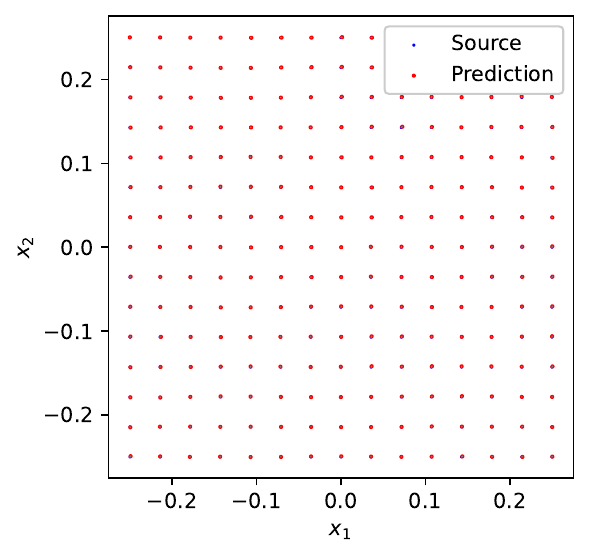}}
\subfloat[Ground truth map]{\includegraphics[width=0.4\columnwidth]{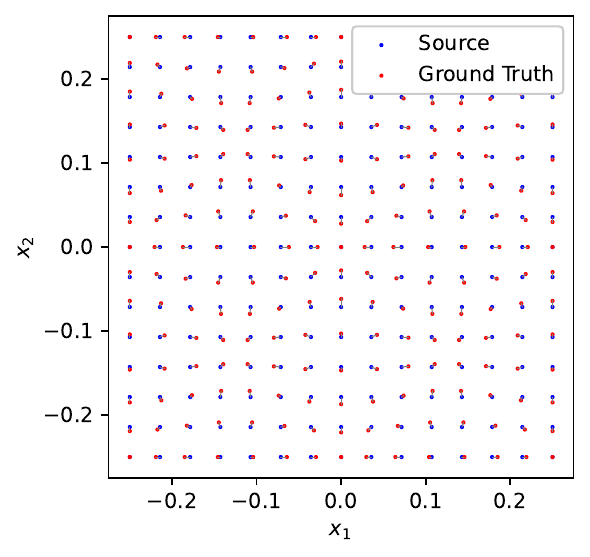}}
\caption{Comparison of the Cartesian grid pushed forward by $T_\theta$ and by $\bar{T}$ with different $\lambda$}
\label{squ_grid_map_com}
\end{figure}

Here we present a comparison of the Cartesian grid points pushed forward by $T_\theta$ and by $\bar{T}$ with different $\lambda$ in Figure \ref{squ_mesh_com}. As illustrated in Figure~\ref{squ_grid_map_com}, $\lambda = 0.3$ exhibits the best approximation for the ground truth map.  

\subsection{Mapping from one ellipse to another ellipse}\label{appendix for ep}

Both the unconditioned network $T_\theta$ and the conditioned one $T_\theta(\cdot|\kappa)$ are implemented using a ResNet architecture comprising three residual blocks. Each block includes four dense layers with $32$ neurons per layer. 

\begin{figure}[htbp]
\centering
\subfloat[$n_\gamma=5$\label{ellispe_ngamma5}]{
\includegraphics[width=0.4\columnwidth]{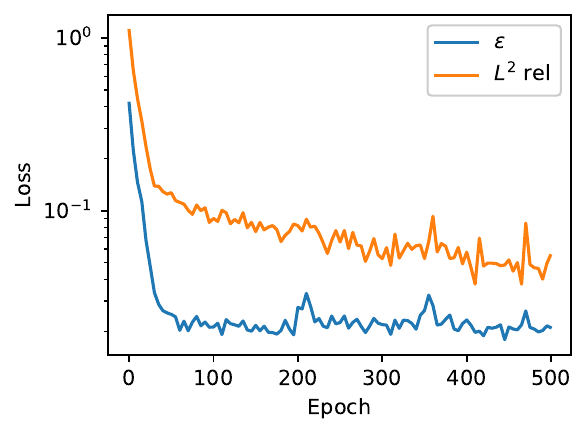}}
\subfloat[$n_\gamma=10$\label{ellispe_ngamma10}]{
\includegraphics[width=0.4\columnwidth]{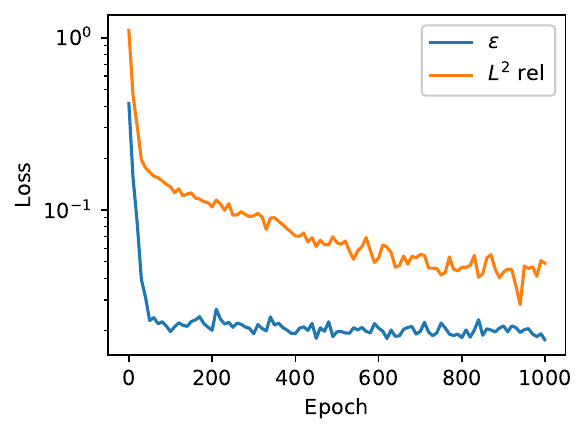}}

\subfloat[$n_\gamma=50$\label{ellispe_ngamma50}]{
\includegraphics[width=0.4\columnwidth]{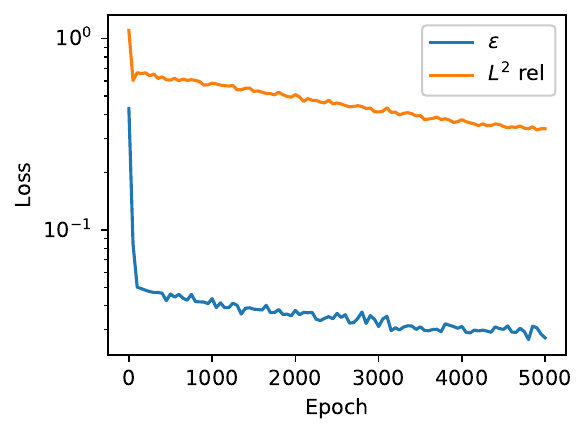}}
\subfloat[$n_\gamma=100$\label{ellispe_ngamma100}]{
\includegraphics[width=0.4\columnwidth]{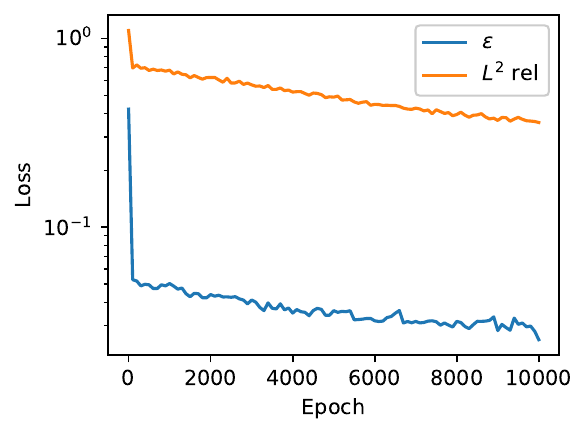}}
\caption{Total optimality gap $\epsilon$ and $L^2$ relative error vs. epochs for ellipse transport $T_\theta$ with varying $\gamma$ update frequencies $n_\gamma$ }
\label{ngamma_convergence}
\end{figure}

We generate a dataset containing $n_\kappa=10$ sets of $N_0=10000$ paired source and target samples. Figure \ref{ngamma_convergence} presents the convergence history of $T_\theta$ for different $n_\gamma$. 
To assess the influence of the update frequency $n_\gamma$, we perform a controlled comparison under a fixed runtime budget. Specifically, we set the total number of training $\mathrm{steps} = n_\gamma \times 100$, so that the number of OT plan updates remains the same across different $n_\gamma$. 
As the optimality gap $\epsilon$ in~\eqref{total_epsilon} is only well-defined at iterations where $\gamma$ is updated, we record $\epsilon$ every $n_\gamma$ steps. Notably, increasing the update frequency from $n_\gamma=5$ to $n_\gamma=10$ results in only a marginal increase in total runtime ($8.59\times 10^2$ seconds vs. $8.91\times 10^2$ seconds), confirming computational cost is dominated by the number of OT problem solves in \eqref{trainingloss} rather than neural network training.

We train for $\mathrm{steps}= 500$ iterations in total and set $n_\gamma=10$ for the conditioned case. 

\begin{figure}[htbp]
\centering
\includegraphics[width=0.45\linewidth]{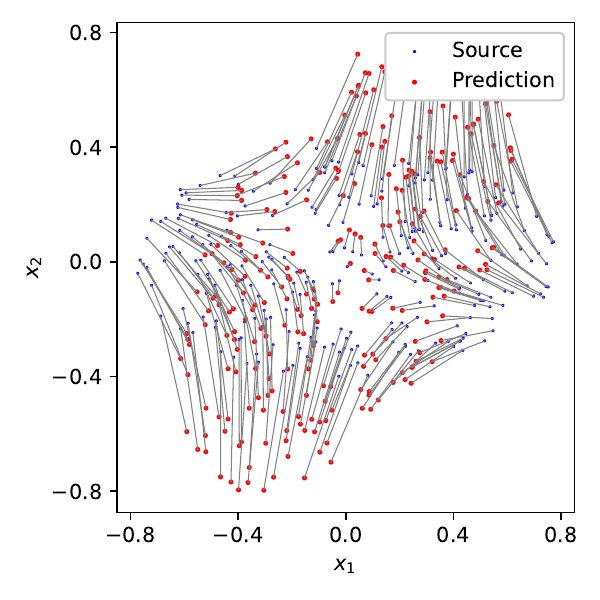}
\caption{Predicted transport map in the ellipse problem}\label{ellispe_uncond_map_com}
\end{figure}

\begin{figure}[hbpt]
\centering
\subfloat[ $\kappa=-0.15$]{ 
\includegraphics[width=0.25\columnwidth]{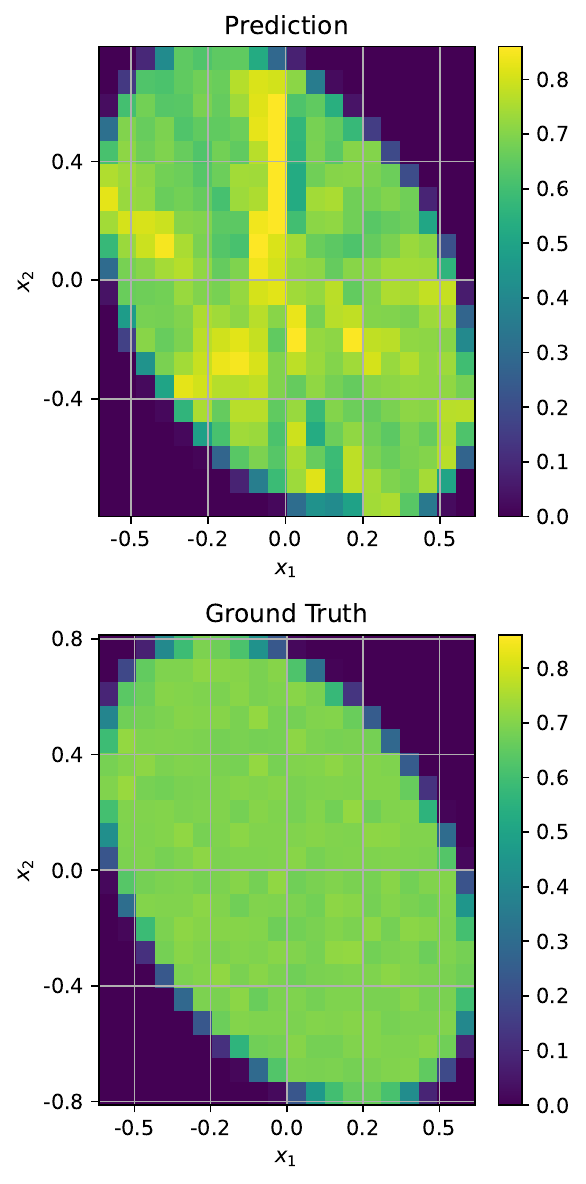}}
\subfloat[$\kappa=-0.47$]
{\includegraphics[width=0.25\columnwidth]{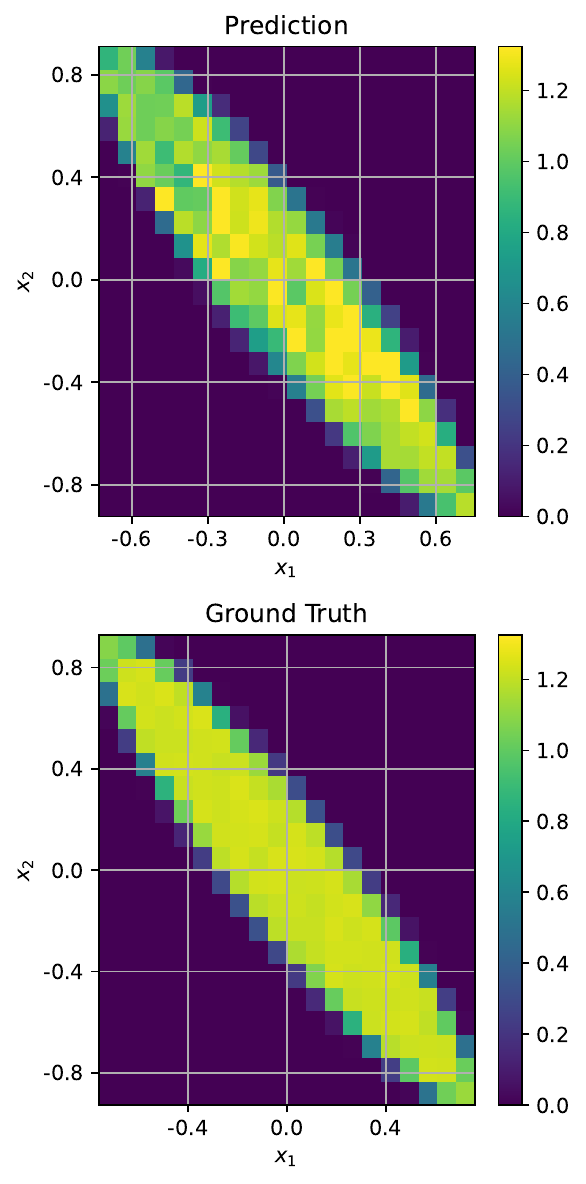}}
\subfloat[$\kappa=0.11$]{ 
\includegraphics[width=0.25\columnwidth]{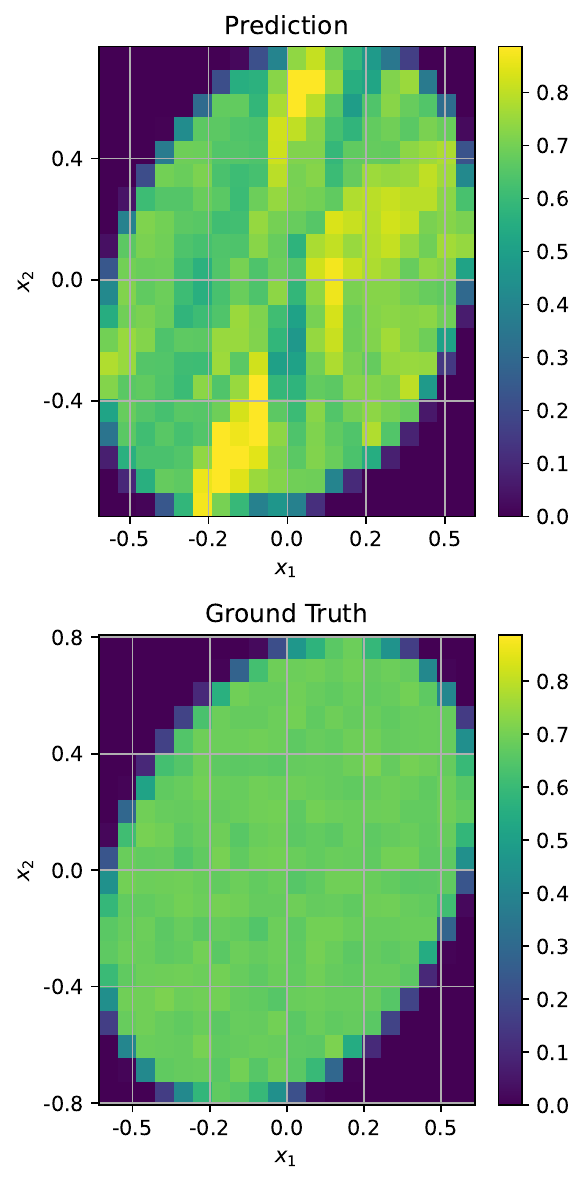}}
\subfloat[ $\kappa=0.32$]{ 
\includegraphics[width=0.25\columnwidth]{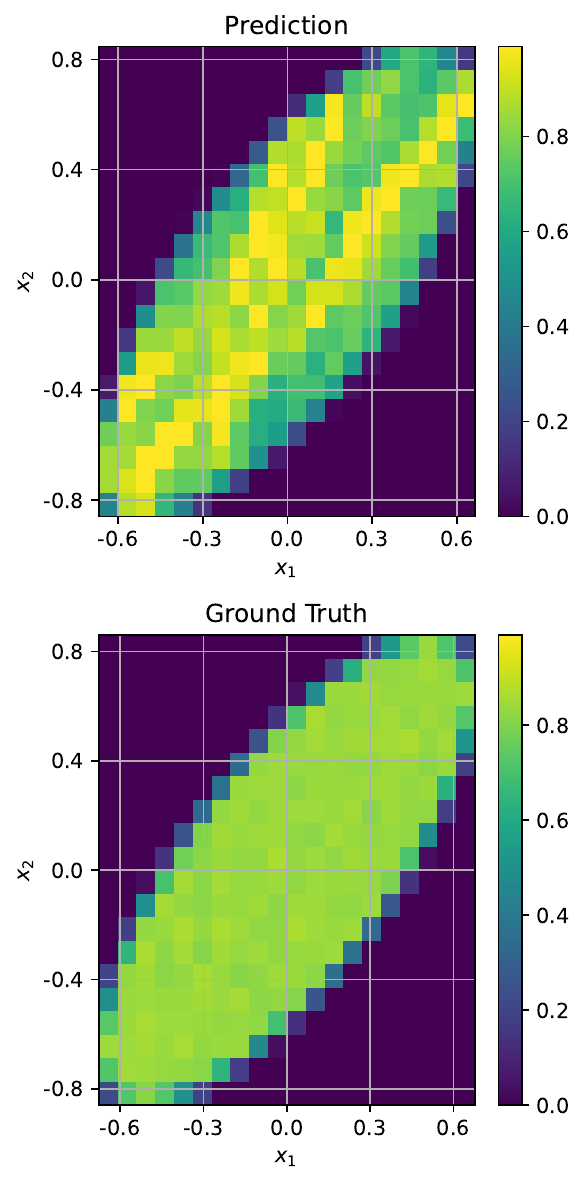}}
\caption{Histogram comparison of prediction inferred by conditioned network $T_\theta(\cdot|\kappa)$ with specific $\kappa$ and ground truth}
\label{ep_cond_hist}
\end{figure}

Figure~\ref{ellispe_uncond_map_com} illustrates the predicted transport map by the unconditioned $T_\theta$. Figure~\ref{ep_cond_hist} presents the histogram comparisons for each conditioned $\kappa$. Across all sampled values, the predicted transports $T_{\theta}(x|\kappa)$ consistently align with the shape and orientation of the target distributions $\nu_\kappa$. 

\subsection{Mapping disjoint two half circles onto the circle } \label{appendix for dis} 
The standard MLPs applied here both have a depth of three hidden layers, each containing $128$ neurons. 

\begin{figure}[htbp]
\centering
\includegraphics[width=0.5\columnwidth]{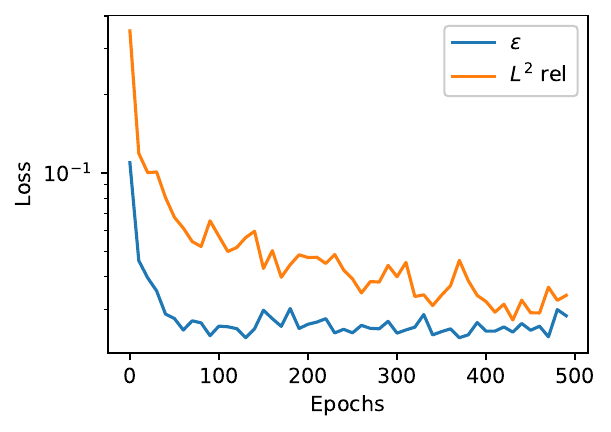}
\caption{Total optimality gap $\epsilon$ and $L^2$ relative error vs. epochs for unconditioned discontinuous transport $T_\theta$} 
\label{dis_uncond_loss}
\end{figure}

\begin{figure}[htbp]
\centering
\subfloat[$\kappa=0.2$]{ 
\includegraphics[width=0.4\columnwidth]{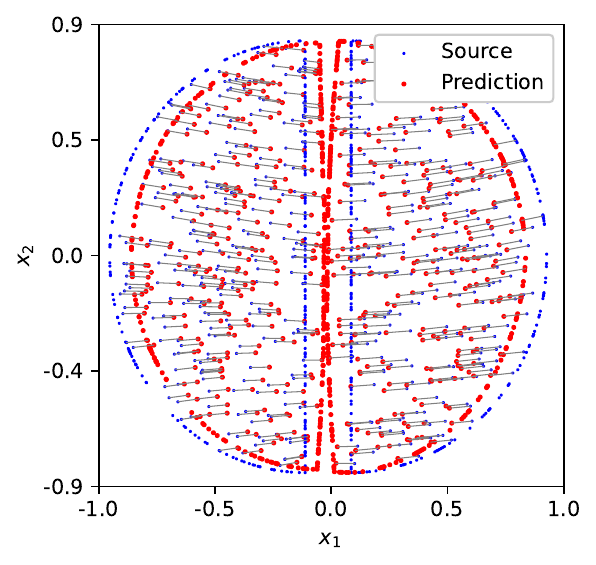}}
\subfloat[ $\kappa=0.4$]{ 
\includegraphics[width=0.4\columnwidth]{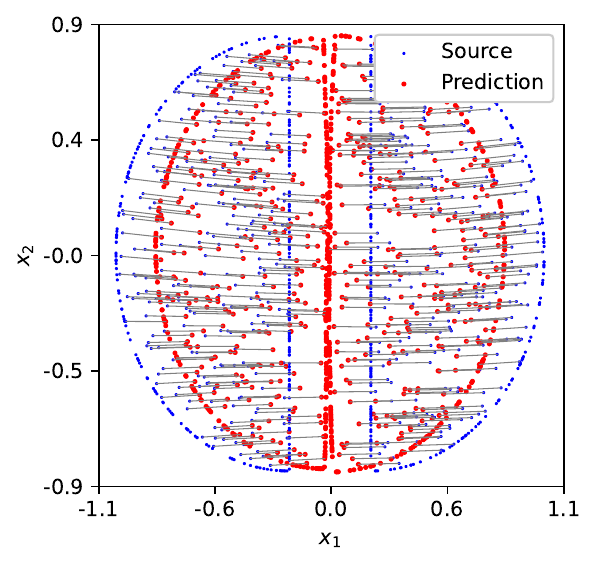}}

\subfloat[$\kappa=0.6$]{ 
\includegraphics[width=0.4\columnwidth]{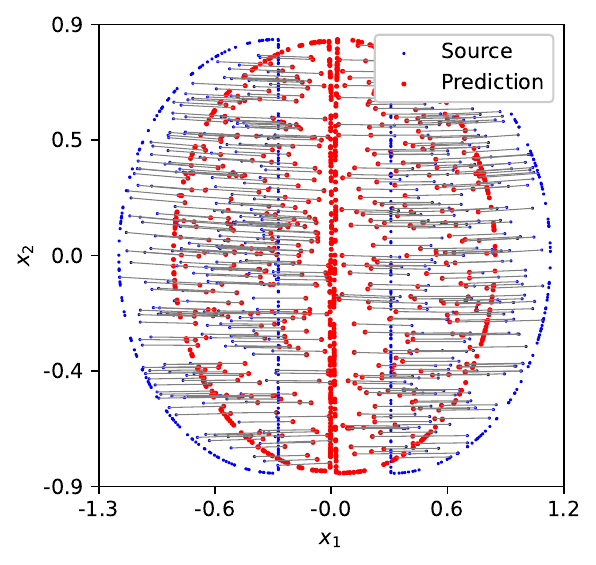}}
\subfloat[$\kappa=0.8$]{ 
\includegraphics[width=0.4\columnwidth]{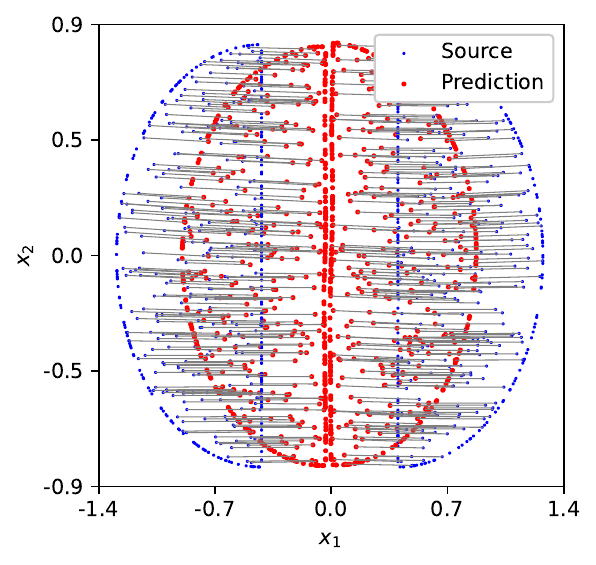}}
\caption{Predicted transport map by conditioned network $T_\theta(\cdot|\kappa)$ with specific $\kappa$ } 
\label{dis_multcond_map}
\end{figure}

For both unconditioned and multi-parameters conditioned settings, we generate $N_0=20000$ uniformly distributed source samples along with the target samples.
And the training takes $\mathrm{steps}= 500$ iterations in total, updating batch and transport plan $\gamma$ every $n_\gamma=10$ iterations.
\begin{figure}[htbp]
\centering
\includegraphics[width=0.45\linewidth]{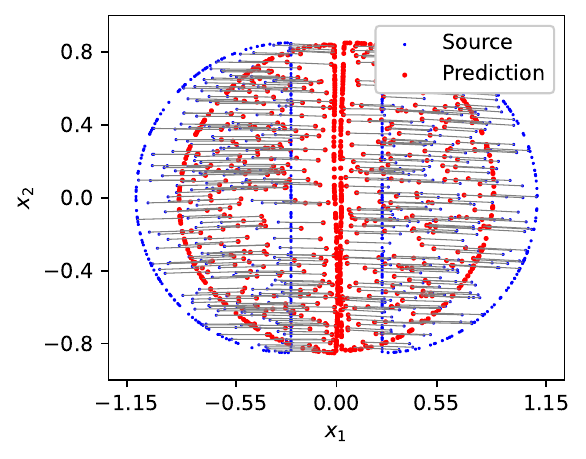}
\caption{Predicted transport map in disjoint support problem}\label{dis_uncond_map}
\end{figure}
Figure~\ref{dis_uncond_map} presents learned map with additional boundary points for clarity, confirming that this unconditioned $T_\theta$ effectively handles this singular transport where GAN-based methods often struggle with convergence and mode collapse \cite{lei2019mode}. Besides, Figure \ref{dis_uncond_loss} reports the convergence behavior under this unconditioned setting, which also verifies our theoretical analysis. Figure~\ref{dis_multcond_map} confirms that our conditioned loss \eqref{condition-DPOT-loss} is valid for this singular transport problem with $\kappa \in \mathcal{O}$. 

\subsection{Inverse mapping} \label{Appdendix for inv}
The source density $\rho_X(x)$ reads, with $x=(x_1,x_2)$,
\begin{equation}
\rho_X\left(x_1, x_2\right)=2+\sum_{\left(c_x, c_y\right) \in\{-1,1\} \times\{-1,1\}} \frac{1}{0.2^2} \exp \left(-\frac{0.5\left(\left(x_1-c_x\right)^2+\left(x_2-c_y\right)^2\right)}{0.2^2}\right),
\label{eqn:rhox_inverse}
\end{equation}
which includes a constant offset $2$, and $(c_x, c_y)$ stand for the $4$ corners of the domain $[-1,1]^2$. Generating $\rho_X$ is challenging due to this constant offset, as there is no direct sampler. 
The target density $\rho_Y({y})$ is a simple gaussian in the center of the domain $[-1,1]^2$:
\begin{align*}
 \rho_Y(y_1,y_2)=2+\frac{1}{0.2^2} \exp (-\frac{0.5|y_1^2 + y_2^2|}{0.2^2}).   
\end{align*} The source density $\rho_X({x})$ is constructed on $[-1,1]^2$ as \eqref{eqn:rhox_inverse}, which is symmetric with respect to $x_1$ and $x_2$ axis. We use the accept-reject algorithm to generate source and target samples.
The generated dataset contains $n_\kappa=10$ sets of source-target pairs, with each set containing $N_0 = 20000$ samples.

\begin{figure}[htbp]
\centering
\subfloat[Forward network output (left) vs. reference (right)]{
\includegraphics[width=0.65\columnwidth]{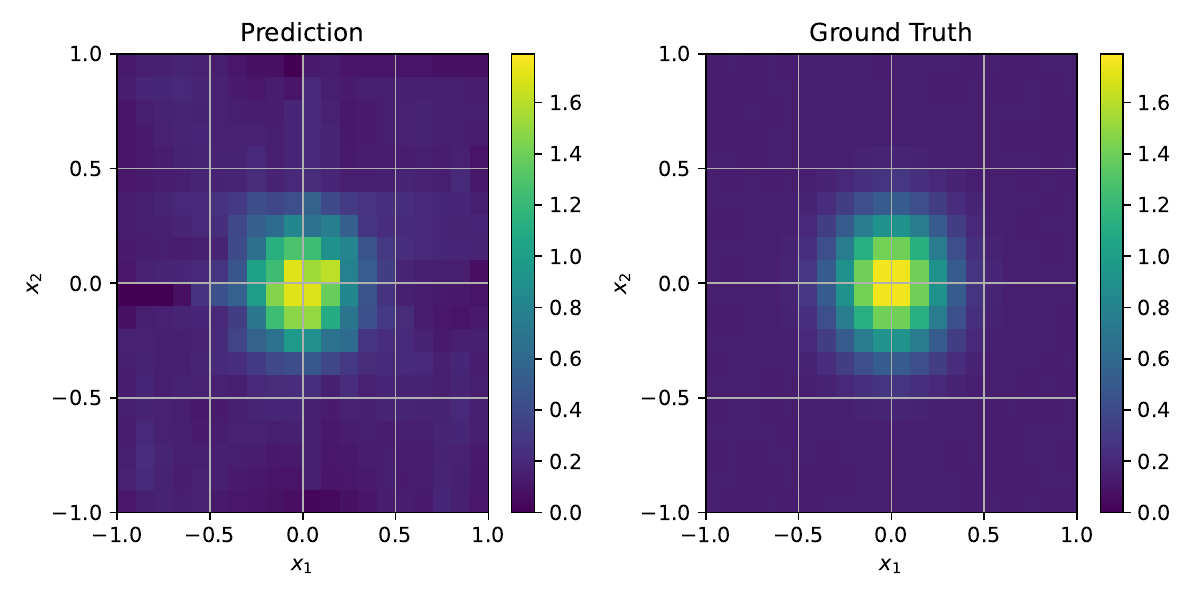}}

\subfloat[Inverse network output (left) vs. reference (right)]{ 
\includegraphics[width=0.65\columnwidth]{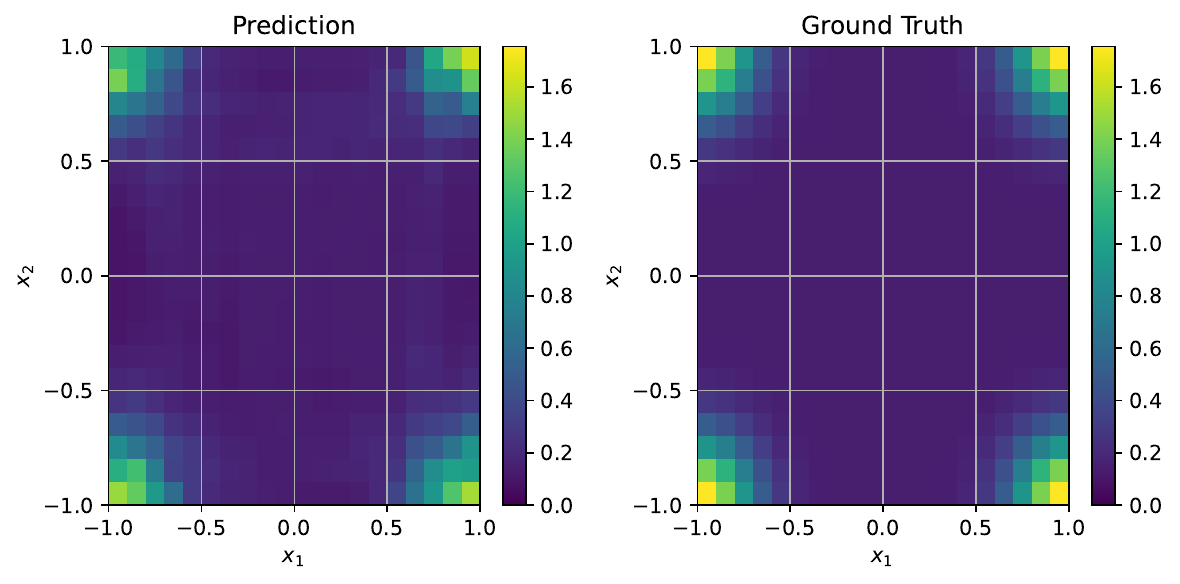}}
\caption{Histograms comparison for inverse transport}
\label{inv_hist}
\end{figure}

We train our model for $\mathrm{steps} = 2000$ iterations in total.
The training process alternates between updating forward and inverse mappings, where the forward network $T_{\theta}$ and the inverse network $T^{-1}_{\theta}$ are implemented as fully connected modified MLP, where $n_{\textrm {in }}=3$ and $n_{\textrm{out }}=2$ with three hidden layers, each consisting $32$ neurons. 
We update training batch and cost matrices for $T_{\theta}$ and $T^{-1}_{\theta}$ every $n_\gamma=50$ iterations. Figure \ref{inv_hist} provides the histograms of forward and inverse network predictions. 

\subsection{CSIR model}\label{Appendix for CSIR}
Following the setup in \cite{cui2023self}, the true parameters ${x}_{\text {true }}=[0.1,1, \ldots, 0.1,1]$, and the observations are simulated by 
\begin{align}
{y}_{i, j}=I_i\left(t_j ; {x}_{\text {true }}\right)+\alpha_{i, j}, \quad t_j=\frac{5 j}{6}, \quad j=1, \ldots, 6,
\label{CSIR posterior}
\end{align}
where $\alpha_{i, j} \sim \mathcal{N}\left(0, 1\right)$ is a zero-mean standard Gaussian noise. The likelihood function then reads as 
\begin{align}
\mathcal{L}^y({x}) \propto \exp \left(-\Phi^y({x})\right), \quad
\Phi^y({x})=\frac{1}{2} \sum_{i=1}^d \sum_{j=1}^6\left[I_i\left(t_j ; {x}\right)-{y}_{i, j}\right]^2.
\label{CSIR likehood}
\end{align}

We consider four CSIR models with $d=1,2,3,4$ compartments respectively, and for each dimension $d$, we generate $n_\kappa = 8$ sets of $N_0=5000$ paired prior and posterior samples. 
For each model the transport map is realized by a ResNet, consisting of three residual blocks with $128$ neurons per layer. 
Training is conducted over $\mathrm{steps}=500$ iterations. We refresh both the transport plan $\gamma$ and the training batches every $n_{\gamma}=10$ iterations.

\subsection{Image-to-Image color transfer}\label{appendix for ct}
We employ two unconditioned modified MLPs for the two color transfer tasks, each consisting of three hidden layers of 128 neurons with $n_{\textrm{in}}=n_{\textrm{out}}=3$. 

\begin{figure}[htbp]
\centering
\subfloat[Original images\label{ColorMLP0_boat}]{ 
\includegraphics[width=0.45\columnwidth]{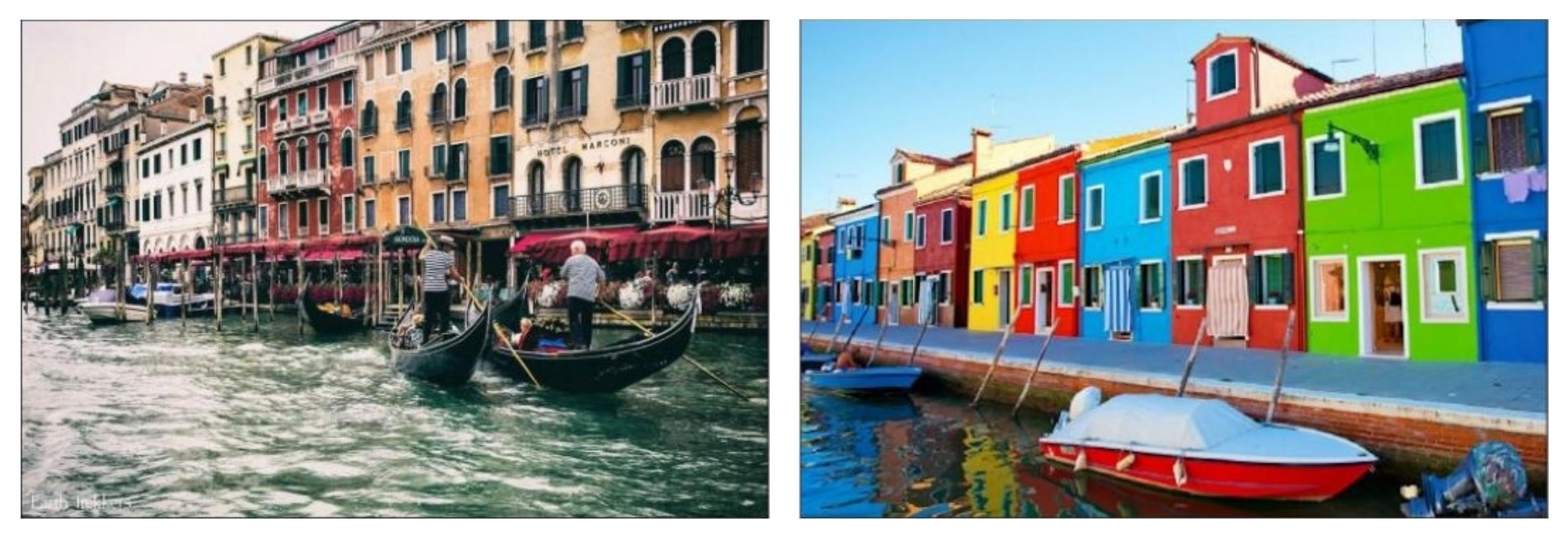}}
\subfloat[MLP+$P_{\lambda=0.3}(T_\theta)$\label{color_boat_MLP}]{ 
\includegraphics[width=0.45\columnwidth]{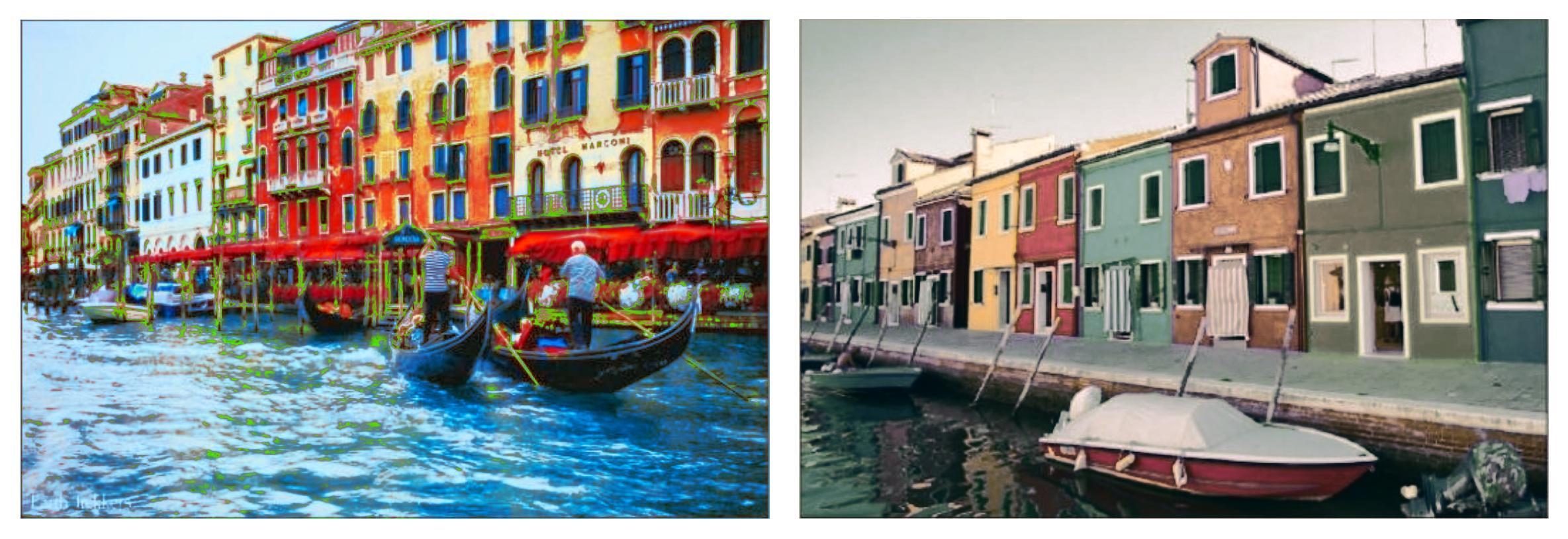}}
%\vspace{-1em}

\subfloat[ICNN+$P_{\lambda=0.3}(T_\theta)$\label{color_icnnlambN0_boat}]{ 
\includegraphics[width=0.45\columnwidth]{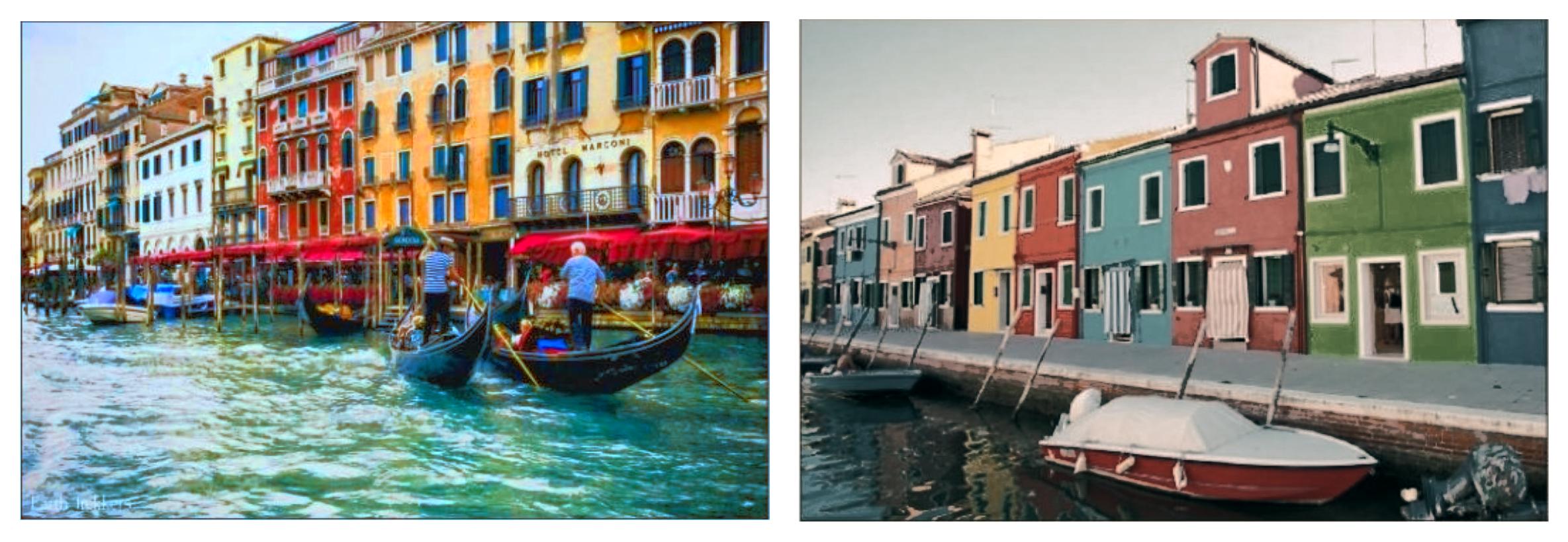}}
\subfloat[ICNN+W2L\eqref{W2GN-loss}\label{color_icnnw2l_boat}]{
\includegraphics[width=0.45\columnwidth]{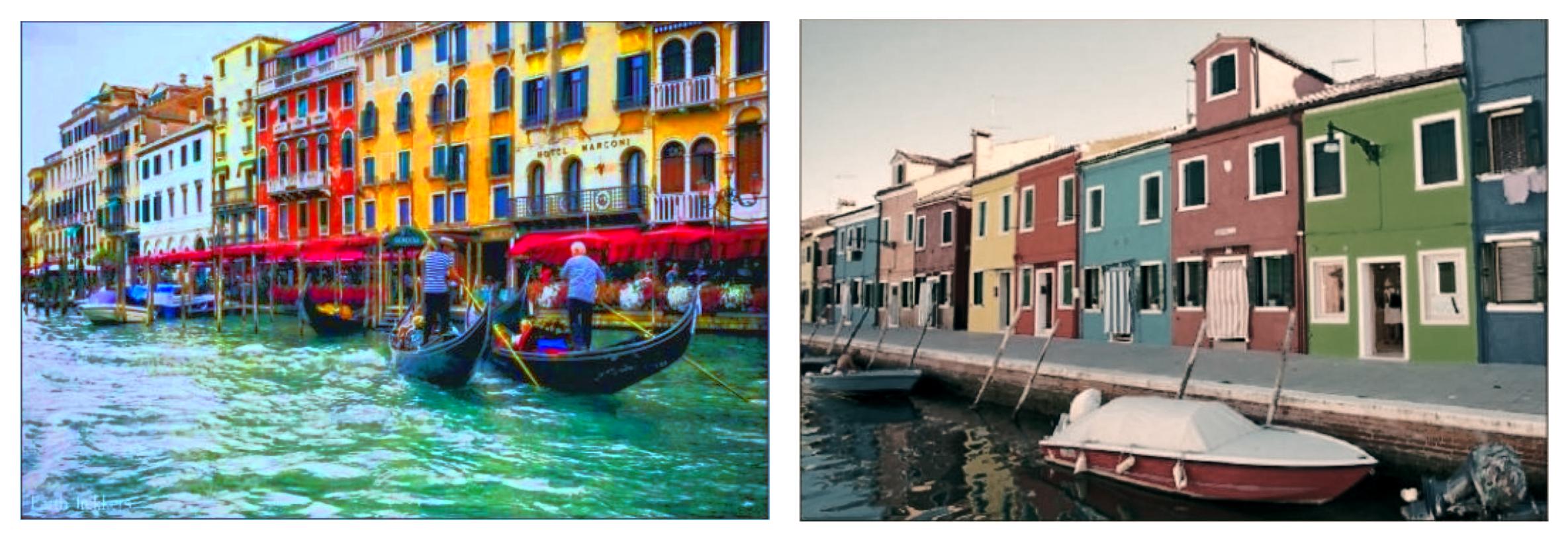}}
%\vspace{-1em}
\caption{Comparison of color transferred results for the second task}
\label{color_com2}
\end{figure}

The training dataset for each task is constructed by pairing corresponding RGB pixels from the source and target images. Since all images have a resolution of $1200\times 800$, this yields $N_0=960000$ pairs of data per task.  
We train for $\mathrm{steps}=1000$ iterations for both experiments.  For Figure~\ref{color_fost_MLP}, the training batch and transport plan $\gamma$ are refreshed every $n_\gamma = 50$ iterations, whereas for Figure~\ref{color_boat_MLP}, they are refreshed every $n_\gamma = 20$ iterations.

For the other three ICNN-based settings, we apply two DenseICNN networks $D$ and $D_{\textrm{conj}}$ with input dimension $n_\mathrm{in}=3$, rank $r=3$, $\mathrm{strong}$-$\mathrm{convexity \;constant}=10^{-6}$ and $\mathrm{dropout\; rate}=10^{-5}$. The gradients of $D$ and $D_{\textrm{conj}}$ define a forward and inverse transport map $\nabla \psi$ and $\nabla \psi^*$, respectively. The cycle-consistency regularizer coefficient $\eta$ in \eqref{W2GN-loss} is $6$. All parameters of $D$ are initialized from a Gaussian distribution $\mathcal{N}\left(0, \frac{1}{4}\right)$. We pretrain $D$ to approximate the identity map by minimizing the loss  
\begin{align*}
\mathcal{L}_{\textrm{pre }}=\mathbb{E}_{X \sim U((-0.5, 1.5)^3)}\|D(X)-X\|^2+10^{-10}\|D\|_{L^1},
\end{align*} 
until either $\mathcal{L}_{\textrm{pre}}<10^{-3}$ or $\mathrm{steps}=10000$. Once pretraining is complete, the final weights of $D$ are copied to initialize the conjugate network $D_{\textrm{conj}}$. We use Adam optimizer with learning rate $\mathrm{LR}=10^{-3}$ and momentum decay rates $\vec{\beta} = (\beta_1 = 0.8, \beta_2 = 0.9)$. This pretraining procedure follows the descriptions in Appendix C.1 of \cite{korotin2019wasserstein}.

For the first experiment shown in Figure \ref{color_com1}, we apply the DenseICNN networks with three hidden layers of 128, 128 and 64 neurons respectively, while for the second one displayed in Figure \ref{color_com2}, we adopt DenseICNN networks with three hidden layers of 64 neurons each.

\begin{figure}[htbp]
\centering
\subfloat[$t=0.2$\label{Color_fost_interp_2}]{ 
\includegraphics[width=0.7\columnwidth]{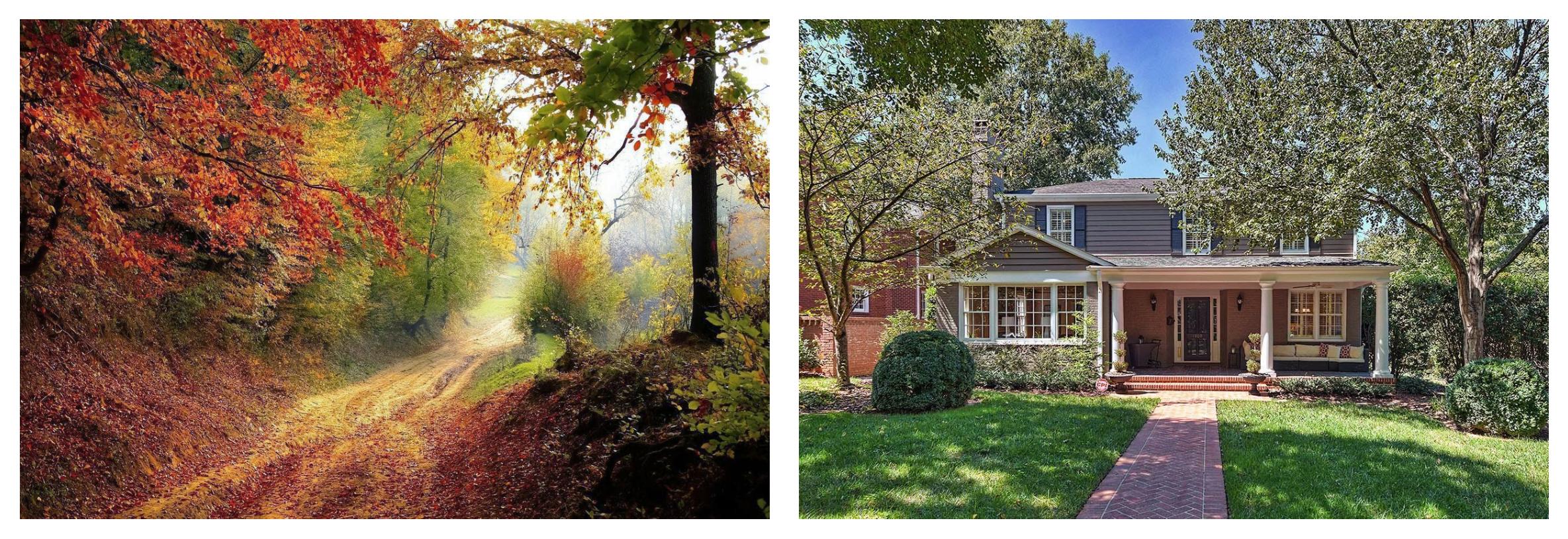}}

\subfloat[$t=0.4$\label{Color_fost_interp_4}]{ 
\includegraphics[width=0.7\columnwidth]{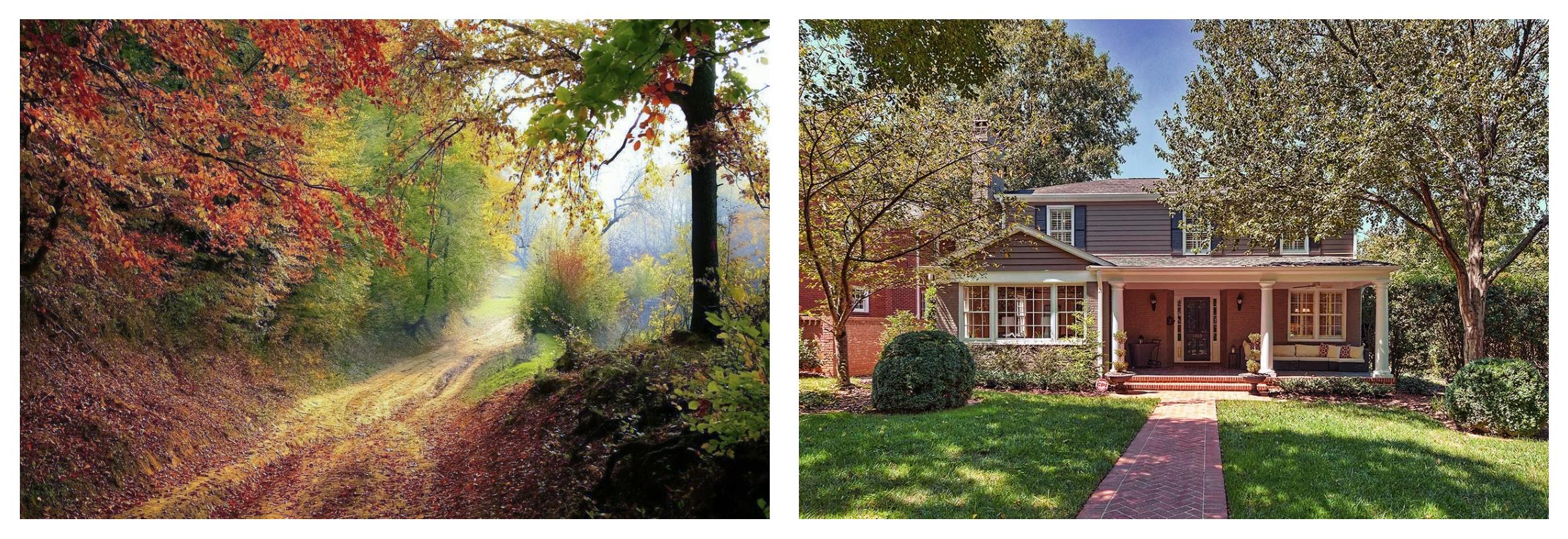}}

\subfloat[$t=0.6$\label{Color_fost_interp_6}]{ 
\includegraphics[width=0.7\columnwidth]{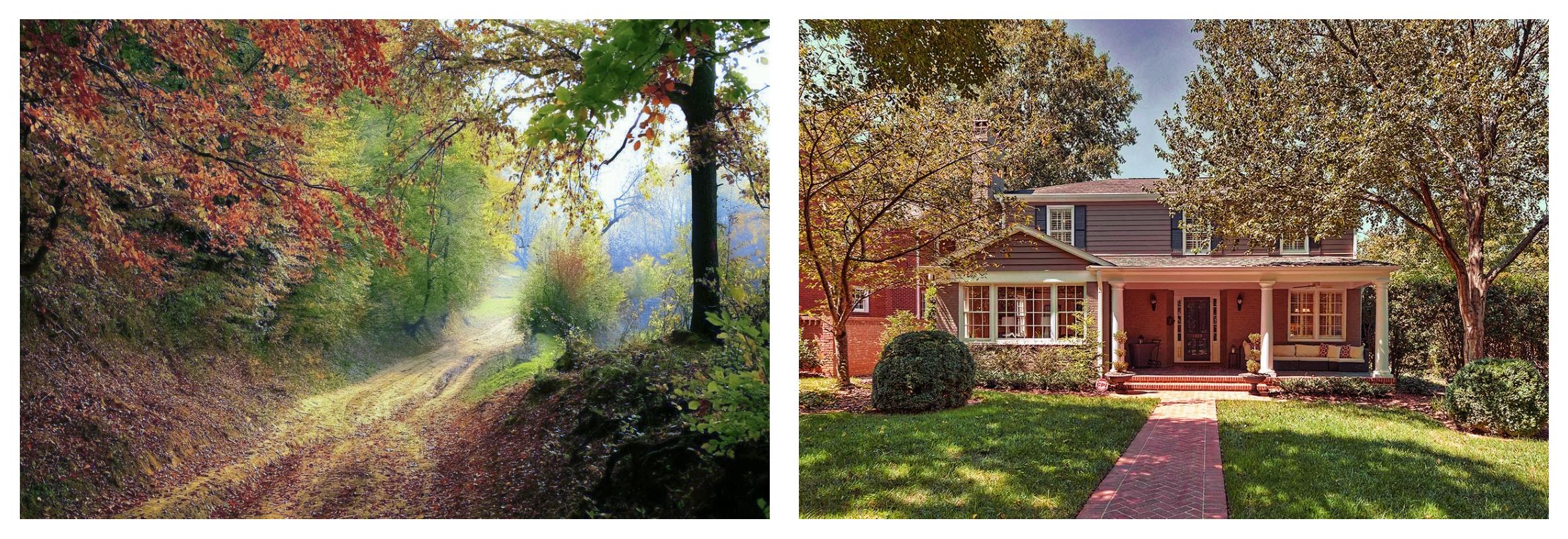}}

\subfloat[$t=0.8$\label{Color_fost_interp_8}]{ 
\includegraphics[width=0.7\columnwidth]{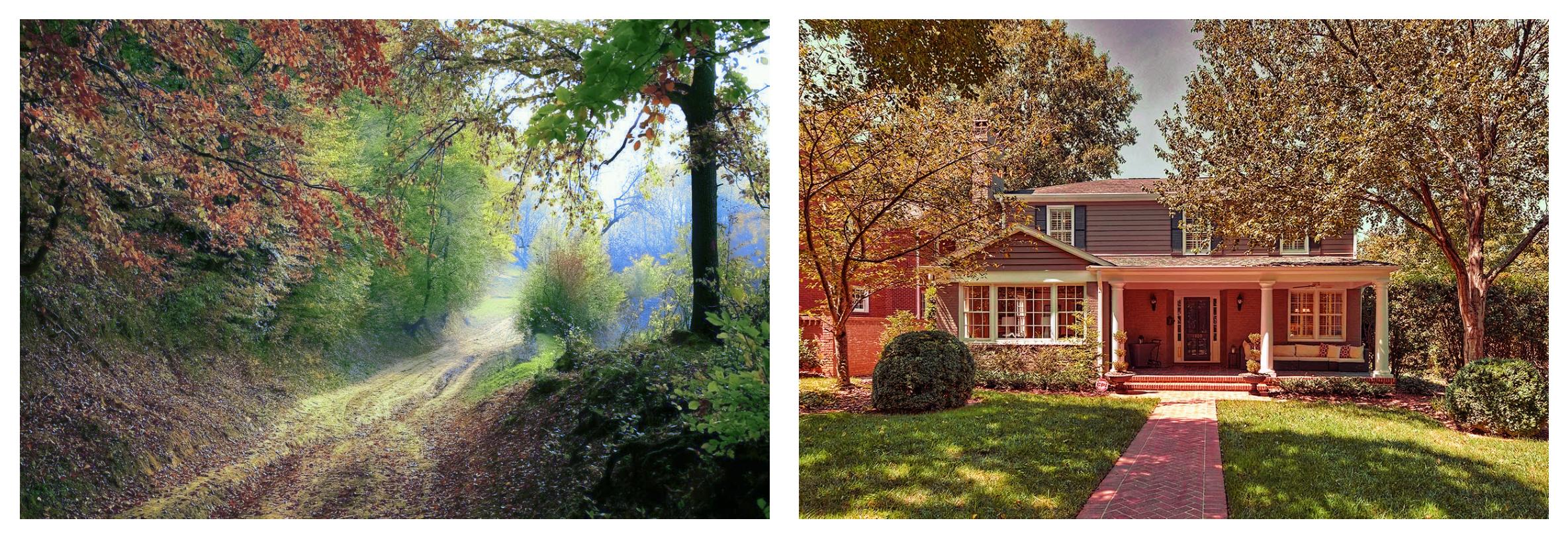}}

\subfloat[$t=1.0$\label{Color_fost_interp_10}]{ 
\includegraphics[width=0.7\columnwidth]{Figures/color_transfer/MLP_fost.jpeg}}
\caption{Interpolations of displacement $\tilde{T}(x)=(1-t) x +  tT_{\theta}(x)$ for modified MLP}
\label{color_process1}
\end{figure}

\begin{figure}[htbp]
\centering
\subfloat[$t=0.2$\label{Color_boat_interp_2}]{ 
\includegraphics[width=0.7\columnwidth]{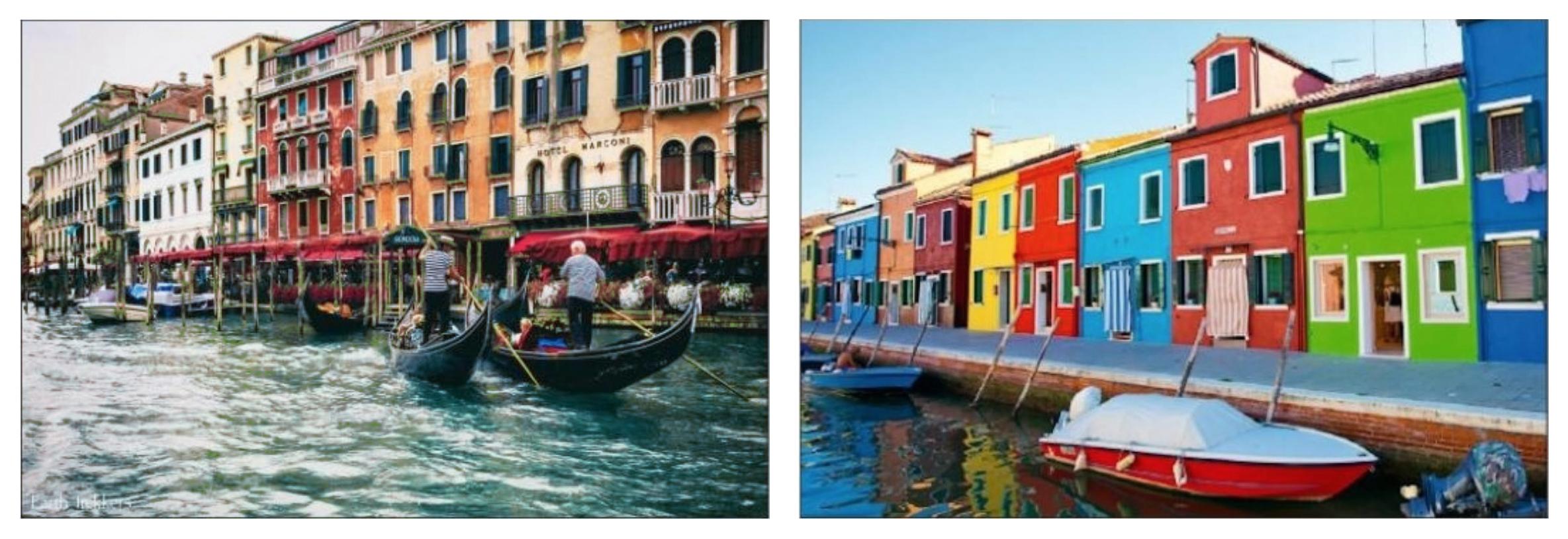}}

\subfloat[$t=0.4$\label{Color_boat_interp_4}]{ 
\includegraphics[width=0.7\columnwidth]{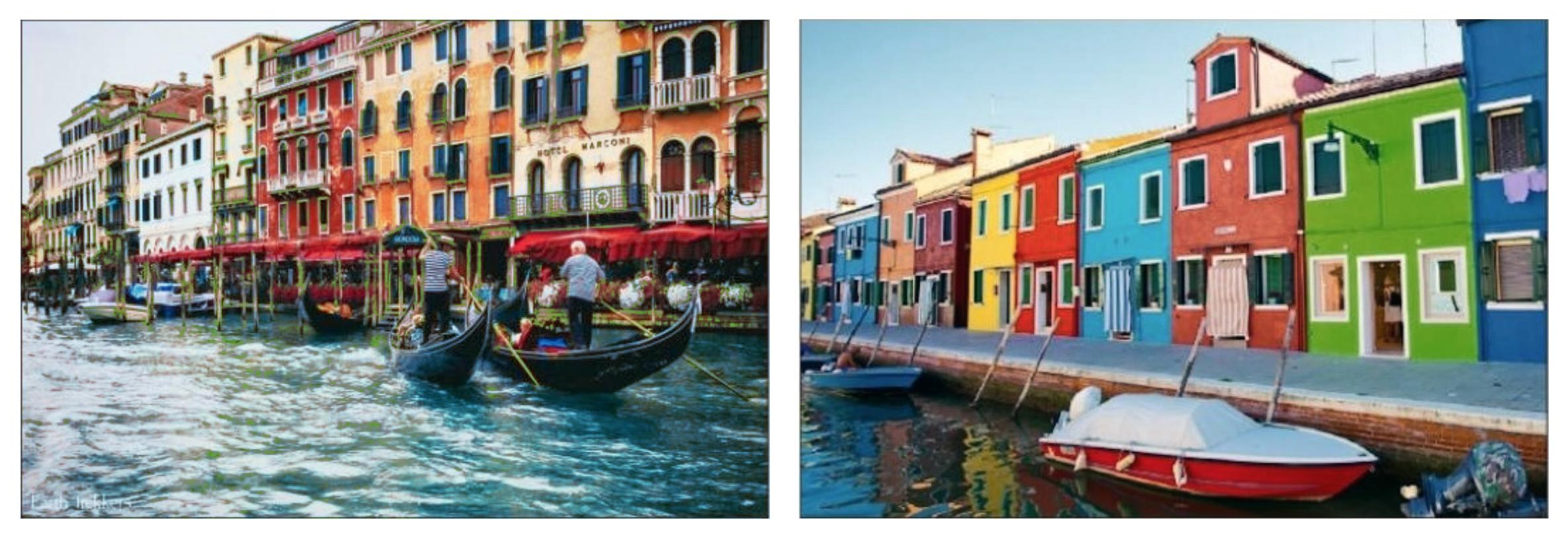}}

\subfloat[$t=0.6$\label{Color_boat_interp_6}]{ 
\includegraphics[width=0.7\columnwidth]{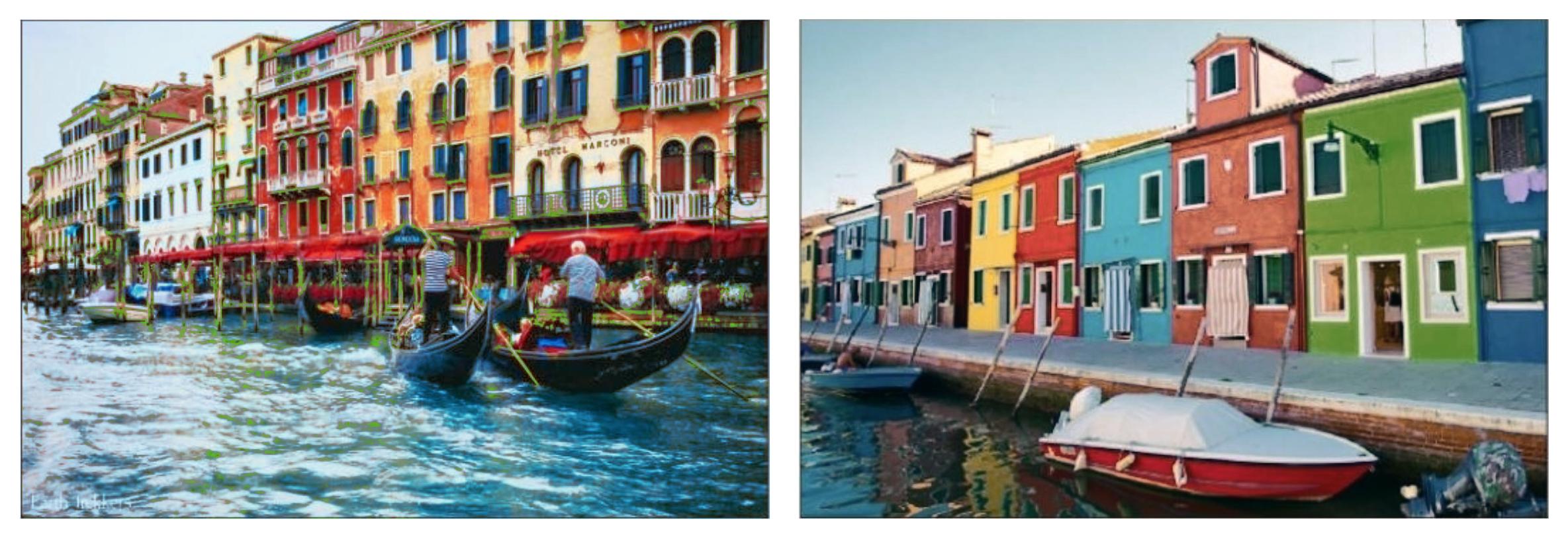}}

\subfloat[$t=0.8$\label{Color_boat_interp_8}]{ 
\includegraphics[width=0.7\columnwidth]{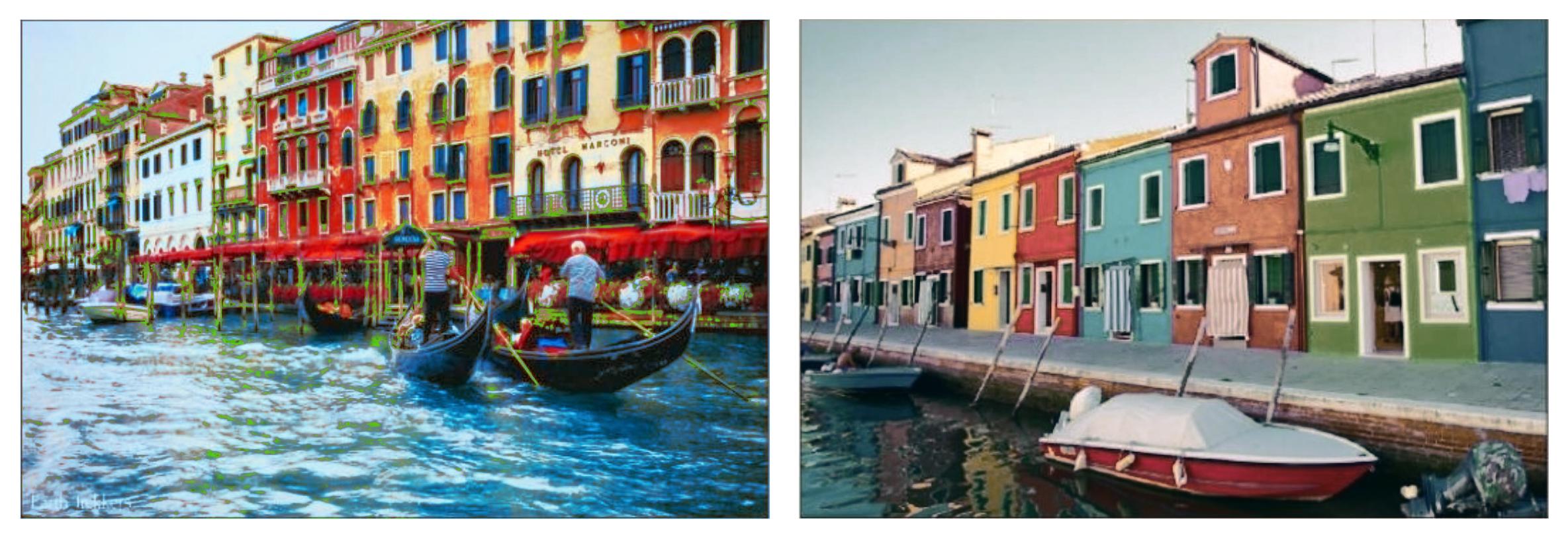}}

\subfloat[$t=1.0$\label{Color_boat_interp_10}]{ 
\includegraphics[width=0.7\columnwidth]{Figures/color_transfer/MLP_boat.jpeg}}
\caption{Interpolations of displacement $\tilde{T}(x)=(1-t) x +  tT_{\theta}(x)$ for modified MLP}
\label{color_process2}
\end{figure}

\end{document}